\documentclass{article}

\usepackage[preprint]{neurips_2025}

\usepackage[utf8]{inputenc} 
\usepackage[T1]{fontenc}    
\usepackage{hyperref}       
\usepackage{url}            
\usepackage{booktabs}       
\usepackage{amsfonts}       
\usepackage{nicefrac}       
\usepackage{microtype}      
\usepackage{xcolor}         
\usepackage{multirow}
\usepackage{array}
\usepackage{enumitem}
\usepackage{xspace}
\usepackage{graphicx}
\usepackage{amssymb}
\usepackage{amsmath}
\usepackage[capitalize,noabbrev]{cleveref}
\usepackage{wrapfig}
\usepackage{caption}
\newcommand{\name}{{\sc CAMILA}\xspace}

\title{CAMILA: Context-Aware Masking for Image Editing with Language Alignment}

\author{%
  \textbf{Hyunseung Kim}\textsuperscript{\textmd{1}}\quad
  \textbf{Chiho Choi}\textsuperscript{\textmd{2}}\quad
  \textbf{Srikanth Malla}\textsuperscript{\textmd{2}}\quad
  \textbf{Sai Prahladh Padmanabhan}\textsuperscript{\textmd{2}}\quad \\
  \textbf{Saurabh Bagchi}\textsuperscript{\textmd{1}}\quad
  \textbf{Joon Hee Choi}\textsuperscript{\textmd{2}}\quad\\
  \textsuperscript{\textmd{1}}Purdue University\quad
  \textsuperscript{\textmd{2}}Samsung Semiconductor, USA\\
  \texttt{\{kim4061, sbagchi\}@purdue.edu}\\
  \texttt{\{chiho1.choi, srikanth.m, sai.prahladh, jh4.choi\}@samsung.com}
}

\begin{document}

\maketitle

\begin{abstract}

Text-guided image editing has been allowing users to transform and synthesize images through natural language instructions, offering considerable flexibility. However, most existing image editing models naively attempt to follow all user instructions, even if those instructions are inherently infeasible or contradictory, often resulting in nonsensical output. To address these challenges, we propose a context-aware method for image editing named as \name (\textbf{C}ontext-\textbf{A}ware \textbf{M}asking for \textbf{I}mage Editing with \textbf{L}anguage \textbf{A}lignment). \name is designed to validate the contextual coherence between instructions and the image, ensuring that only relevant edits are applied to the designated regions while ignoring non-executable instructions. For comprehensive evaluation of this new method, we constructed datasets for both single- and multi-instruction image editing, incorporating the presence of infeasible requests. Our method achieves better performance and higher semantic alignment than state-of-the-art models, demonstrating its effectiveness in handling complex instruction challenges while preserving image integrity.

\end{abstract}
\vspace{-1.5em}
\section{Introduction}
\label{sec:intro}

In recent years, the growing demand for visual content has made image editing essential across various fields. With advancements in technology, text-guided image editing has emerged as a powerful tool, enabling users to manipulate images using natural language instructions~\cite{brooks2023instructpix2pix,huang2023smartedit,fu2024guiding,tumanyan2023plug, guo2023focus,hertz2022prompt}. This innovation has streamlined the editing process, enabling users to perform sophisticated edits. Among these advancements, diffusion-based models have particularly excelled in image generation~\cite{DBLP:journals/corr/abs-2006-11239,rombach2022high,rombach2022text,nichol2021glide,zhang2023adding,xue2024raphael,balaji2022ediff} and editing tasks~\cite{kawar2023imagic,couairon2022diffedit,hertz2022prompt,brooks2023instructpix2pix,tumanyan2023plug,zhang2023adding}. However, models relying on simple text encoders such as CLIP~\cite{radford2021learning} struggle to achieve user-intended fine-grained edits. These difficulties become more apparent when the editing prompt involves multi-step instructions with intricate details.

To address this limitation, recent research has introduced two notable improvements in model design. First, the CLIP-like text encoder has been replaced by Multimodal Large Language Models (MLLMs)~\cite{huang2023smartedit,fu2024guiding}. These models effectively parse user instructions and interpret textual prompts, improving the capabilities of natural language understanding. Second, regions requiring editing within the image are identified and modified using various methods, such as cross-attention maps and segmentation models, to align each edit prompt with its corresponding regions~\cite{guo2023focus,li2024zone}. Although the region-based image editing model~\cite{guo2024focus} shows more effective results on multi-instruction tasks than other state-of-the-art methods, its attention maps often fail to consistently align with intended editing regions. This misalignment is especially pronounced when modifications involve spatial relationships or regions not directly associated with primary instruction keywords.

These limitations become evident in multi-instruction scenarios containing challenging instructions that cannot be directly applied to the current image. Such instructions may request alterations to non-existent objects, logically inconsistent modifications, or edits that are incompatible with the image’s content. Parsing and interpreting such inputs makes editing systems impractical, introducing suboptimal edits or even unrealistic, incoherent images. Additionally, relying on pretrained Large Language Models~\cite{gpt4v} to parse or reorganize these instructions introduces further complexity in the editing pipeline and increases the potential for errors at intermediate steps. Any misinterpretation or bias in LLM output may propagate downstream, leading to incorrect region selection or over-editing.

\begin{figure}
    \centering
    \includegraphics[width=0.95\linewidth]{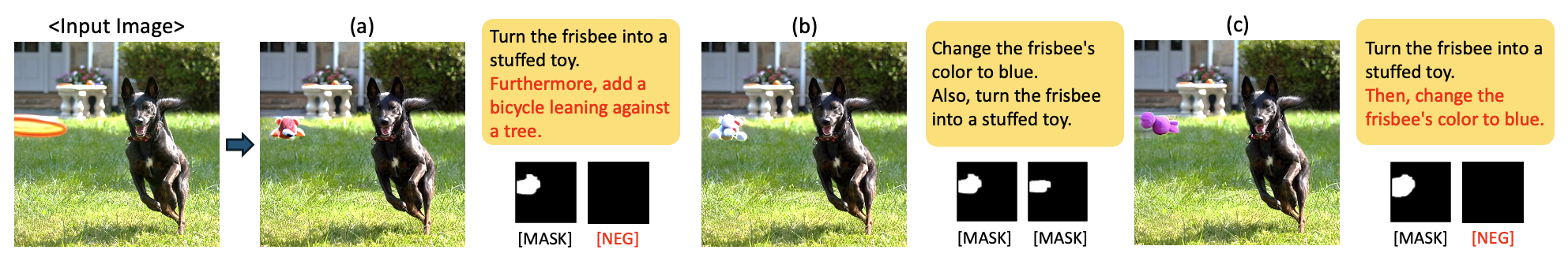}
    \vspace{-0.5em}
    \caption{Three scenarios demonstrate how our method handles context-aware multi-instruction editing across various combinations of feasible and infeasible prompts.
    By leveraging \texttt{[MASK]} and \texttt{[NEG]} specialized tokens, it accurately identifies executable instructions.}
    \label{fig:teaser}
    \vspace{-1.5em}
\end{figure}

Despite the growing research interest in comprehensive image editing, most existing methods overlook instruction executability, often leading to over-edited results. Our proposed approach addresses these concerns by explicitly assessing the executability of the instruction throughout the editing process. Building on pioneering research in this domain, we leverage the MLLM to jointly interpret both text instructions and images, then we extend its capabilities to enable image editing with context awareness. Here, \textit{context} refers to the model’s ability to interpret the relevance of various instructions within a given image, allowing it to focus on applicable regions while ignoring irrelevant areas. A key feature of \name is the use of specialized tokens and broadcast mechanism. Our model assigns \texttt{[MASK]} tokens to editable regions and \texttt{[NEG]} tokens to suppress irrelevant edits. The following broadcasting module then consistently aligns token assignments with user prompts. Overall, our context-aware pipeline helps to validate the coherence of instructions, resulting in improved performance across all image editing scenarios, including non-executable prompts.

To properly evaluate our approach, we extend the conventional single- and multi-instruction image editing tasks by introducing the possibility of non-executable prompts. This results in new evaluation scenarios: \textit{Context-Aware Image Editing} that evaluate how the model handles the number of instructions and the presence of infeasible requests within the same sequence. We compare our method against several state-of-the-art baselines, observing substantial improvements in editing accuracy, particularly L1 and L2 distances, as well as enhanced performance on CLIP and DINO scores, with a human preference-based evaluation also indicating strong performance.

Our main contributions to this work are as follows:
\begin{itemize}[leftmargin=1em, itemsep=0em, topsep=0pt]
    \item We introduce a context-aware image editing model that precisely identifies prompt executability and corresponding editing regions, allowing user-aligned and consistent modifications.
    \vspace{-0.25em}
    \item We propose a new task setting: \textit{Context-Aware Image Editing}. New datasets are created to evaluate model behavior and context-awareness in challenging scenarios.
    \vspace{-0.25em}
    \item Our model demonstrates significant improvements over existing methods in varying evaluation scenarios, achieving lower pixel-level errors and higher semantic alignment, while also showing qualitative superiority in effectively handling complex instructions.
\end{itemize}

Note that we formally define `non-executable instruction' as any request that cannot be executed given the visual constraints or inherent semantics of the image.
\section{Related Works}

\noindent\textbf{Multimodal Large Language Models.} Multimodal Large Language Models (MLLMs)~\cite{li2023blip,liu2024visual,dai2023instruct, team2023gemini,zhu2023minigpt, liu2024improved} integrate multiple modalities, such as images and text. Recent MLLMs have advanced to handle complex tasks such as referring visual grounding~\cite{zhao2023bubogpt, lai2024lisa,chng2024mask, xia2024gsva}, which aims to distinguish specific objects based on context. Additionally, MLLMs have been applied to image editing task~\cite{huang2023smartedit, fu2024guiding}. For instance, SmartEdit~\cite{huang2023smartedit} improves instruction comprehension with bidirectional interactions between image and text, while MGIE~\cite{fu2024guiding} jointly trains an MLLM and diffusion model to guide editing tasks with visual-aware instructions. However, these models often lack context-awareness and fail to distinguish between relevant and irrelevant prompts. We thus break new ground by being the first to incorporate a context-aware MLLM specially for image editing. Unlike prior research, we do not limit our scope to single instruction tasks, enabling our model to handle both multi and context-aware instructions.

\begin{figure*}[t]
    \centering
    \includegraphics[width=0.88\linewidth]{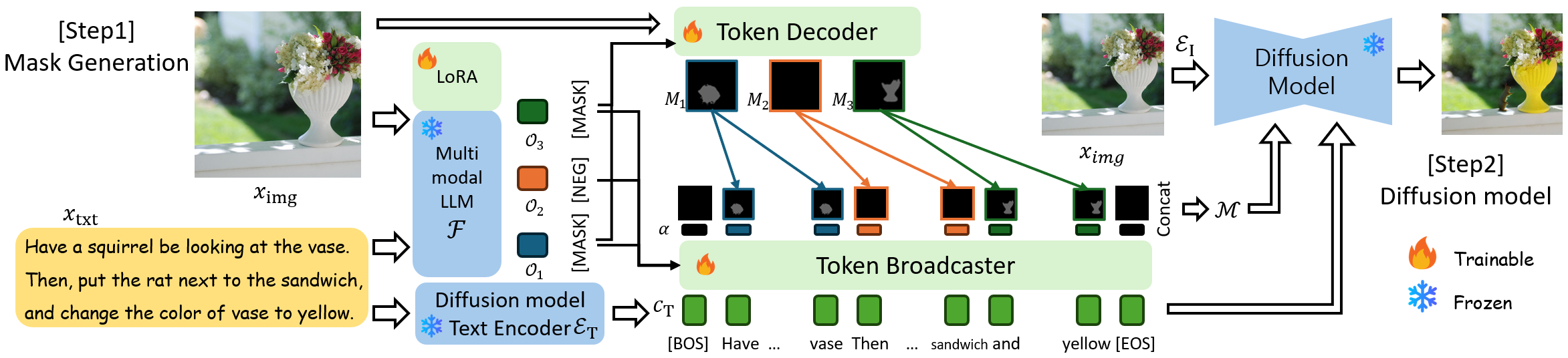}
    \caption{The architecture of \name begins by jointly processing the image \( x_{\text{img}} \) and text instructions \( x_{\text{txt}} \) using an MLLM. Output tokens are classified as either \texttt{[MASK]} or \texttt{[NEG]}, indicating regions to modify or leave unchanged. These tokens are aligned with the text embeddings using the Token Broadcaster, and the final binary mask is generated by the Token Decoder. The mask is then applied in a diffusion model to produce the edited image.}\vspace{-0.7em}
    \label{fig:architecture}
    \vspace{-0.7em}
\end{figure*}

\noindent\textbf{Image Editing by Diffusion Model.} Diffusion models have become prominent in image editing~\cite{richardson2021encoding,abdal2019image2stylegan,nichol2021glide,meng2021sdedit,kawar2023imagic,couairon2022diffedit,hertz2022prompt,tumanyan2023plug,zhang2023adding, zhang2024hive}. While text-guided image editing enable basic modifications, instruction-based image editing offers more nuanced control by interpreting complex, user-directed commands via natural language. InstructPix2Pix~\cite{brooks2023instructpix2pix} introduced a dataset combining GPT-3-generated texts~\cite{brown2020language} and Prompt2Prompt-based images~\cite{hertz2022prompt}, which powers natural language-guided editing. MGIE~\cite{fu2024guiding} utilizes an MLLM with visual-aware instructions for editing, and FoI~\cite{guo2023focus} uses cross-attention maps for multi-instruction scenarios. However, these methods struggle with ambiguous or incorrect instructions, as they lack mechanisms to interpret prompt feasibility. This limitation often leads to unintended modifications when the model encounters unclear instructions.
\section{Preliminary}

We briefly introduce InstructPix2Pix (IP2P)~\cite{brooks2023instructpix2pix}, a standard framework for instruction-guided image editing and its cross-attention mechanism. This overview serves as the background for our work.

\subsection{InstructPix2Pix}

IP2P~\cite{brooks2023instructpix2pix} is built upon Stable Diffusion~\cite{rombach2022high} to modify images based on textual instructions. In this framework, conditioning on both input image and text instructions is necessary for guiding diffusion network to produce editing results aligned with user instruction. The input image \( x_{\text{img}} \) is first encoded into a latent vector $z$ by the encoder $\mathcal{E}_{\text{I}}$. At each time step $t$, the noisy latent vector $z_t$ is progressively denoised by the score network. Then, the denoised latent vector $z$ is decoded into the output image.

To achieve conditional generation, diffusion models often employ classifier-free guidance~\cite{ho2022classifier}, which eliminates the need for an external classifier. In their score network, two conditioning factors are introduced for use during inference: the image conditioning $c_I$ and the text instruction conditioning $c_T$. $c_I$ and $c_T$ are the encoded outputs from the image encoder $\mathcal{E}_{\text{I}}$ and the text encoder $\mathcal{E}_{\text{T}}$, respectively. The final score estimation $\tilde{e_{\theta}}(z_t, c_I, c_T)$ is computed as follows:
\begin{equation}
\begin{split}
    \tilde{e_{\theta}}(z_t, c_I, c_T) = e_{\theta}(z_t, \varnothing, \varnothing)
    &+ s_I \cdot (e_{\theta}(z_t, c_I, \varnothing) - e_{\theta}(z_t, \varnothing, \varnothing)) \\
    &+ s_T \cdot (e_{\theta}(z_t, c_I, c_T) - e_{\theta}(z_t, c_I, \varnothing)).
    \label{eq:cfg2}
\end{split}
\end{equation}

In this equation, $e_{\theta}(z_t, \varnothing, \varnothing)$ represents the base score prediction without any conditioning applied. The second term modulates the score with image conditioning $c_I$, where $s_I$ modulates how much the model preserves the characteristics of the input image. Similarly, the last term incorporates text conditioning $c_T$, with $s_T$ controls the degree of adherence to the edit instruction provided.

\subsection{Cross Attention in Stable Diffusion}
\label{prelim_ca}

IP2P employs cross-attention network modulation within the denoising U-Net architecture of the Stable Diffusion network. A key component is the cross-attention layer, which generates attention maps $\mathcal{A} \in \mathbb{R}^{r \times r \times m}$, where $r$ is the spatial size and $m$ is the number of text tokens. Several studies~\cite{agarwal2023star,chefer2023attend,guo2023focus} have shown that cross-attention maps with $r = 16$ capture the most significant semantic information, compared to maps at other spatial resolutions. Thus, by modulating the computation of these cross-attention layers, it is possible to alter the image, as adjustments in the attention maps guide the model’s focus on specific aspects of the text and image content~\cite{couairon2022diffedit, xie2023smartbrush}.

\section{Methods}
\label{sec:proposedmethods}
\vspace{-0.5em}
 
We build our framework upon a pretrained MLLM~\cite{liu2024visual} and diffusion model~\cite{rombach2022high}, but our key contribution lies in explicitly assessing the executability of instruction and leveraging specialized tokens to guide editing process in diffusion model. A key feature of our approach is its ability to validate the contextual coherence between instructions and the image, ensuring that only relevant edits are applied to designated regions while ignoring non-executable instructions. This context-aware mechanism distinguishes our method from existing MLLM-based approaches~\cite{fu2024guiding, huang2023smartedit}.

\subsection{Architecture}
\label{sec:architecture}

The architecture of \name is shown in~\cref{fig:architecture}. Given an image \( x_{\text{img}} \) and text instructions \( x_{\text{txt}} \), both inputs are jointly processed by the MLLM \( \mathcal{F} \). The model is designed to encode and combine the visual and textual inputs, enabling it to capture the relationships between the textual instructions and corresponding regions in the image. Specifically, the image is processed through a vision encoder, while the text instructions are tokenized and processed by a language encoder. These representations are then combined into a unified sequence within the MLLM architecture, which interprets the joint context of the image and instructions. The output sequence \( \mathcal{O}\) is generated from the image input \( x_{\text{img}} \) and text input \( x_{\text{txt}} \). Each output token \( \mathcal{O}_i \) in \( \mathcal{O} = \{\mathcal{O}_1, \mathcal{O}_2, \dots, \mathcal{O}_n\} \), where $n$ denotes the number of generated tokens, is classified as either a \texttt{[MASK]} or \texttt{[NEG]} token. The \texttt{[MASK]} tokens correspond to regions of the image that are to be modified based on the text instructions, while the \texttt{[NEG]} tokens indicate areas of the image that should remain unaffected.

\begin{wrapfigure}{l}{0.45\textwidth}
    \centering
    \vspace{-0.5em}
    \includegraphics[width=0.85\linewidth]{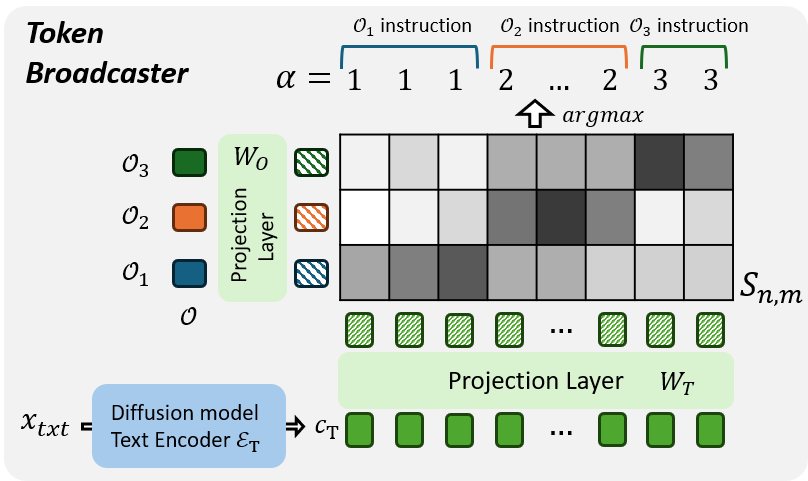}
    \caption{\textbf{Architecture of the Token Broadcaster.} It calculates similarity between MLLM output tokens and encoded text features, assigning each output token to the text embedding that best matches its corresponding semantic region.}
    \label{fig:module}
    \vspace{-1.0em}
\end{wrapfigure}

By combining the visual and textual inputs, the MLLM is able to determine the relevance of each instruction to specific regions in the image, ensuring that only applicable edits are applied. This joint processing aligns each generated output token, either \texttt{[MASK]} or \texttt{[NEG]}, with specific instructions. The \texttt{[MASK]} tokens are decoded, resulting in masks that accurately highlight the regions in the image that require modification according to the instructions. This targeted approach improves the precision of the editing process by ensuring that modifications are applied solely to relevant areas. The following content will elaborate on how \texttt{[MASK]} and \texttt{[NEG]} tokens are aligned with instructions by Token Broadcaster and how \texttt{[MASK]} tokens are decoded into the actual editing mask by Token Decoder.

\subsection{Token Broadcaster and Token Decoder}

\noindent \textbf{Token Broadcaster.} The output sequence \( \mathcal{O} \) generated by the MLLM is processed by the Token Broadcaster module to ensure that the \texttt{[MASK]} and \texttt{[NEG]} tokens align accurately with the corresponding text embeddings. As illustrated in \cref{fig:module}, the text instructions \( x_{\text{txt}} \) are embedded through the text encoder $\mathcal{E}_{\text{T}}$ of the diffusion model, resulting in a set of text embeddings $c_T$. Using the diffusion model's text encoder $\mathcal{E}_{\text{T}}$ allows the model to ensure that the generated editing masks will align precisely with $c_T$, facilitating integration into the diffusion model.

The MLLM output tokens \( \mathcal{O} \) and the text embeddings $c_T$ reside in different latent spaces, so we need to align them into a single space. Many studies~\cite{kim2022transferring, yuan2021multimodal} use cosine similarity-based alignment to measure and organize relationships or similarities between different modalities. We project them into a shared space for alignment by applying trainable transformations \( W_O \) and \( W_T \) to each, directly within the similarity matrix:
\begin{equation}
S_{i,j} = \frac{(\mathcal{O}_i W_O) \cdot (c_{Tj} W_T)}{\|(\mathcal{O}_i W_O)\| \|(c_{Tj} W_T)\|},
\end{equation}
where each element \( S_{i,j} \) represents the cosine similarity score between the \( i \)-th transformed output token \( (\mathcal{O}_i W_O) \) and the \( j \)-th transformed text embedding \( (c_{Tj} W_T) \), indicating their compatibility in the shared latent space.

To convert similarity scores into alignment probabilities, a softmax is applied along each column of \( S \). For each text embedding \( j \), we then determine the index \( \alpha_j \) that maximizes this probability:
\begin{equation}
    \alpha_j = \arg\max_i \left( \frac{\exp(S_{i,j})}{\sum_{k} \exp(S_{k,j})} \right), \forall j \in \{1, 2, \dots, m\},
\end{equation}
where $m$ denotes the length of text embeddings. This alignment process ensures that each text embedding maps to the output token best reflecting its semantic region within the image.

\noindent \textbf{Token Decoder.} The Token Decoder processes tokens differently based on their type: only tokens labeled as \texttt{[MASK]} are converted into editing masks, while \texttt{[NEG]} tokens are directly replaced with black masks, indicating regions where no modification is applied. Designed as a two-layer Transformer decoder, the Token Decoder generates a set of binary masks \( M_1, M_2, \dots, M_n \), each specifying regions of the image to be edited according to the text instructions.

In the first decoder layer, we employ a cross-attention mechanism between image and text embeddings. This allows the model to extract contextually relevant features from the image that are aligned with the text instructions. By attending to both modalities, the decoder effectively maps the semantic content of the text to corresponding regions in the image. The second decoder layer further refines this information by incorporating the \texttt{[MASK]} tokens into the key and value projections of the attention mechanism. This enables the model to focus more precisely on the regions identified by each \texttt{[MASK]} token. After the second decoder layer, these intermediate masks are passed through sigmoid thresholding to produce the final 0-1 binary masks, denoted as \( M_i \). Through this process, Token Decoder is able to generate the final binary mask \( M_i \), with each mask serving as an editing mask for the corresponding MLLM output tokens \( \mathcal{O}_i \), defining the specific areas of the image to be modified.

\subsection{Diffusion Model}

For each text embedding \( j \), the alignment index \( \alpha_j \) determines the specific binary mask \( M_{\alpha_j} \) to be used. The individual masks are concatenated to form a unified binary mask \( \mathcal{M} \), which is then used in the diffusion model to guide the editing process:
\begin{equation}
\mathcal{M} = \text{concat}( M_{\alpha_1}, M_{\alpha_2}, \dots, M_{\alpha_m}).
\end{equation}
This binary mask \( \mathcal{M} \) ensures that each region is modified according to alignment indices from the Token Broadcaster, enabling precise, context-aware edits that reflect the intended modifications.

We modulate the cross-attention layers of the diffusion model, focusing specifically on the 16-sized cross-attention map, which captures the most semantically relevant features, as explained in \cref{prelim_ca}. The U-Net's cross-attention map \( \mathcal{A} \) is modulated using the following equation:
\begin{equation}
\label{eq:modulating_attention}
\mathcal{A'} = \text{softmax}\left(\frac{\mathcal{X} \odot \mathcal{M} + \mathcal{Y} \odot (1-\mathcal{M})}{\sqrt{d}}\right),
\end{equation}
where $d$ is the latent projection dimension, $\mathcal{X} = Q_{I,T}K^\mathsf{T}_{I,T}$, and $\mathcal{Y} = Q_{I,\varnothing}K^\mathsf{T}_{I,\varnothing}$. In this formulation, \( Q_{I,T} \) and \( K_{I,T} \) represent the query and key projections in \( e_{\theta}(z_t, c_I, c_T) \), respectively, while \( Q_{I,\varnothing} \) and \( K_{I,\varnothing} \) are the query and key projections in \( e_{\theta}(z_t, c_I, \varnothing) \). 

This modulation approach leverages \( \mathcal{A} \) to align each text embedding precisely with the regions specified by the concatenated binary mask \( \mathcal{M} \), enhancing editing accuracy by concentrating on the relevant areas as dictated by the instructions. Then, the binary mask \( \mathcal{M} \) selectively applies the text-conditioned attention map \( \mathcal{X} \) to editable regions and \( \mathcal{Y} \) to unaltered areas, ensuring that only the specified areas are modified. By modulating the attention layer as in \cref{eq:modulating_attention}, we generate the final output image following the score estimation formulated in \cref{eq:cfg2}.

\subsection{Training Details}
\noindent \textbf{Training Loss Function.} The training of our MLLM-based approach is optimized with four primary loss components, each designed to target a specific aspect of model performance for accurate token classification, alignment, and mask generation. The total loss \( \mathcal{L}_{\text{main}} \) is formulated as follows:
\begin{equation}
\mathcal{L}_{\text{main}} = \lambda_1 \mathcal{L}_{\text{CE}}^{\text{token}} + \lambda_2 \mathcal{L}_{\text{CE}}^{\text{broadcast}} + \lambda_3 \mathcal{L}_{\text{dice}} + \lambda_4 \mathcal{L}_{\text{BCE}},
\label{losseq}
\end{equation}
where \( \lambda_1, \lambda_2, \lambda_3, \lambda_4 \) are hyperparameters that balance the influence of each loss component.

The first element, token classification loss \( \mathcal{L}_{\text{CE}}^{\text{token}} \), applies cross-entropy (CE) loss to the MLLM output tokens. The second element, broadcasting alignment loss \( \mathcal{L}_{\text{CE}}^{\text{broadcast}} \), also utilizes CE loss to align MLLM output tokens with their respective text embeddings, ensuring precise correspondence between instructions and image regions. For mask quality, the mask dice loss \( \mathcal{L}_{\text{dice}} \) measures overlap between predicted and ground truth masks, encouraging accurate spatial targeting. Lastly, the binary cross-entropy loss \( \mathcal{L}_{\text{BCE}} \) enforces accuracy at the pixel level in the generated mask.

\noindent \textbf{Trainable Parameters.} To efficiently fine-tune the pre-trained MLLM while preserving its learned knowledge, we adopt the Low-Rank Adaptation technique~\cite{hu2021lora}. In our training, we freeze the vision backbone and text encoder of the MLLM, while the remaining parts of the model are fine-tuned. Additionally, the Token Broadcaster and Token Decoder are also trained, ensuring that the model aligns the output tokens with the text instructions and generates accurate masks for the diffusion model.
All other training details are provided in \cref{appendix:implementation}.

\subsection{Surrogate Module Training for Enhanced Masking}

To further improve the quality of the binary mask \( \mathcal{M} \) provided to the diffusion model, we conduct additional training beyond the initial MLLM training. Through empirical analysis, we found that certain outputs misalign with the description of the goal image. To better align the generated image with the intended modifications, we consider it useful to focus on improving CLIP-T score, which measures the similarity between the global description and the generated image. By optimizing the model for a higher CLIP-T score, we aim to generate higher quality binary masks, which lead to improved quality in the final output image.

However, due to the inherent complexity and the large number of steps involved in the forward pass of the diffusion model, directly backpropagating the loss from the final output image through the diffusion model to the MLLM is infeasible. To address this limitation, we develop a lightweight surrogate module that approximates the CLIP-T score based on the input image \( x_{\text{img}} \), the edit instruction \( x_{\text{txt}} \), and the binary masks \( \mathcal{M} \). Designed as a single-layer transformer, the surrogate module offers a streamlined alternative to the complex, multi-step diffusion model. It is trained using a mean squared error (MSE) loss between the actual CLIP-T score and the predicted CLIP-T score. During this training phase, all other parts of the model are kept frozen, and only the surrogate module is updated. The overall loss function for training the surrogate module is formulated as:
\begin{equation}
\mathcal{L}_{\text{surrogate}} = \mathbb{E} \left[ \left( \text{CLIP-T}_{\text{output}} - \text{CLIP-T}_{\text{surrogate}} \right)^2 \right],
\end{equation}
where \(\text{CLIP-T}_{\text{output}}\) and \(\text{CLIP-T}_{\text{surrogate}}\) denote the actual CLIP-T score of the target output and the predicted score, respectively.
This approach ensures that the surrogate module learns to accurately estimate the CLIP-T score without requiring multi-step backpropagation of the diffusion model.

\noindent \textbf{Refining Mask Generation via Surrogate Module.} Once the surrogate module is fully trained, we use estimated values to fine-tune the MLLM, Token Broadcaster, and Token Decoder. In this stage, the surrogate module is kept frozen, and the focus is on improving mask generation to maximize the predicted CLIP-T score. The objective is to modify the MLLM's outputs to generate binary masks with a higher CLIP-T score when processed by the diffusion model.

During training, the loss function is augmented to include both $\mathcal{L}_{\text{main}}$ as well as the MSE loss between the predicted CLIP-T score and the oracle CLIP-T score. The updated loss  $\mathcal{L}_{\text{updated}}$ is defined as:
\begin{equation}
\mathcal{L}_{\text{updated}} = \mathcal{L}_{\text{main}}  + \lambda_5 \mathcal{L}_{\text{MSE}},
\end{equation}
where \(\mathcal{L}_{\text{MSE}} \) is the MSE loss between the predicted and oracle CLIP-T score, and \( \lambda_5 \) is a hyperparameter controlling the weight of the CLIP-T score loss.

\section{Evaluation}

\subsection{Task Categorization}
For a comprehensive assessment, we evaluate our method on both single-instruction tasks aligned with standard benchmarks, and multi-instruction image editing tasks that require multiple edit turns in a single sequence. In a single-instruction scenario, a single directive is tested either in a single-turn or multi-turn setting, whereas multi-instruction tasks involve multiple directives that must be applied simultaneously. We further divide multi-instruction tasks into two types: \textit{Multi-instruction Image Editing}, which includes only applicable instructions, and \textit{Context-Aware Instruction Image Editing}, which includes a mix of applicable and non-applicable instructions.

\begin{table*}[t]
\centering
\setlength{\tabcolsep}{4pt}
\small
\vspace{-0.5em}
\caption{\textbf{Quantitative comparison across multi-instruction and context-aware instruction tasks.} Our model demonstrates overall superior performance, especially excelling in the context-aware instruction task. This highlights our method’s superb capability to handle context-aware instructions with high precision, applying edits that closely align with the intended modifications without over-editing. \textbf{Bold} and \underline{underlining} indicates the best and the second-best performance for each metric.}
\resizebox{\linewidth}{!}{
\begin{tabular}{cccccc|ccccc}
\toprule
 &\multicolumn{5}{c|}{\textbf{Multi Instruction}} &\multicolumn{5}{c}{\textbf{{Context-Aware Instruction}}}\\ \midrule

 Method & L1$\downarrow$ & L2$\downarrow$ & CLIP-I$\uparrow$ & DINO$\uparrow$ & CLIP-T$\uparrow$ & L1$\downarrow$ & L2$\downarrow$ & CLIP-I$\uparrow$ & DINO$\uparrow$ & CLIP-T$\uparrow$ \\ \midrule

IP2P~\cite{brooks2023instructpix2pix} 
& 0.1402 & 0.0526 & 0.8327 & 0.7122 & \underline{0.2977} 
& 0.1460 & 0.0514 & 0.7975 & 0.6429 & 0.2715 \\
MGIE~\cite{fu2024guiding} 
& 0.1639 & 0.0777 & 0.8205 & 0.6723 & 0.2787 
& 0.1592 & 0.0750 & 0.8090 & 0.6519 & 0.2637 \\
SmartEdit~\cite{huang2023smartedit} 
& 0.1295 & 0.0573 & 0.8630 & 0.7516 & 0.2971 
& 0.1111 & 0.0495 & 0.8739 & 0.7726 & 0.2824 \\
FoI~\cite{guo2023focus} 
& \underline{0.1054} & \underline{0.0385} & \underline{0.8811} & \underline{0.8096} & 0.2941
& \underline{0.0891} & \underline{0.0284} & \underline{0.8895} & \underline{0.8190} & \underline{0.2888} \\

\name \textbf{(ours)} & \textbf{0.0945} & \textbf{0.0366} & \textbf{0.8980} & \textbf{0.8392} & \textbf{0.2984} 
& \textbf{0.0661} & \textbf{0.0222} & \textbf{0.9296} & \textbf{0.8932} & \textbf{0.3006}
\\

\bottomrule
\end{tabular}
}

\vspace{-1.0em}
\label{tab:multi}
\end{table*}

\subsection{Evaluation Settings}
\label{sec:evalsettings}

\noindent \textbf{Datasets:} For evaluating single instruction tasks, we use the MagicBrush~\cite{zhang2024magicbrush} dataset, which covers both single-turn and multi-turn scenarios as detailed in \cref{ablation}, along with the EMU~\cite{sheynin2024emu} dataset. 
However, the literature lacks dedicated benchmark datasets for multi-instruction or context-aware instruction editing. To address this gap, we introduce two new tasks and curate corresponding datasets: \textit{Multi-instruction Image Editing} and \textit{Context-Aware Instruction Image Editing} as detailed in \cref{subsec:mainresults}. In Multi-instruction Image Editing, we concatenate applicable instructions from MagicBrush’s multi-turn dataset into a single instruction sequence. In the Context-Aware Instruction Image Editing task, we introduce non-applicable instructions generated with ChatGPT-4V(ision)~\cite{gpt4v} alongside images. More details on data creation are detailed in \cref{appendix:dataset}.

\noindent \textbf{Metrics:} To evaluate our proposed method, we employ a diverse set of metrics, including L1/L2, CLIP-I, DINO, CLIP-T, CLIP-dir, and PickScore~\cite{kirstain2023pick}. Detailed descriptions of these metrics are provided in \cref{appendix:metrics}.

\noindent \textbf{Baselines:} We compare \name with five different state-of-the-art image editing methods: IP2P~\cite{brooks2023instructpix2pix}, EMILIE~\cite{joseph2024iterative}, MGIE~\cite{fu2024guiding}, SmartEdit (SE)~\cite{huang2023smartedit}, and FoI~\cite{guo2023focus}.

\begin{figure*}[t]
    \renewcommand{\arraystretch}{0.3}
    \setlength\tabcolsep{2.5pt}
    \centering
    \resizebox{0.75\linewidth}{!}{%
    \normalsize
    \begin{tabular}{ccccccc}
        Input Image & \shortstack{IP2P~\cite{brooks2023instructpix2pix}} & \shortstack{MGIE~\cite{fu2024guiding}} & \shortstack{SmartEdit~\cite{huang2023smartedit}} & \shortstack{FoI~\cite{guo2023focus}} & \multicolumn{2}{c}{\name~\textbf{(Ours)}} \\

        \includegraphics[width=0.16\textwidth]{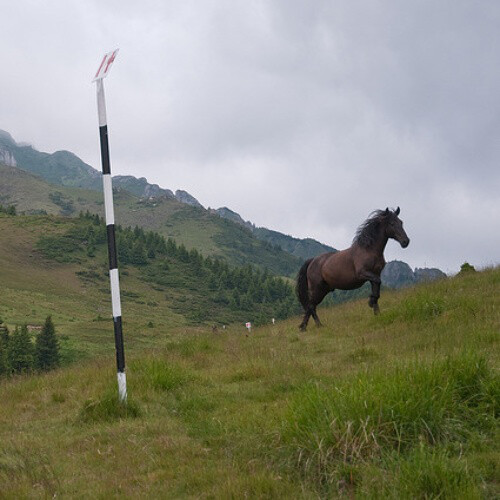}&
        \includegraphics[width=0.16\textwidth]{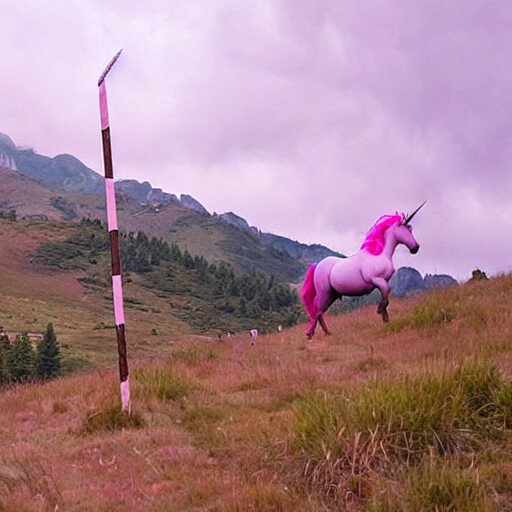}&
        \includegraphics[width=0.16\textwidth]{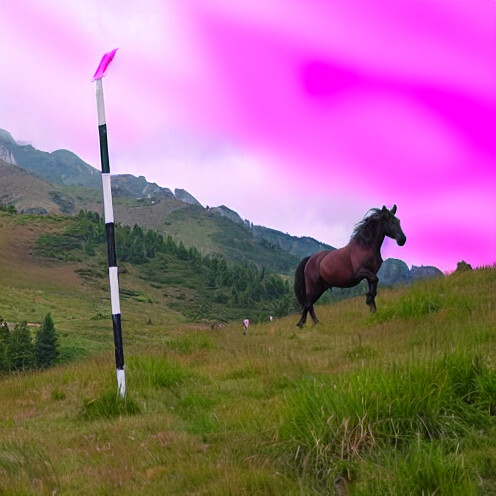}&
        \includegraphics[width=0.16\textwidth]{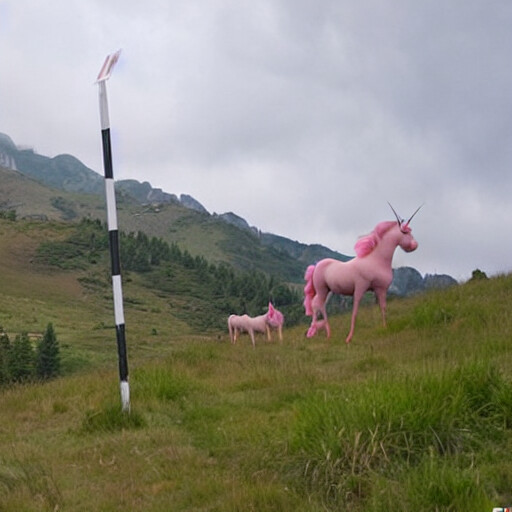}&
        \includegraphics[width=0.16\textwidth]{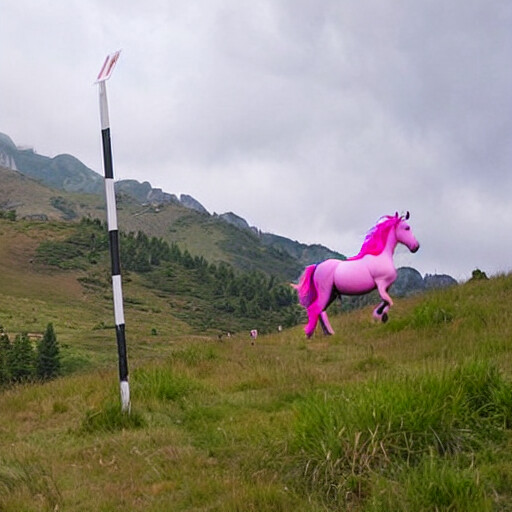}&
        \includegraphics[width=0.16\textwidth]{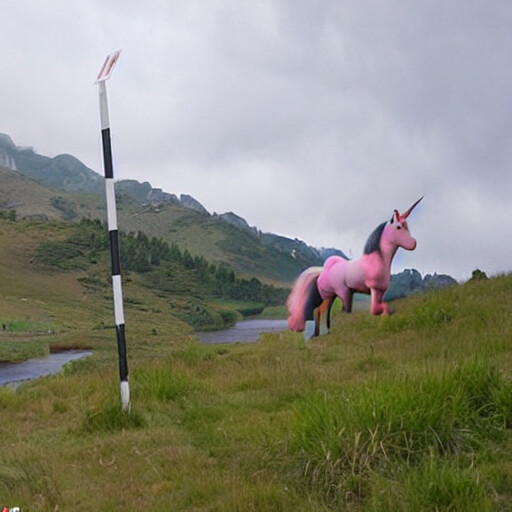}&
        \shortstack{
        \includegraphics[width=0.053\textwidth]{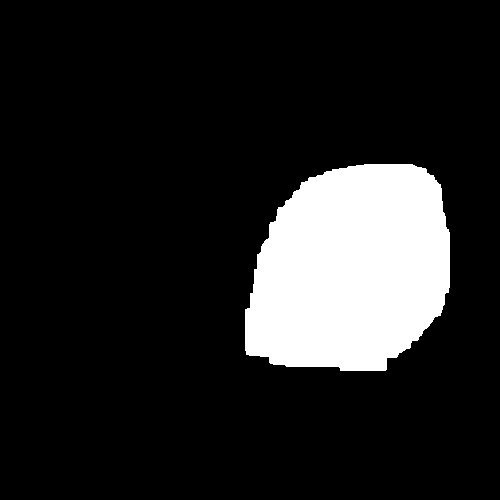} \\
        \texttt{[MASK]} \\
        \includegraphics[width=0.053\textwidth]{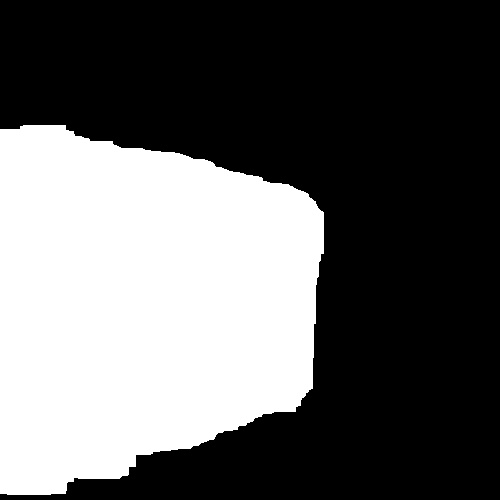} \\
        \texttt{[MASK]}} \\
        \multicolumn{7}{c}{(a) Edit instruction: \emph{``Turn the brown \textbf{horse} into a pink unicorn, and put a \textbf{river} nearby.''}} \\

        \includegraphics[width=0.16\textwidth]{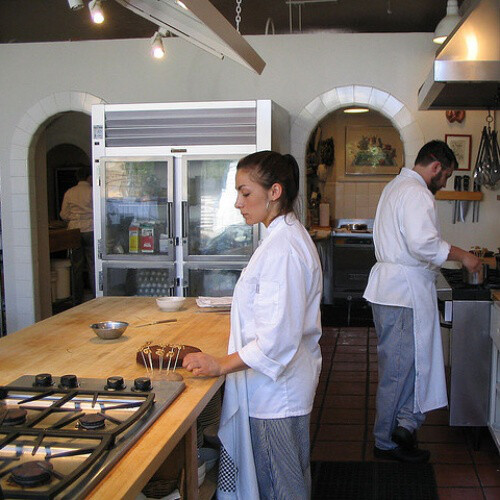}&
        \includegraphics[width=0.16\textwidth]{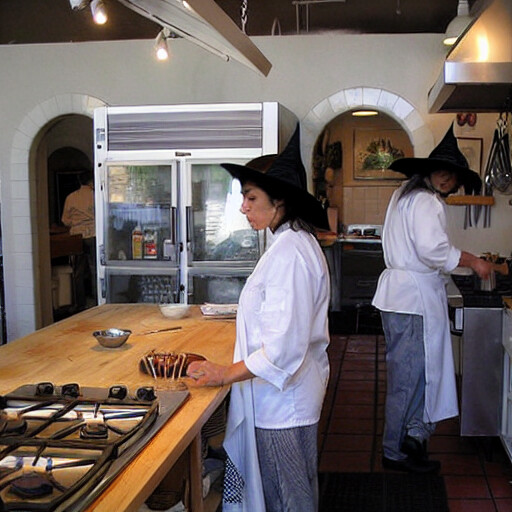}&
        \includegraphics[width=0.16\textwidth]{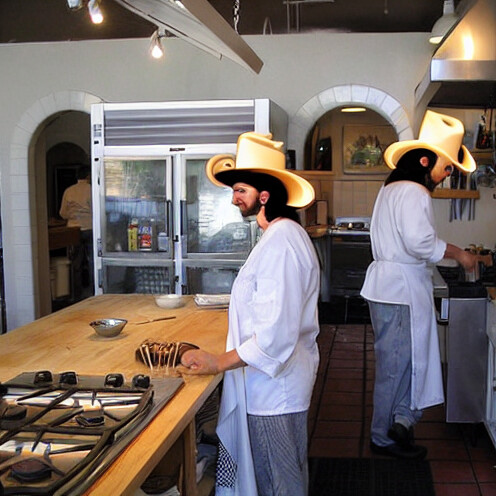}&
        \includegraphics[width=0.16\textwidth]{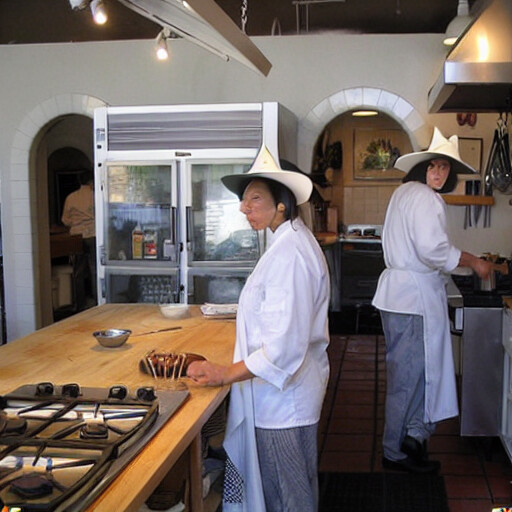}&
        \includegraphics[width=0.16\textwidth]{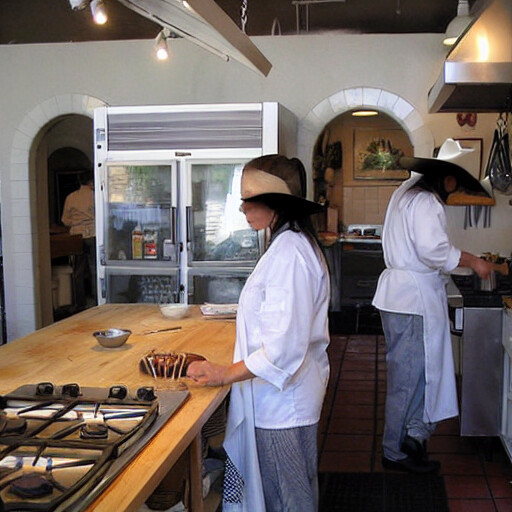}&
        \includegraphics[width=0.16\textwidth]{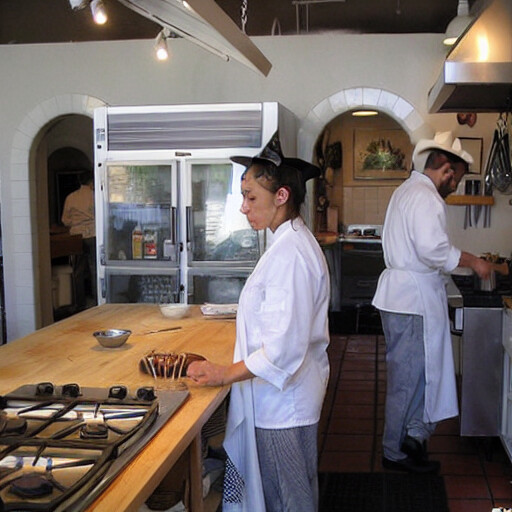}&
        \shortstack{
        \includegraphics[width=0.053\textwidth]{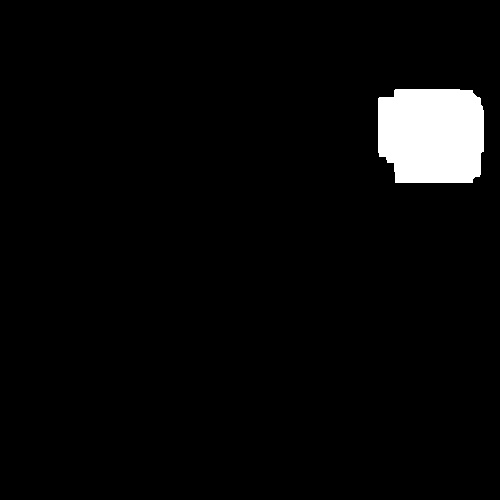} \\
        \texttt{[MASK]} \\
        \includegraphics[width=0.053\textwidth]{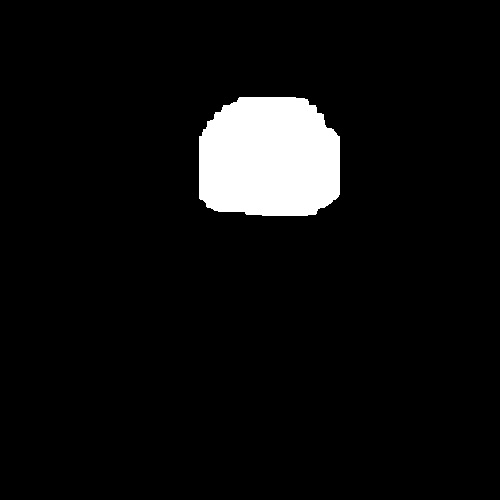} \\
        \texttt{[MASK]}} \\
        \multicolumn{7}{c}{(b) Edit instruction: \emph{``Give the \textbf{man} a cowboy hat, and put a witch hat in the \textbf{woman}.''}} \\

        \includegraphics[width=0.16\textwidth]{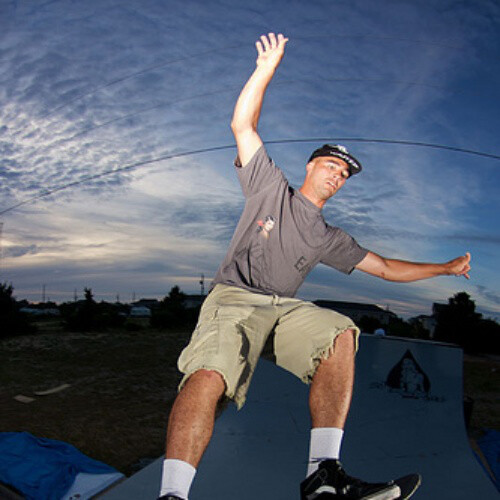}&
        \includegraphics[width=0.16\textwidth]{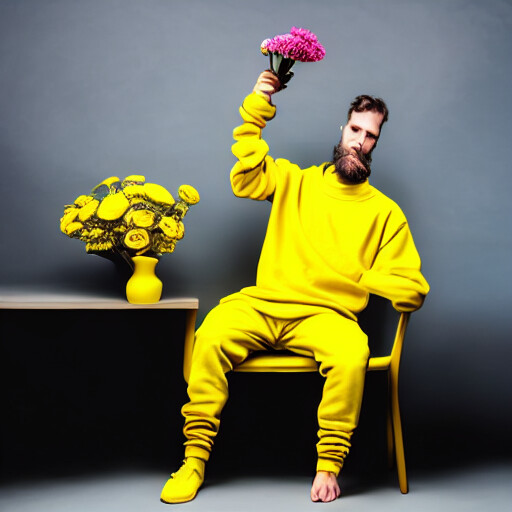}&
        \includegraphics[width=0.16\textwidth]{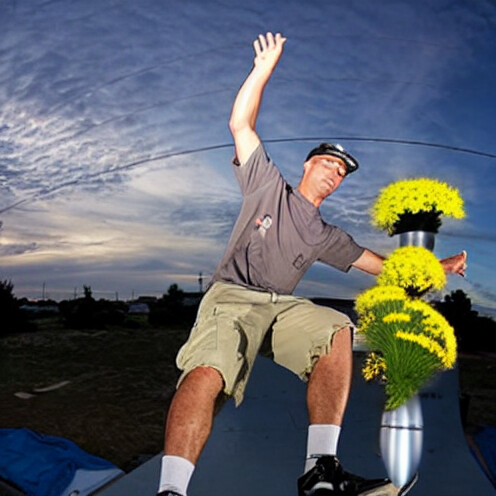}&
        \includegraphics[width=0.16\textwidth]{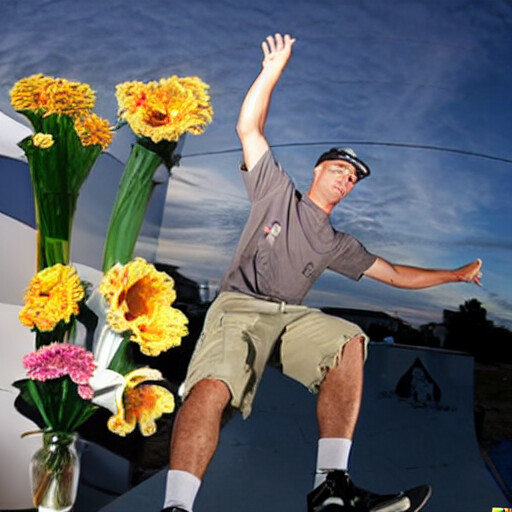}&
        \includegraphics[width=0.16\textwidth]{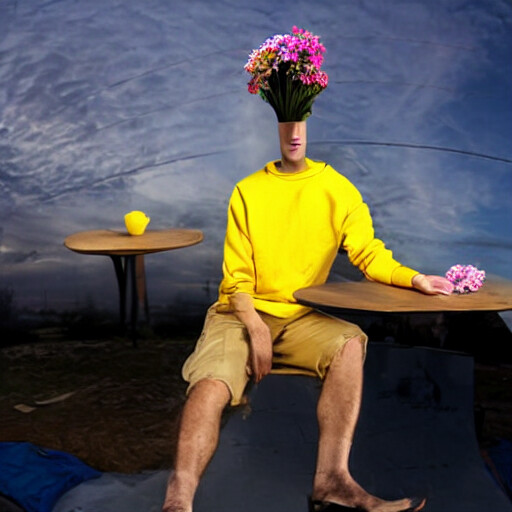}&
        \includegraphics[width=0.16\textwidth]{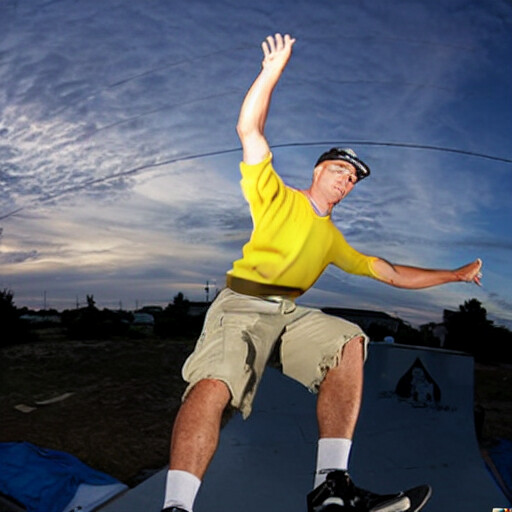}&
        \shortstack{
        \includegraphics[width=0.053\textwidth]{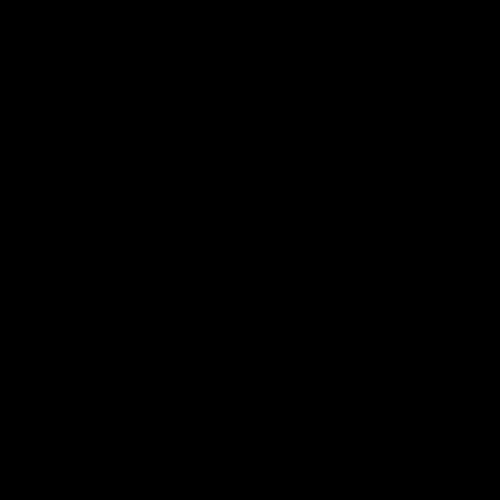} \\
        \texttt{[NEG]} \\
        \includegraphics[width=0.053\textwidth]{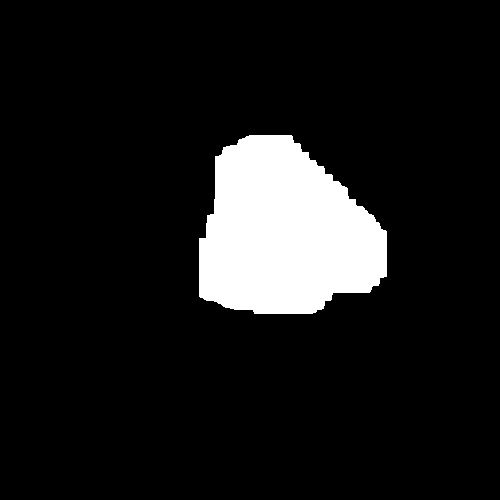} \\
        \texttt{[MASK]}} \\
        \multicolumn{7}{c}{(c) Edit instruction: \emph{``Put a flower \textbf{vase} on the table next to the chair.}} \\
        \multicolumn{7}{c}{\emph{On top of that, have the \textbf{man} be wearing a yellow sweatshirt.''}} \\
        
        \includegraphics[width=0.16\textwidth]{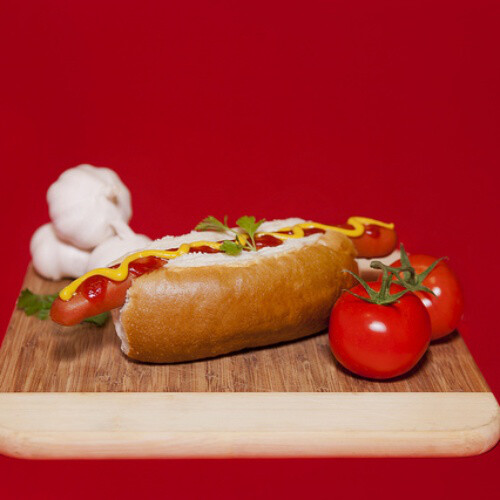}&
        \includegraphics[width=0.16\textwidth]{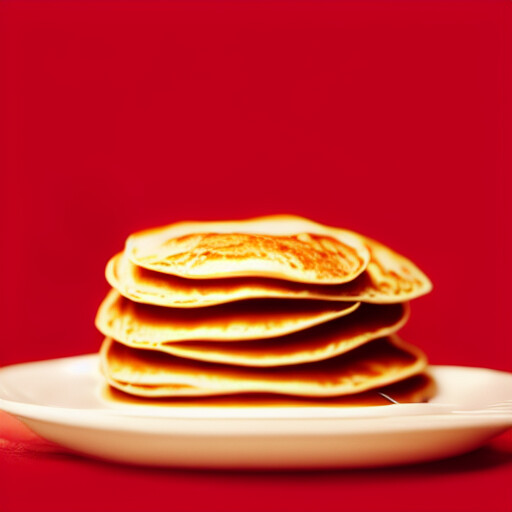}&
        \includegraphics[width=0.16\textwidth]{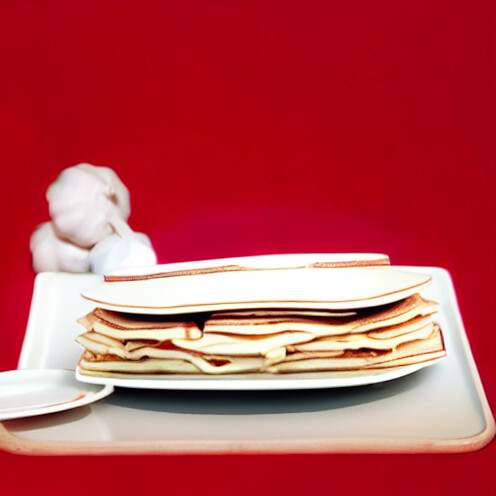}&
        \includegraphics[width=0.16\textwidth]{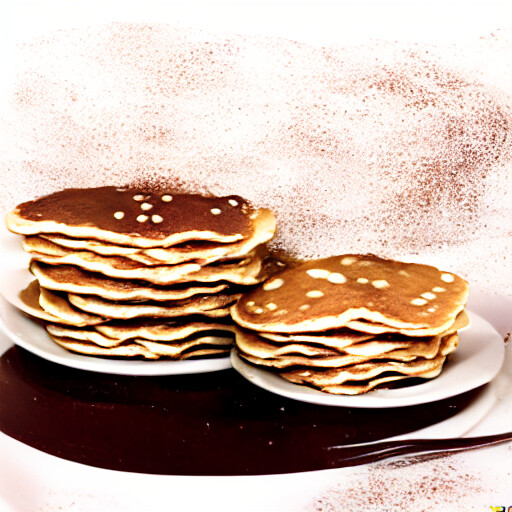}&
        \includegraphics[width=0.16\textwidth]{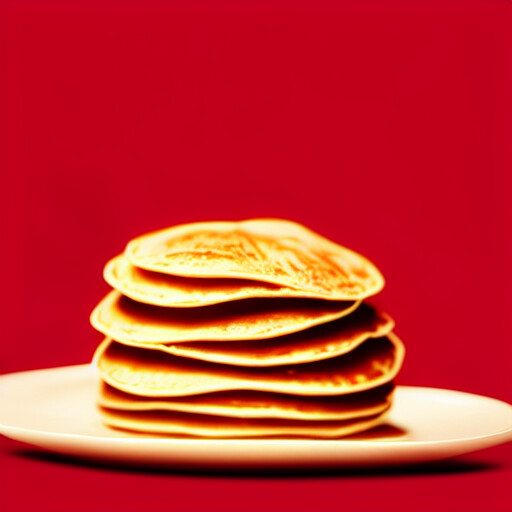}&
        \includegraphics[width=0.16\textwidth]{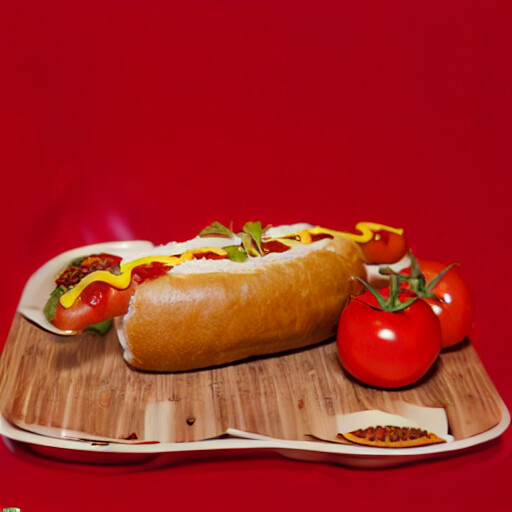}&
        \shortstack{
        \includegraphics[width=0.053\textwidth]{figure/negmask.jpg} \\
        \texttt{[NEG]} \\
        \includegraphics[width=0.053\textwidth]{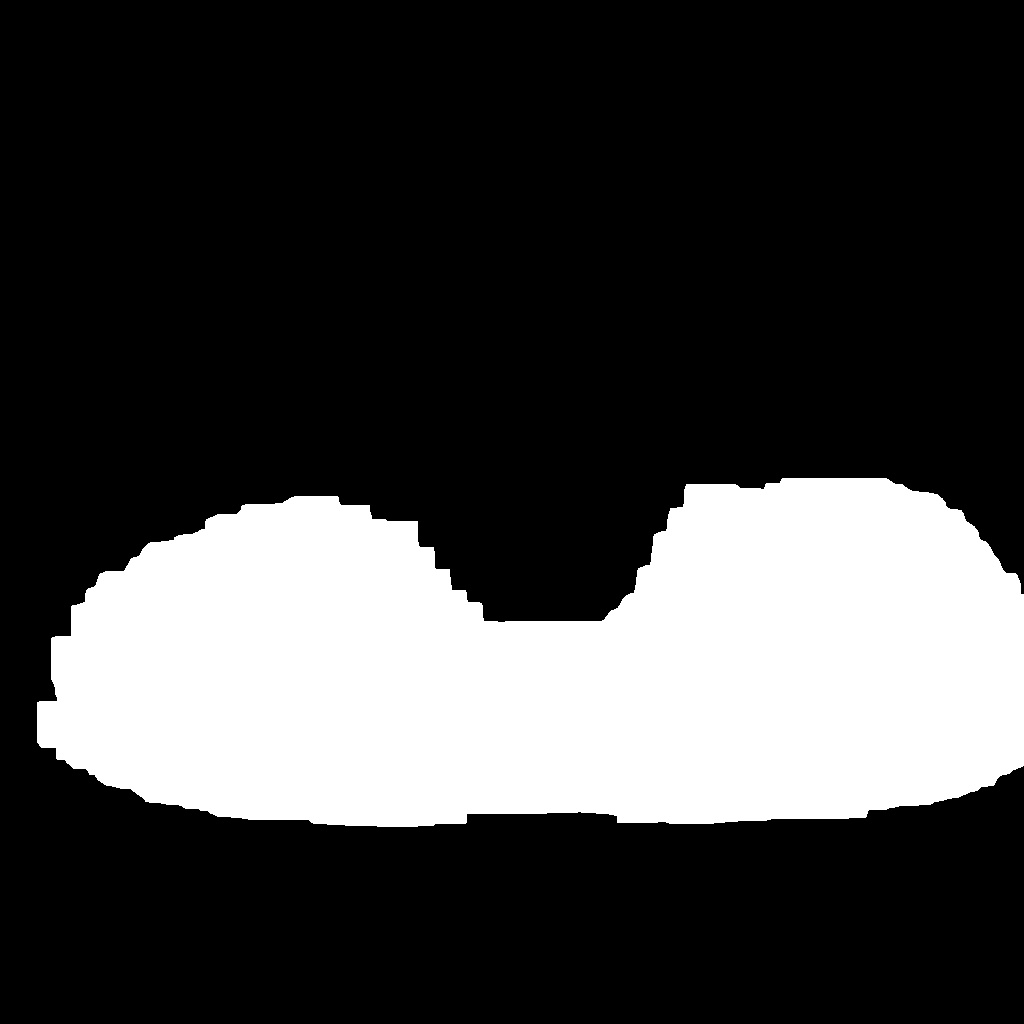} \\
        \texttt{[MASK]}} \\
        \multicolumn{7}{c}{(d) Edit instruction: \emph{``Remove the stack of \textbf{pancakes} from the plate and put the \textbf{food} on a plate.''}} \\

    \end{tabular}}
    
    \caption{\textbf{Qualitative comparisons:} FoI needs to extract keywords from each instruction using pretrained GPT model before running the model. Furthermore, due to inaccuracies in the attention map of diffusion model, FoI often fails to make precise modifications. In the case of context-aware instructions, \name accurately identifies applicable instructions by generating \texttt{[MASK]} and \texttt{[NEG]} tokens from MLLM. We present the decoded mask results for each instruction of the \texttt{[MASK]} token.}
    \label{fig:qualitative_main}
    \vspace{-1.75em}
    
\end{figure*}

\subsection{Main Results}
\label{subsec:mainresults}

\noindent\textbf{Quantitative Result.} As illustrated in \cref{tab:multi}, \name demonstrates state-of-the-art results across both multi-instruction and context-aware instruction tasks, particularly excelling in metrics such as CLIP-I, CLIP-T, and DINO similarity, and overall distance metrics (L1 and L2). This indicates that our model aligns closely with human perception in maintaining fidelity to edited images.

The existing methods exhibit notable limitations. MGIE, which relies on a summarization approach to compress instructions, proves to be vulnerable to non-applicable instructions, leading to potential inaccuracies in execution. While SmartEdit shows improved understanding due to the integration of MLLMs, it suffers from a lack of robustness by feeding all instructions into the diffusion model simultaneously, which can lead to oversights in handling complex editing requests. Additionally, FoI struggles with imprecise attention maps, which reduces its performance below \name though it is the most competitive baseline. In stark contrast, our approach effectively manages multi-instruction editing tasks, demonstrating superior capability in processing context-aware instructions. This proficiency enables our model to execute edits with high precision, aligning closely with the intended modifications while minimizing the risk of over-editing. Overall, our results underscore the advantages of our method in navigating the complexities inherent to multi-instruction tasks.

In addition to distance-based metrics and similarity-based metrics, we further evaluate human perceptual alignment using the PickScore metric across two editing tasks: Multi-instruction and Context-Aware image editing. \name outperforms the strongest baseline, FoI, by 18.1\% and 24.0\% in each setting, respectively. The advantage is more evident in the Context-Aware setting, demonstrating our model’s ability to effectively filter non-applicable instructions. More detailed results are presented in \cref{sec:pick}.

\noindent\textbf{Qualitative Result.} As shown in \cref{fig:qualitative_main}, we present qualitative results and observe the following: All models, except for FoI, frequently execute only a single instruction when multiple instructions are provided. In (a), while FoI successfully performs the first instruction, it fails to generate an accurate attention map for the keyword `river', resulting in the incomplete application of the second instruction. Similarly, in (b), the lack of a fine-grained attention map results in the hat being placed incorrectly. The remaining models predominantly execute only one instruction and demonstrate a tendency toward over-editing; for instance, IP2P and MGIE alter the background color in (a), and SmartEdit generates an additional unicorn.

In (c) and (d), most models exhibit erroneous edits in response to non-applicable instructions. In (c), despite the absence of a chair or table in the input image, the models add incorrect floral elements or a table. Similarly, in (d), although the input image does not contain a pancake, it erroneously appears due to the removal instruction, illustrating an inability to correctly handle the instruction. Especially compared to FoI, our model demonstrates greater precision in mask extraction for areas requiring modification, enabling more refined edits. Furthermore, through the use of \texttt{[MASK]} and \texttt{[NEG]} tokens, our proposed model facilitates robust, context-aware image editing. Further qualitative results are provided in \cref{appendix:results}.
\section{Ablation Study}
\label{ablation}

\noindent\textbf{Robustness of \name.}
In our framework, distinguishing applicable from non-applicable instructions is critical to prevent unintended edits. Each input may contain multiple instructions, which are classified by the MLLM into \texttt{[MASK]} and \texttt{[NEG]} tokens. On the Context-Aware Image Editing dataset, our model achieves a token classification accuracy of 90.21\%, highlighting the robustness of \name in filtering non-applicable instructions. Furthermore, we evaluate the alignment of generated masks with ground-truth edited regions using standard segmentation metrics. For applicable instructions, classified as \texttt{[MASK]} token, our model achieves an IoU of 0.3819 and a Dice score of 0.4986. As described in \cref{losseq}, our model is trained with multiple loss objectives, not solely for segmentation accuracy. The generated masks are designed as high-level guidance, reflecting our focus on instruction fidelity and plausibility rather than strict spatial matching.

\noindent\textbf{Evaluation on Instruction-Following Accuracy.}
We evaluate our method on the EMU dataset for both single-instruction and context-aware instruction tasks. Since the EMU dataset does not provide ground-truth target images, we utilize CLIP-T and CLIP-dir as evaluation metric. CLIP-dir measures how accurately the generated image aligns with the intended semantic direction of the instructions. As shown in \cref{tab:emu}, \name achieves the highest CLIP-dir score in the context-aware instruction task, demonstrating its effectiveness in handling non-executable instructions.

\begin{table}[t!]
\small

\begin{minipage}[t]{0.39\linewidth}
\caption{\textbf{Quantitative comparison on EMU dataset.} Achieving the highest CLIP-dir score in the Context-Aware task shows that our model effectively distinguishes non-executable instructions.}
\centering
\setlength{\tabcolsep}{3pt}
\resizebox{\linewidth}{!}{%
\begin{tabular}{ccc|cc}
\toprule
Task & \multicolumn{2}{c}{Single-inst} & \multicolumn{2}{c}{Context-Aware}\\
Method & CLIP-T$\uparrow$ & CLIP-dir$\uparrow$ & CLIP-T$\uparrow$ & CLIP-dir$\uparrow$\\ \midrule
IP2P & 0.2616 & 0.075  & 0.2446 & 0.064  \\
MGIE & 0.2680 &  0.082 & 0.2543 & 0.066  \\
SE & \underline{0.2680} &  \textbf{0.094} & 0.2448 &  \underline{0.067} \\
FoI & 0.2673 & 0.068 & \underline{0.2651} & 0.054 \\
\textbf{ours} & \textbf{0.2687} & \underline{0.092}  & \textbf{0.2679} & \textbf{0.092}  \\
\bottomrule
\end{tabular}}
\label{tab:emu}
\end{minipage}
\quad
\begin{minipage}[t]{0.59\linewidth}
\caption{\textbf{Comparison of results before and after additional training with the surrogate module.} We apply the surrogate module to improve the CLIP-T score, which also enhances L1/L2 losses, as well as CLIP-I and DINO scores.}
       \centering
\setlength{\tabcolsep}{3pt}
\resizebox{\linewidth}{!}{
\begin{tabular}{c|c|c|ccccc}
\toprule
\multicolumn{2}{c|}{Task}& Config & L1$\downarrow$ & L2$\downarrow$ & CLIP-I$\uparrow$ & DINO$\uparrow$ & CLIP-T$\uparrow$ \\ \midrule

\multirow{4}{*}{\shortstack{Single \\ Inst.}}&\multirow{2}{*}{\shortstack{Single \\ -Turn}} & before  & 0.0602 & 0.0194 & 0.9367 & 0.9067 & 0.3020 \\
&& after  & 0.0596 & 0.0191 & 0.9375 & 0.9069 & 0.3022 \\
\cline{2-8}
&\multirow{2}{*}{\shortstack{Multi \\ -Turn}} & before  & 0.0931 & 0.0339 & 0.8969 & 0.8357 & 0.3011 \\
&& after  & 0.0782 & 0.0268 & 0.9127 & 0.8659 & 0.3019 \\
\midrule
\multirow{4}{*}{\shortstack{Multi \\ Inst.}}&\multirow{2}{*}{Multi} & before  & 0.0957 & 0.0372 & 0.8961 & 0.8329 & 0.2975 \\
&& after  & 0.0945 & 0.0366 & 0.8980 & 0.8392 & 0.2984 \\
\cline{2-8}
&\multirow{2}{*}{\shortstack{Context \\ -Aware}} & before  & 0.0673 & 0.0228 & 0.9284 & 0.8910 & 0.3002 \\
&& after  & 0.0661 & 0.0222 & 0.9296 & 0.8932 & 0.3006 \\
\bottomrule

\end{tabular}}
\label{tab:surrogate}
\end{minipage}

\vspace{-1.5em}
\end{table}

\begin{wrapfigure}{r}{0.5\textwidth}
\vspace{-1.0em}
\captionof{table}{\textbf{Quantitative comparison on single instruction tasks.} 
\name excels in single instruction tasks by generating precise masks that accurately target modification areas.}

\renewcommand{\arraystretch}{0.8}
\setlength{\tabcolsep}{1.5pt}

\vspace{-0.25em}
\resizebox{\linewidth}{!}{
\tiny
\begin{tabular}{ccccccc}
\toprule
Task& Method & L1$\downarrow$ & L2$\downarrow$ & CLIP-I$\uparrow$ & DINO$\uparrow$ & CLIP-T$\uparrow$ \\ \midrule
\multirow{5}{*}{\rotatebox{90}{Single-turn}} &
IP2P & 0.1129 & 0.0373 & 0.8540 & 0.7423 & 0.2918 \\
&MGIE & 0.0931 & 0.0383 & 0.8853 & 0.8088 & 0.2935 \\
&SE& 0.0895 & 0.0353 & 0.9030 & 0.8308 & \textbf{0.3024} \\
&FoI & \underline{0.0699} & \underline{0.0206} & \underline{0.9207} & \underline{0.8779} & 0.2980 \\
&\textbf{ours} & \textbf{0.0596} & \textbf{0.0191} & \textbf{0.9375} & \textbf{0.9069} & \underline{0.3022} \\
\midrule

\multirow{6}{*}{\rotatebox{90}{Multi-turn}} &
IP2P & 0.1538 & 0.0575 & 0.8103 & 0.6511 & 0.2866 \\
&EMILIE& 0.1268 & 0.0509 & 0.8557 & 0.7591 & 0.2916\\
&MGIE & 0.1312 & 0.0574 & 0.8571 & 0.7507 & 0.3013 \\
&SE & 0.1333 & 0.0575 & 0.8567 & 0.7421 & \textbf{0.3021} \\
&FoI & \underline{0.1084} & \underline{0.0379} & \underline{0.8681} & \underline{0.7838} & 0.2935 \\
&\textbf{ours} & \textbf{0.0782} & \textbf{0.0268} & \textbf{0.9127} & \textbf{0.8659} & \underline{0.3019} \\
\bottomrule
\label{tab:single}
\end{tabular}
}

\renewcommand{\arraystretch}{0.3}
\setlength\tabcolsep{1.5pt}
\resizebox{\linewidth}{!}{%
    \begin{tabular}{cccccc}
        Input & IP2P & MGIE & SmartEdit & FoI & \textbf{\name} \\
        \\
        \includegraphics[width=0.13\textwidth]{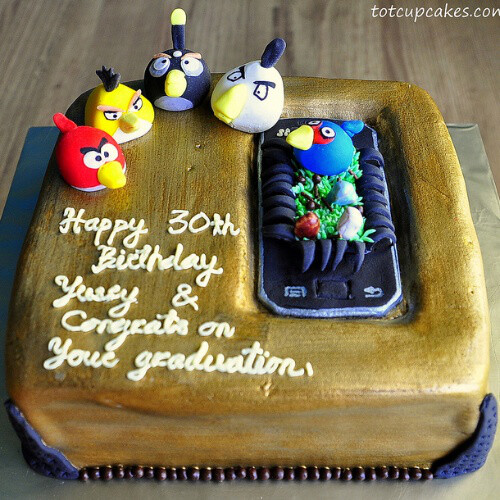}&
        \includegraphics[width=0.13\textwidth]{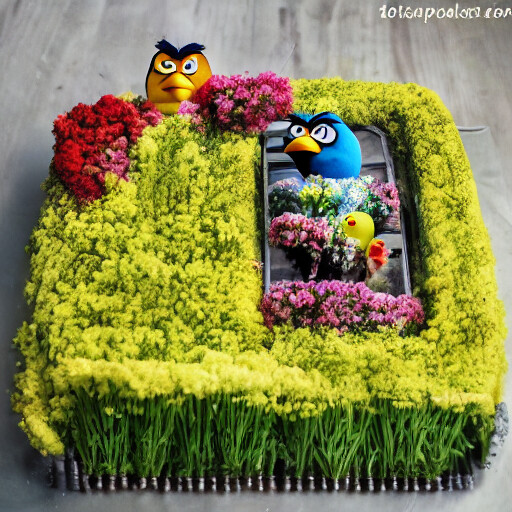}&
        \includegraphics[width=0.13\textwidth]{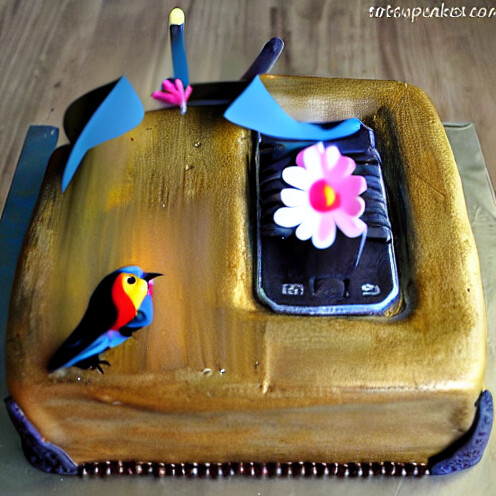}&
        \includegraphics[width=0.13\textwidth]{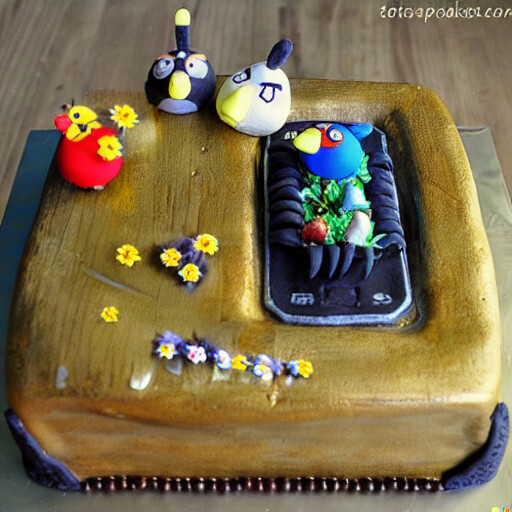}&
        \includegraphics[width=0.13\textwidth]{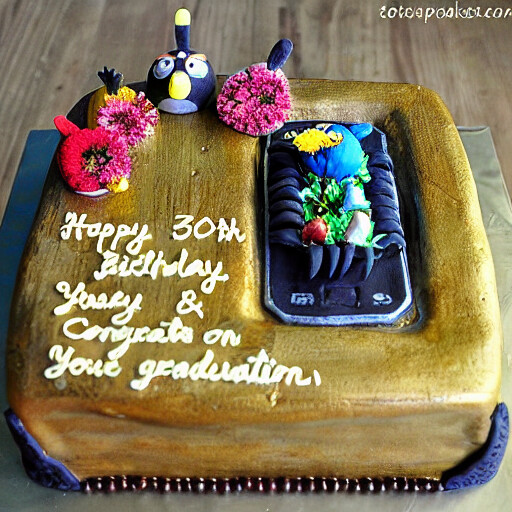}&
        \includegraphics[width=0.13\textwidth]{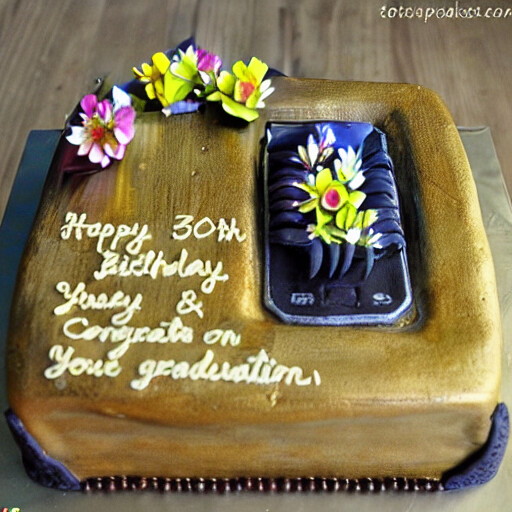}
        \\
        \\
        \multicolumn{6}{c}{(a) Edit instruction: \emph{``Replace the angry \textbf{birds} with flowers''}}
        \\
        \\
        \includegraphics[width=0.13\textwidth]{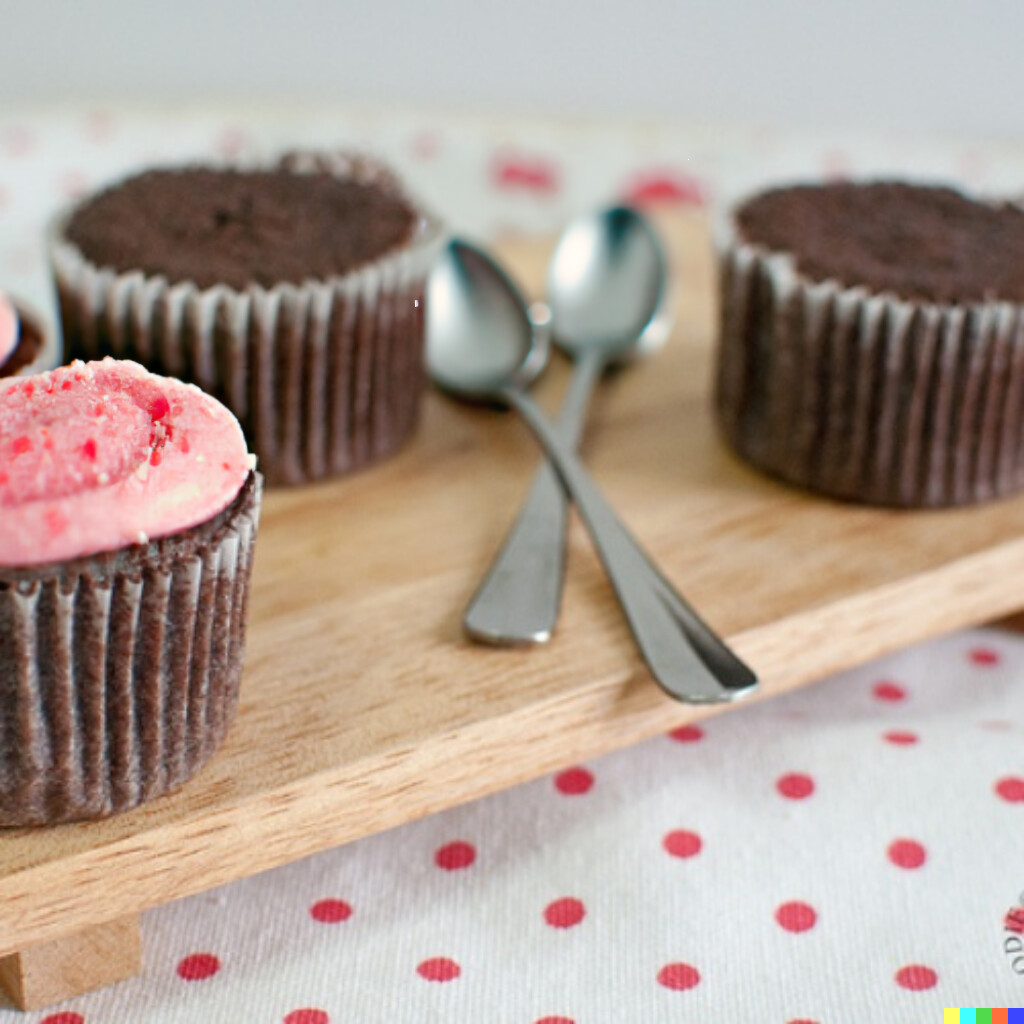}&
        \includegraphics[width=0.13\textwidth]{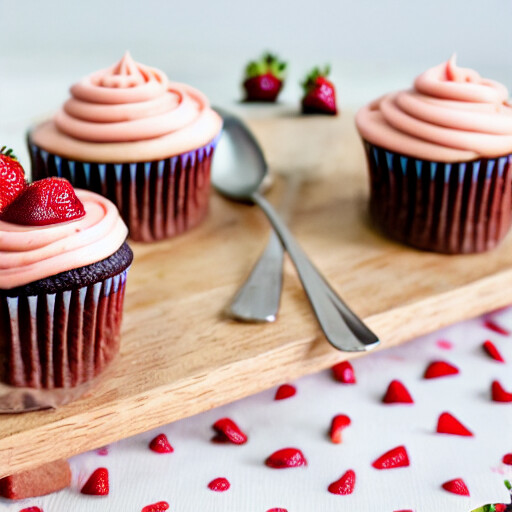}&
        \includegraphics[width=0.13\textwidth]{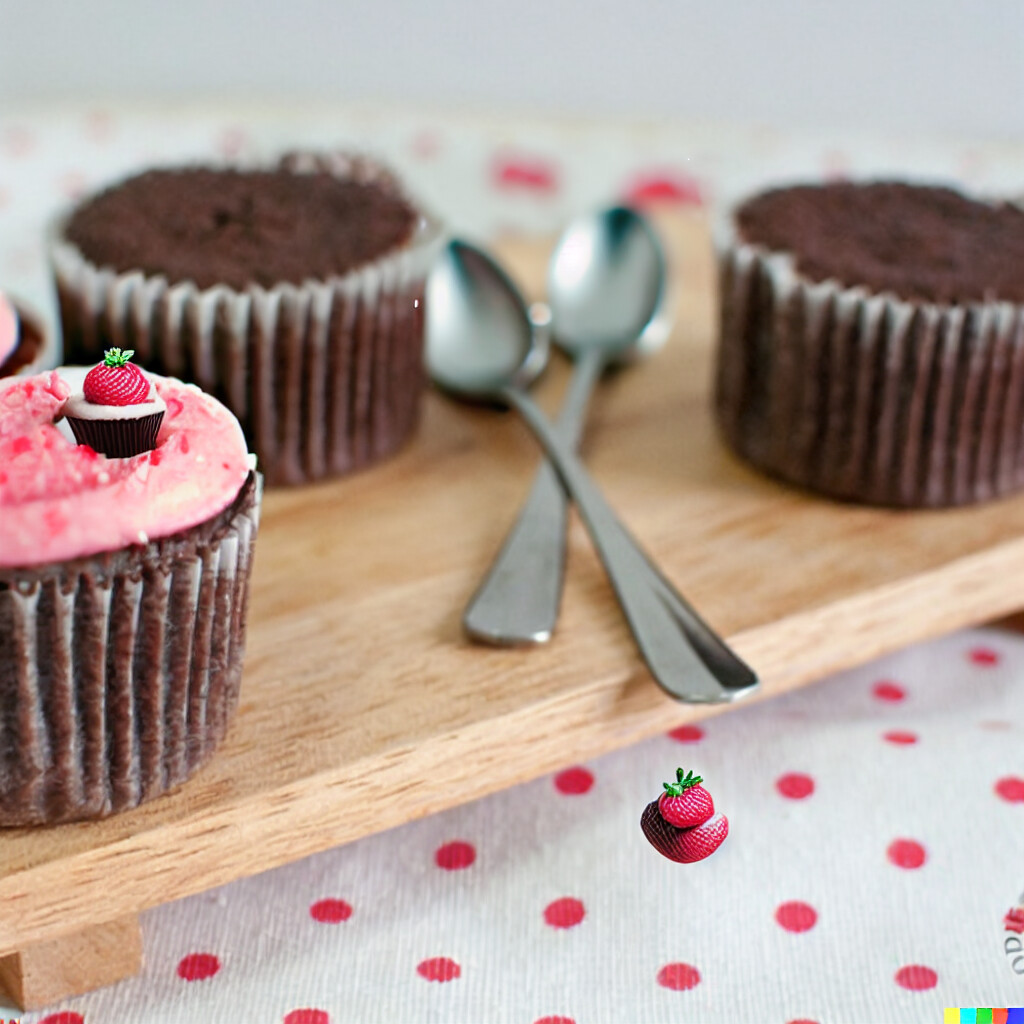}&
        \includegraphics[width=0.13\textwidth]{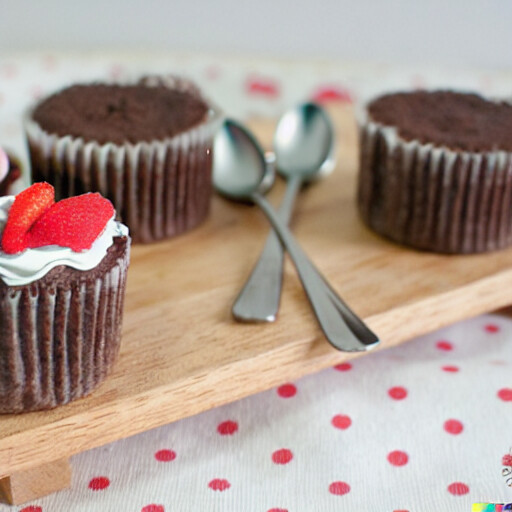}&
        \includegraphics[width=0.13\textwidth]{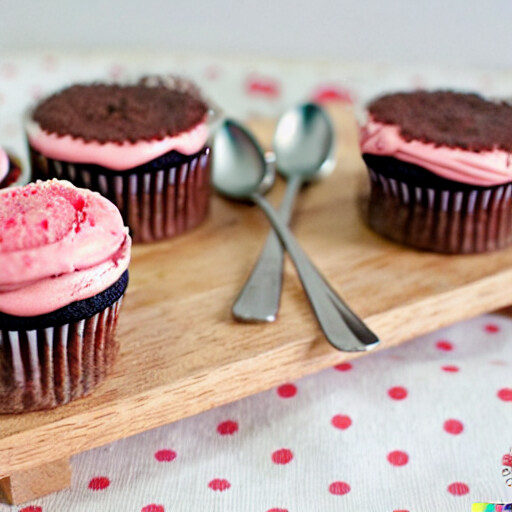}&
        \includegraphics[width=0.13\textwidth]{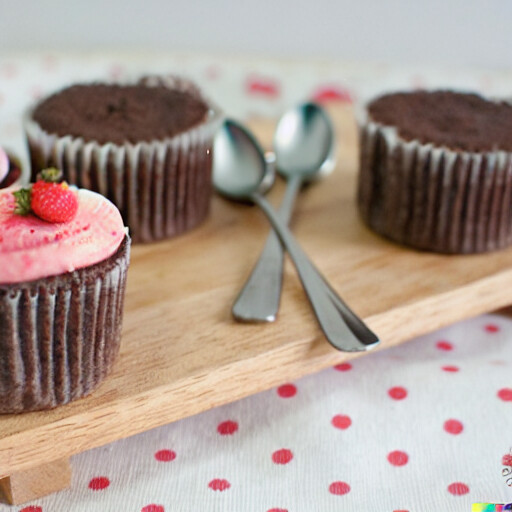}
        \\
        \\
        \multicolumn{6}{c}{(b) \emph{``Add a strawberry on top of the \textbf{cupcake} with frosting''}} \\
        \\

    \end{tabular}}
    \caption{\textbf{Qualitative comparisons for single instruction task.} \name demonstrates successful editing even in the single instruction task.}
    \label{fig:qualitative_single}

\vspace{-2.5em}
\end{wrapfigure}

\noindent\textbf{Single-Instruction Task Performance.} 
\name, optimized for multi-instruction tasks, also performs strongly on single-instruction tasks, as shown in \cref{tab:single}. \name achieves strong performance across most evaluation metrics.
In contrast, SmartEdit shows a tendency toward over-editing. This reflects \name's balanced approach, minimizing over-editing while maintaining high fidelity.
As shown in \cref{fig:qualitative_single}, most models exhibit over-editing issues. Although FoI is designed to minimize over-editing, it encounters specific issues as follows: 
inaccurate attention maps in (a) prevent precise modifications to the `angry birds' object, while in (b), additional frosting is incorrectly applied to cupcakes. These cases show that \name achieves accurate edits without over-editing.

\noindent\textbf{Impact of Surrogate Module Training.} We compare the results on the multi-instruction tasks and single-instruction tasks before and after additional surrogate module training. As shown in \cref{tab:surrogate}, we demonstrate that mask generation through this surrogate module improves model performance. Interestingly, the improvement is not just for CLIP-T but also for the other metrics.

Additional ablation studies on the variation of the Token Decoder and the inference time comparison are provided in \cref{appendix:token} and \cref{appendix:time}, respectively.

\section{Conclusion}

In this paper, we addressed the limitations of current text-guided image editing models, particularly their difficulty in handling fine-grained edits, multi-instruction edits, and distinguishing between executable and non-executable instructions. Leveraging MLLMs, we generated specialized tokens (\texttt{[MASK]} and \texttt{[NEG]}) and designed token broadcaster to ensure the validity of the contextual coherence between instructions and the image, so that only relevant edits can be applied to the designated regions while ignoring non-executable instructions. For comprehensive evaluation, we created new datasets that can evaluate Context-Aware Image Editing task, where our approach achieves superior results across both qualitative and quantitative evaluations compared to state-of-the-art solutions.

\bibliographystyle{plain} 
\bibliography{ref}

\clearpage
\appendix
\onecolumn

\section{Implementation Details}
\label{appendix:implementation}

In this section, we provide more details about the training process. For the main training phase, we set all loss function weights (\(\lambda_1\) to \(\lambda_4\)) to 1. The AdamW optimizer was used with a learning rate of 0.0003 and no weight decay. Training was conducted for 400,000 iterations, with LLaVA-7B~\cite{liu2024visual} employed as the Multimodal Large Language Model (\(\mathcal{F}\)) in our architecture. The main training utilized 2 Nvidia A100 80GB GPUs for approximately three days, with a batch size of 16.

In the additional training phase, the surrogate module was trained using the AdamW optimizer with a learning rate of 0.0003. This phase involved 1,000 iterations of training.

Following the completion of the surrogate module training, we resumed fine-tuning the MLLM, Token Broadcaster, and Token Decoder components by setting the weight of the MSE loss term (\(\lambda_5\)) to 10. During this phase, we employed the AdamW optimizer with a decreased learning rate of 0.0001 and conducted training for an additional 400,000 iterations.

\section{Metrics Details}
\label{appendix:metrics}

We utilize a diverse set of metrics including L1/L2, CLIP-I, DINO, CLIP-T, CLIP-dir, and PickScore.
L1 and L2 are used to calculate the mean absolute pixel-wise discrepancy between the generated and ground truth images. Beyond pixel-level evaluation, we use CLIP-I~\cite{radford2021learning} and DINO~\cite{oquab2023dinov2}, both of which assess image quality via cosine similarity between embeddings of the edited and ground truth images. CLIP-T evaluates the alignment between the generated image and the goal image global description. Additionally, we employ CLIP-dir, which measures the agreement between changes in images and changes in captions. Lastly, PickScore~\cite{kirstain2023pick} serves as a proxy for human preference by quantifying how likely a generated image is to be favored by humans.

\section{Dataset Details}
\label{appendix:dataset}

\subsection{Dataset Design}
\label{appendix:datasetdesign}

The MagicBrush dataset~\cite{zhang2024magicbrush} is specifically designed to support single-instruction image editing tasks, encompassing both single-turn and multi-turn scenarios. Single-turn tasks involve a single instruction for an image edit, while multi-turn tasks consist of sequential single-instruction edits, such as two-turn and three-turn edits. Two-turn edits are composed of two sequential single-instruction edits, and three-turn edits consist of three sequential single-instruction edits. Using the multi-turn data from the MagicBrush dataset, specifically the two-turn and three-turn edits, we restructured them into multi-instruction tasks. By combining the individual instructions into cohesive multi-instruction prompts, we created more complex editing scenarios that simulate real-world use cases. This restructuring involved generating connecting phrases to logically link the instructions, enabling seamless transitions between steps in the multi-instruction tasks. In addition to MagicBrush dataset, we incorporate the EMU dataset~\cite{sheynin2024emu}, which is a single-turn image editing dataset.

For Context-Aware Instruction Image Editing task, we utilized ChatGPT-4V~\cite{gpt4v} to generate non-applicable instructions. For each input image, we obtained five randomly generated non-applicable instructions from the model. During training, one of these instructions was randomly selected and incorporated into the multi-instruction tasks, enhancing the model's robustness in distinguishing between applicable and non-applicable instructions. During testing, one of the five non-applicable instructions was fixed and consistently used for evaluation, ensuring a standardized assessment of the model's performance. The number of test dataset examples for each task is shown in \cref{tab:dataset}.
\begin{table}[h]
    \centering
    \caption{\textbf{Details on the number of image editing samples for each task in MagicBrush dataset.}}
    \setlength\tabcolsep{5.5pt}
    \small
    \begin{tabular}{r|cccc}
    \toprule
        Sessions with & Single-turn Inst. & Multi-turn Inst. & Multi Inst. & Context-Aware Inst. \\
        \midrule
        One Edit & 216 & 216 & - & -\\
        Two Edits & 240 & 120 & 120 & 1053\\
        Three Edits& 597 & 199 & 597 & 1571\\
        \midrule
        Total  & 1053 & 535 & 717 & 2624\\
        \bottomrule
    \end{tabular}
    \label{tab:dataset}
\end{table}

\subsection{Dataset Biases}
To mitigate potential biases introduced by using ChatGPT-4V~\cite{gpt4v} for generating non-applicable instructions, we carefully designed the prompting process to align with the linguistic structure of the MagicBrush dataset. Specifically, we instructed the model to generate instructions that begin with common syntactic patterns found in MagicBrush dataset (e.g., "Put something", "Remove something", "Replace something", "Let there be", "Make something to something").

To validate the representational fidelity of the generated instructions, we categorized all instructions into four major types: \textit{Add object}, \textit{Replace object}, \textit{Remove object}, and \textit{Change something (action, color, etc.)}. The distribution of instruction categories in our generated non-applicable dataset (3,677 instances) was compared with the original MagicBrush's instructions. As shown in \cref{tab:instruction}, although there is a slight variation in category proportions, the overall instruction distribution was designed to reflect the original dataset’s structure and characteristics.
\begin{table}[h]
    \centering
    \caption{\textbf{Instruction distribution comparison between MagicBrush and our generated dataset.}}
    \setlength\tabcolsep{10.5pt}
    \small
    \begin{tabular}{c|cc}
    \toprule
        Instruction Category & MagicBrush & Generated Non-applicable Instructions \\
        \midrule
        Add object &	39.0\% &	34.3\% \\
        Replace object	&17.9\% &	20.5\% \\
        Remove object	&7.0\% &    21.1\% \\
        Change something (action, color, etc.)	&36.1\%	&24.1\% \\
        \bottomrule
    \end{tabular}
    \label{tab:instruction}
\end{table}
\begin{table}[t]
\centering
\small
\caption{\textbf{Comparison of results using different segmentation models.} Our Token Decoder demonstrates superior alignment in CLIP and DINO metrics compared to SAM segmentation models.}
\setlength{\tabcolsep}{2.5pt}
\begin{tabular}{c|c|c|ccccc}
    \toprule
    \multicolumn{2}{c|}{Task}& Segmentation Model & L1$\downarrow$ & L2$\downarrow$ & CLIP-I$\uparrow$ & DINO$\uparrow$ & CLIP-T$\uparrow$ \\ \midrule
    \multirow{4}{*}{\shortstack{Single \\ Inst.}}&\multirow{2}{*}{\shortstack{Single \\ -Turn}} & SAM & 0.0585&	0.0184&	0.9355&	0.9056&	0.2974  \\
    && \textbf{Ours}  & 0.0602 & 0.0194 & 0.9367 & 0.9067 & 0.3020  \\
    \cline{2-8}
    &\multirow{2}{*}{\shortstack{Multi \\ -Turn}} & SAM   & 0.0903	 & 0.0321 &	0.8932 &	0.8298	& 0.2917  \\
    && \textbf{Ours}    & 0.0931 & 0.0339 & 0.8969 & 0.8357 & 0.3011  \\
    \midrule
    \multirow{4}{*}{\shortstack{Multi \\ Inst.}}&\multirow{2}{*}{\shortstack{Multi}} & SAM & 0.0934& 0.0359 & 0.8946 & 0.8303& 0.2895  \\
    && \textbf{Ours}  & 0.0957 & 0.0372 & 0.8961 & 0.8329 & 0.2975   \\
    \cline{2-8}
    &\multirow{2}{*}{\shortstack{Context \\ -Aware}} & SAM & 0.0647 & 0.0215 & 0.9273 & 0.8905 & 0.2952  \\
    && \textbf{Ours}  & 0.0673 & 0.0228 & 0.9284 & 0.8910 & 0.3002  \\
    \bottomrule
\end{tabular}
\label{tab:variation}
\end{table}

\begin{table}[ht] 
\centering 
\small 
\caption{\textbf{Inference time comparison of MLLM part and diffusion model part.}}
\begin{tabular}{c|ccccc} 
    \toprule 
    \textbf{Method} & IP2P & MGIE & SmartEdit & FoI & \name (Ours) \\ 
    \midrule 
    MLLM part & – & 2.1s & 1.5s & – & \textbf{0.7s} \\ 
    Diffusion part & 7.2s & 3.4s & 4.0s & 9.1s & 8.5s \\ 
    \midrule 
    Total Inference Time & 7.2s & 5.5s & 5.5s & 9.1s & 9.2s \\ 
    \bottomrule 
\end{tabular} 
\label{time} 
\end{table}

\begin{table}[t]
\centering
\small
\caption{\textbf{PickScore comparison across baseline methods.}}
\begin{tabular}{lccccc}
\toprule
Method & IP2P & MGIE & SmartEdit & FoI & CAMILA (Ours) \\
\midrule
Multi-instruction & 0.1598 & 0.1404 & 0.1844 & 0.2363 & \textbf{0.2790} \\
Context-Aware & 0.1666 & 0.1350 & 0.1865 & 0.2285 & \textbf{0.2834} \\
\bottomrule
\end{tabular}
\label{tab:pickscore}
\end{table}

\section{Additional Ablation Study}
\label{appendix:ablation}

\subsection{Token Decoder Variations}
\label{appendix:token}

A straightforward approach might be to use pretrained segmentation models~\cite{kirillov2023segment, ke2024segment,rasheed2024glamm} to replace the Token Decoder for generating masks. Thus, we replaced our Token Decoder with SAM~\cite{kirillov2023segment}, and evaluated its performance within our framework, as shown in \cref{tab:variation}. However, SAM does not take edit instructions as input, lacking information about which parts of the image actually need editing. In contrast, our 2-layer transformer-based model utilizes edit instructions to produce targeted masks, resulting in superior CLIP and DINO scores, showing we can better preserve the intended meaning of edits.

\subsection{Inference Time Comparison}
\label{appendix:time}

We measure the inference time of each method and report the results in \cref{time}. All experiments are conducted using an NVIDIA A100 80GB GPU. We separately measure the time for MLLM-related modules and diffusion-based modules. \name achieves the fastest inference time for the MLLM module while maintaining comparable total latency. The slightly higher diffusion time in FoI and \name is attributed to attention modulation applied to the IP2P backbone. Unlike other baselines, \name supports single-shot, multi-instruction editing without requiring preprocessing steps such as keyword extraction.

\begin{figure*}[t]
    \setlength\tabcolsep{2.5pt}
    \centering
    \resizebox{0.8\linewidth}{!}{%
    \large
    \begin{tabular}{c |cc |cc}
        Input Image  &  \shortstack{FoI  \\ \normalsize{\cite{guo2024focus}}} & \shortstack[c]{Keyword's\\ Attention Map} &\textbf{\name (Ours)} & \shortstack[c]{Binary Mask \\ of \texttt{[MASK]}} \\

        \includegraphics[width=0.18\textwidth]{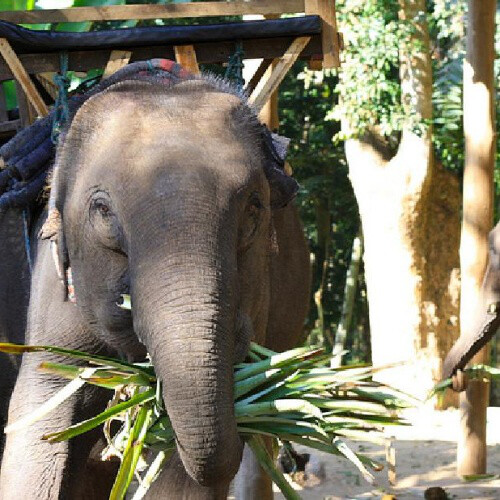}&
        \includegraphics[width=0.18\textwidth]{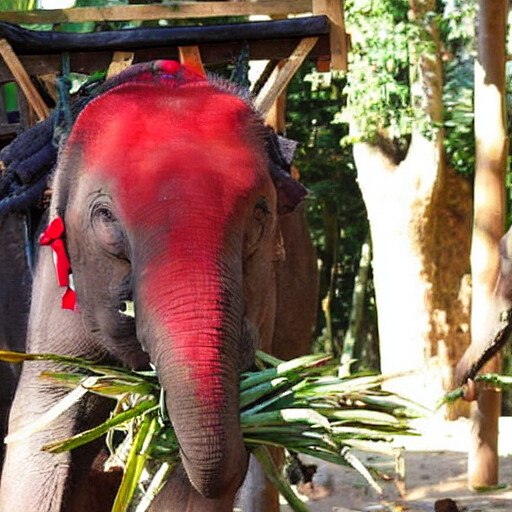}&
        \includegraphics[width=0.18\textwidth]{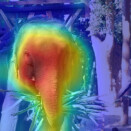}&
        \includegraphics[width=0.18\textwidth]{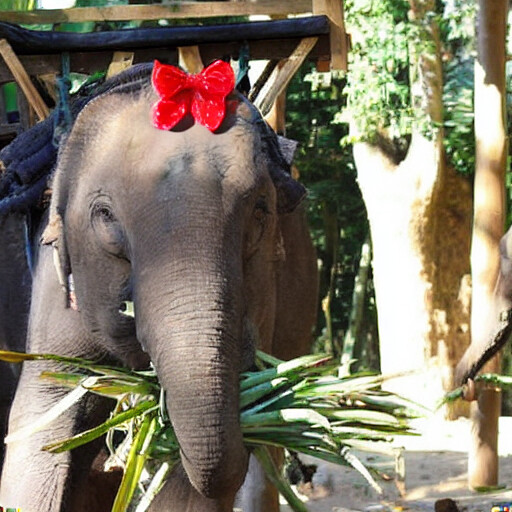}&
        \includegraphics[width=0.18\textwidth]{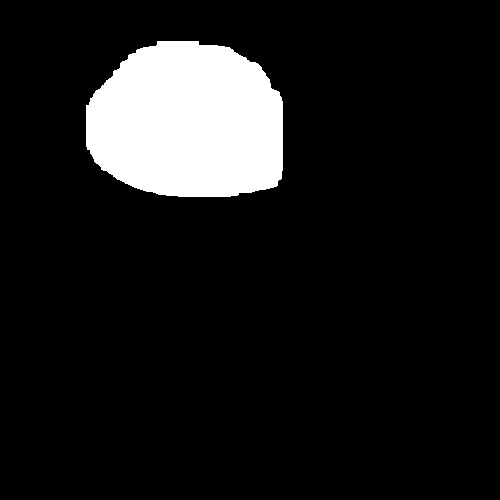}
        \\
        \multicolumn{5}{c}{\large (a) Edit instruction: \emph{``Put a red bow on the \textbf{elephant}'s head."}}
        \\
        
        \includegraphics[width=0.18\textwidth]{figure/Single/425609/425609-output2.jpg}&
        \includegraphics[width=0.18\textwidth]{figure/Single/425609/425609_2_foi.jpg}&
        \includegraphics[width=0.18\textwidth]{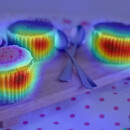}&
        \includegraphics[width=0.18\textwidth]{figure/Single/425609/425609_2_camedit.jpg}&
        \includegraphics[width=0.18\textwidth]{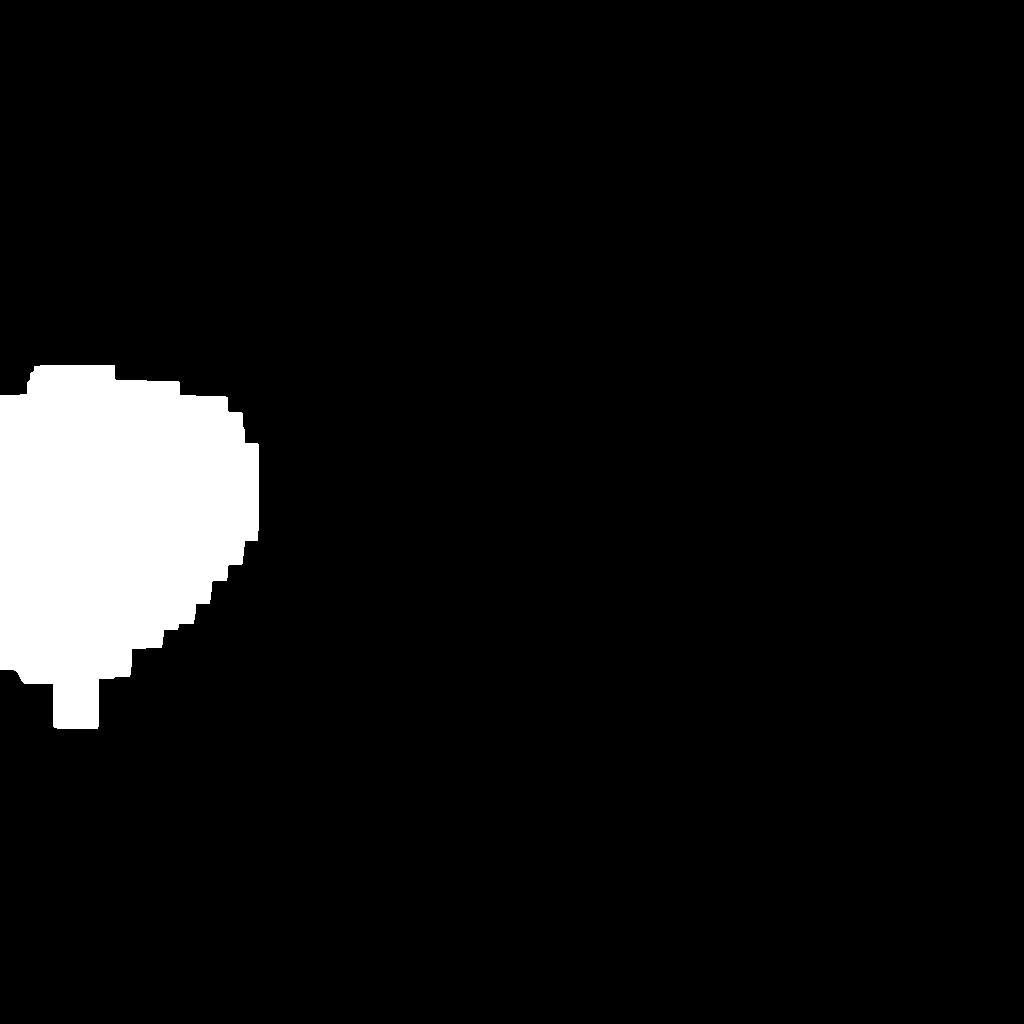}
        \\
        \multicolumn{5}{c}{\large (b) Edit instruction: \emph{``Add a strawberry on top of the \textbf{cupcake} with frosting."}}
        \\
        
        \includegraphics[width=0.18\textwidth]{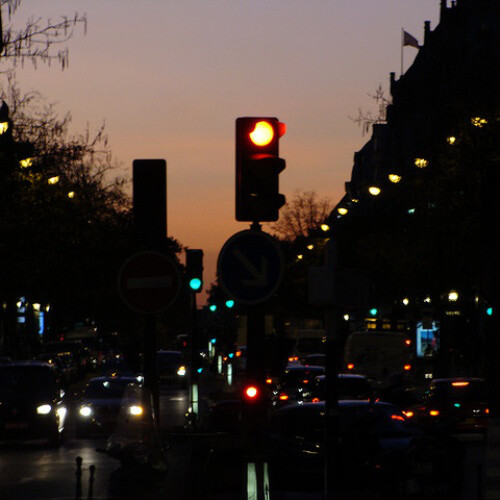}&
        \includegraphics[width=0.18\textwidth]{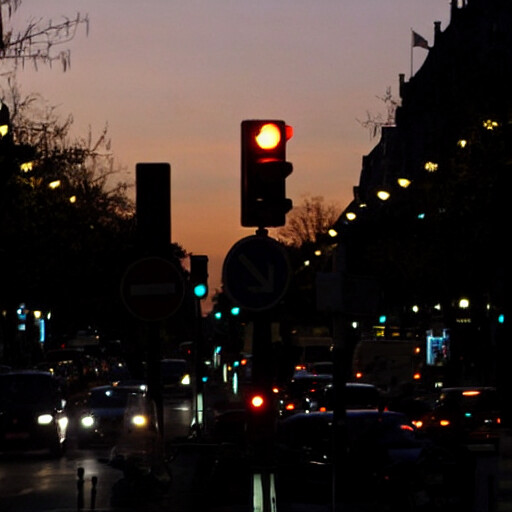}&
        \includegraphics[width=0.18\textwidth]{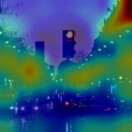}&
        \includegraphics[width=0.18\textwidth]{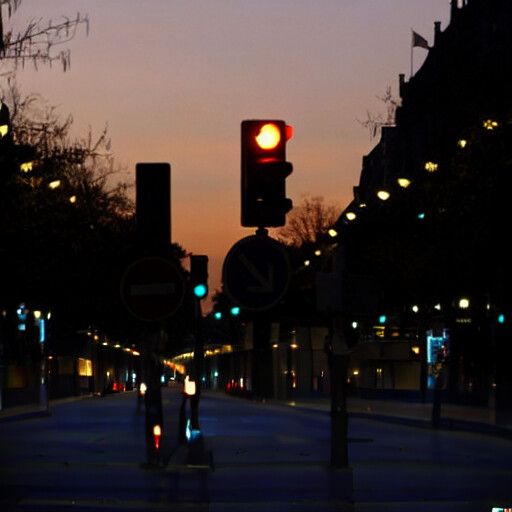}&
        \includegraphics[width=0.18\textwidth]{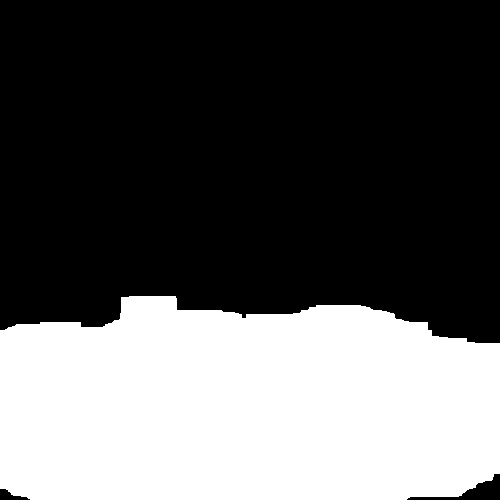}
        \\
        \multicolumn{5}{c}{\large (c) Edit instruction: \emph{``Make the \textbf{street} empty."}}
        \\
        
        \includegraphics[width=0.18\textwidth]{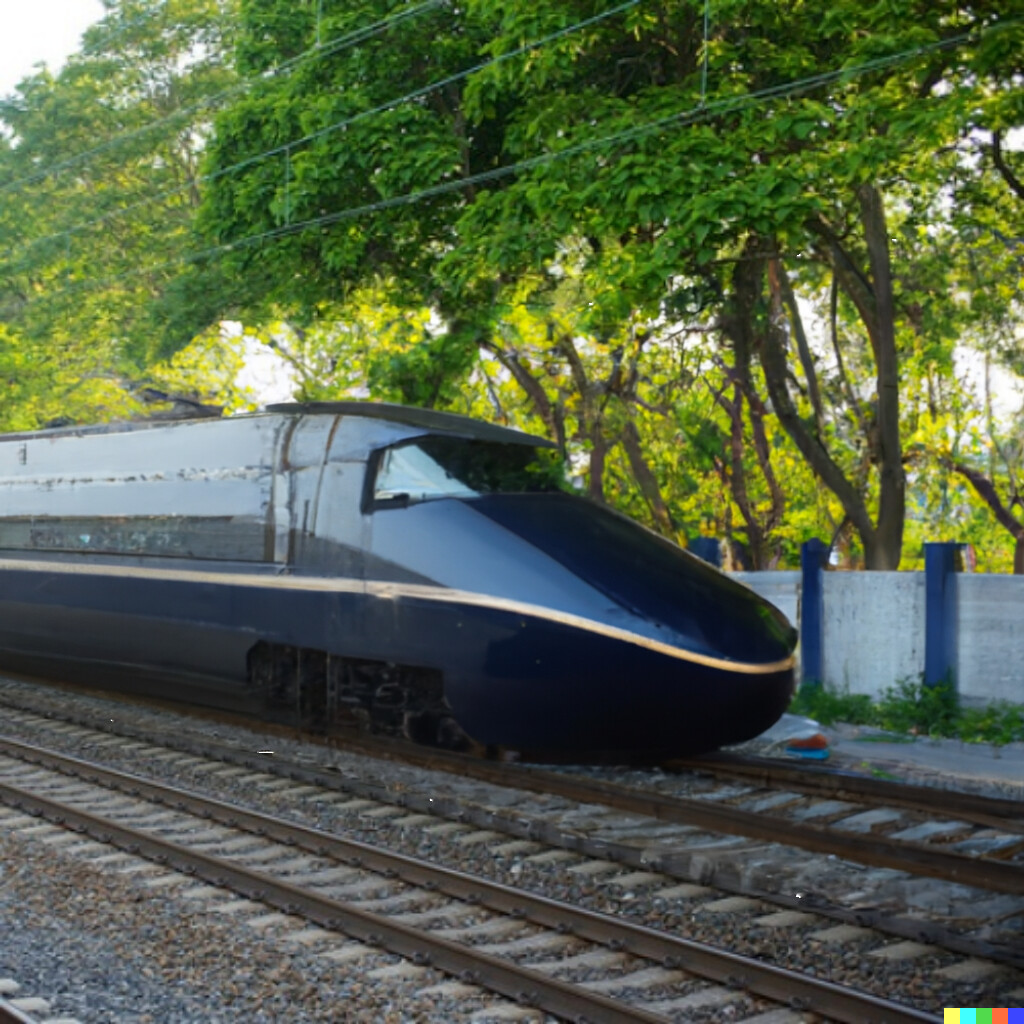}&
        \includegraphics[width=0.18\textwidth]{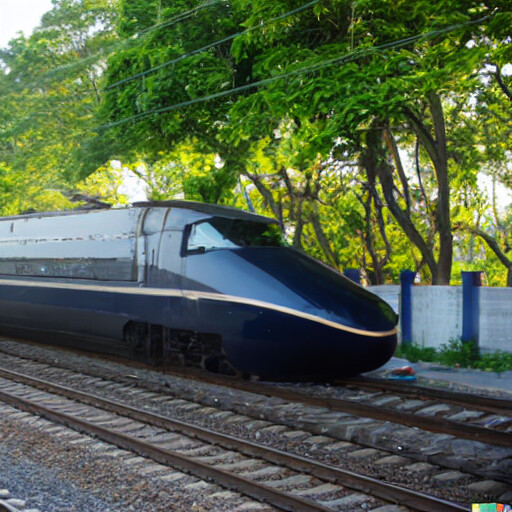}&
        \includegraphics[width=0.18\textwidth]{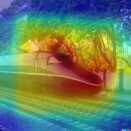}&
        \includegraphics[width=0.18\textwidth]{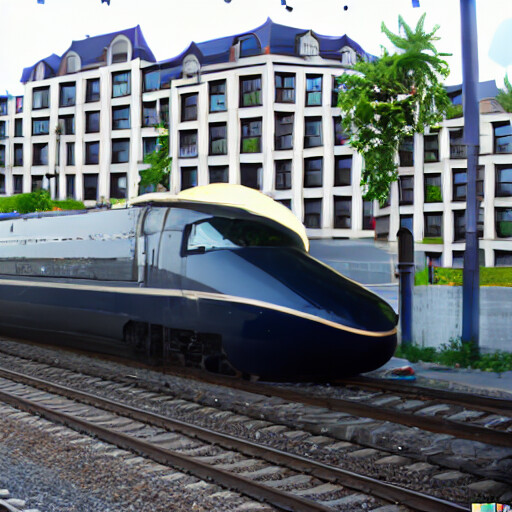}&
        \includegraphics[width=0.18\textwidth]{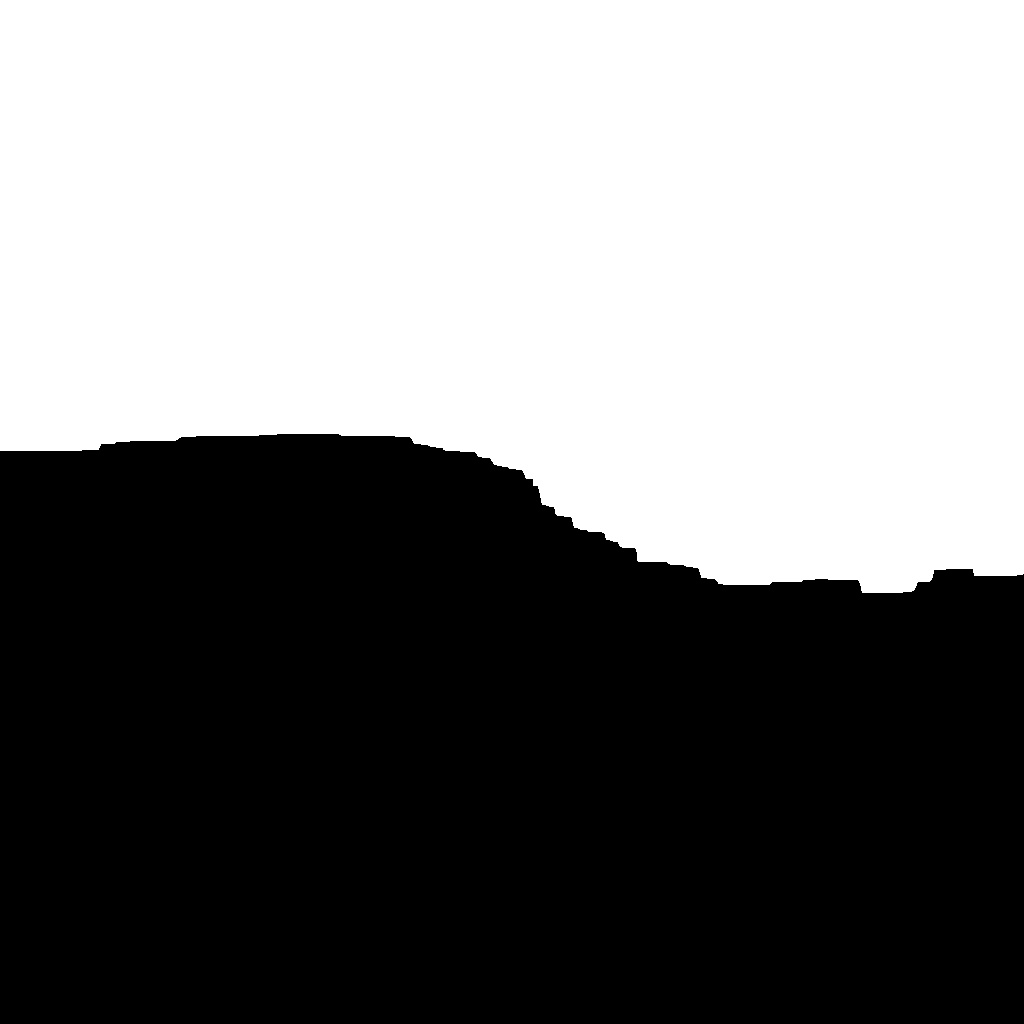}
        \\
        \multicolumn{5}{c}{\large (d) Edit instruction: \emph{``Put buildings in the \textbf{background} of the image."}}
        \\
        
        \includegraphics[width=0.18\textwidth]{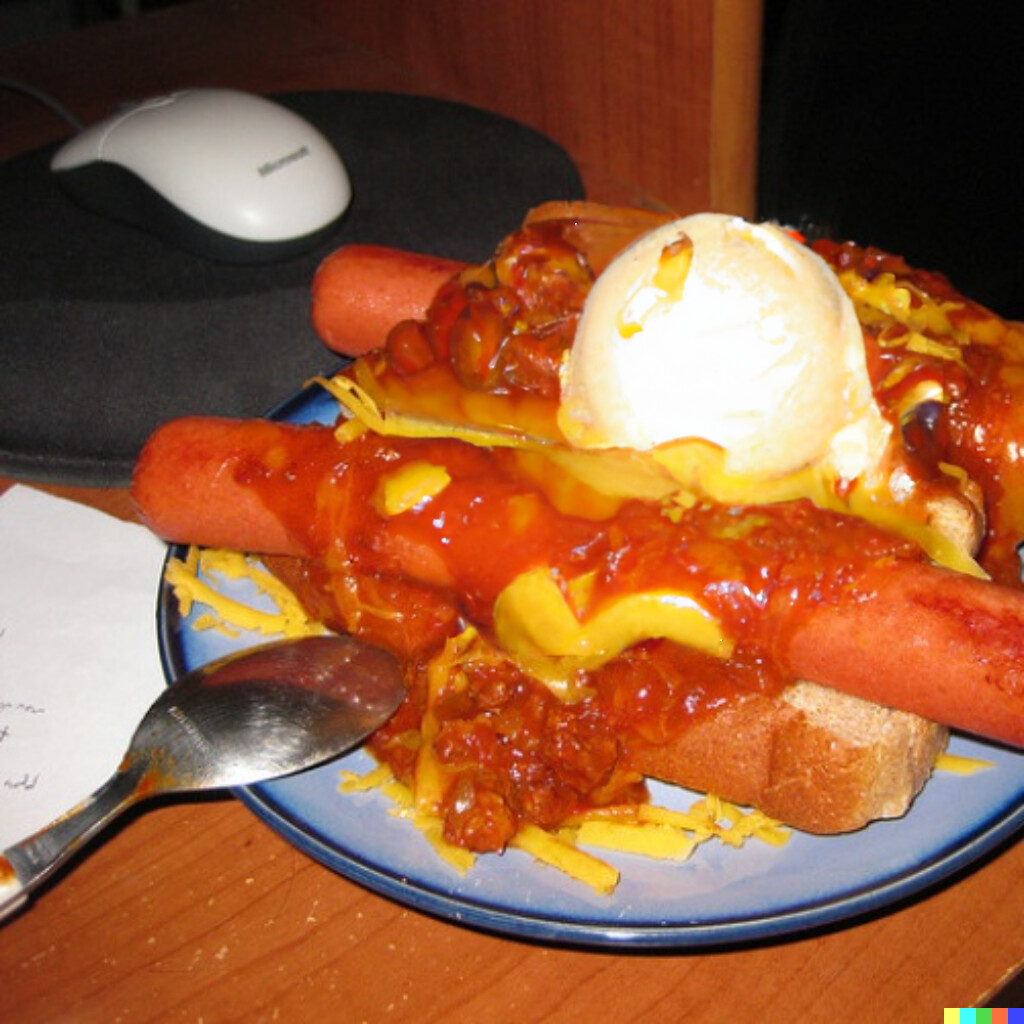}&
        \includegraphics[width=0.18\textwidth]{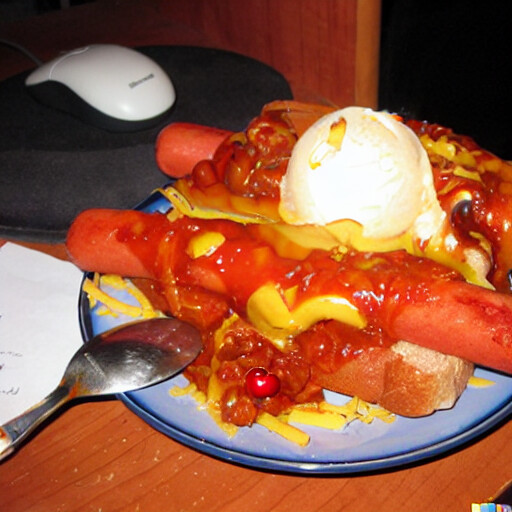}&
        \includegraphics[width=0.18\textwidth]{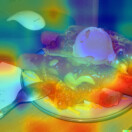}&
        \includegraphics[width=0.18\textwidth]{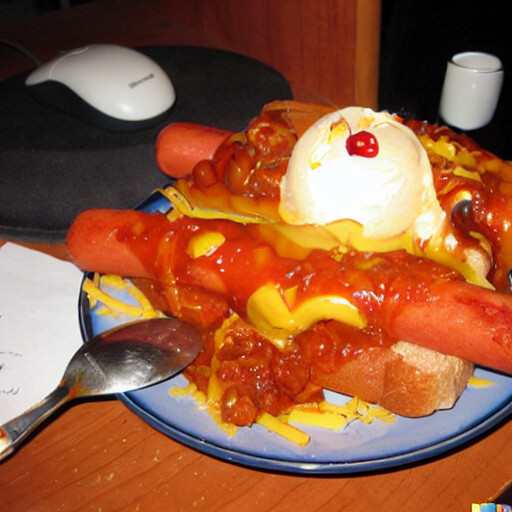}&
        \includegraphics[width=0.18\textwidth]{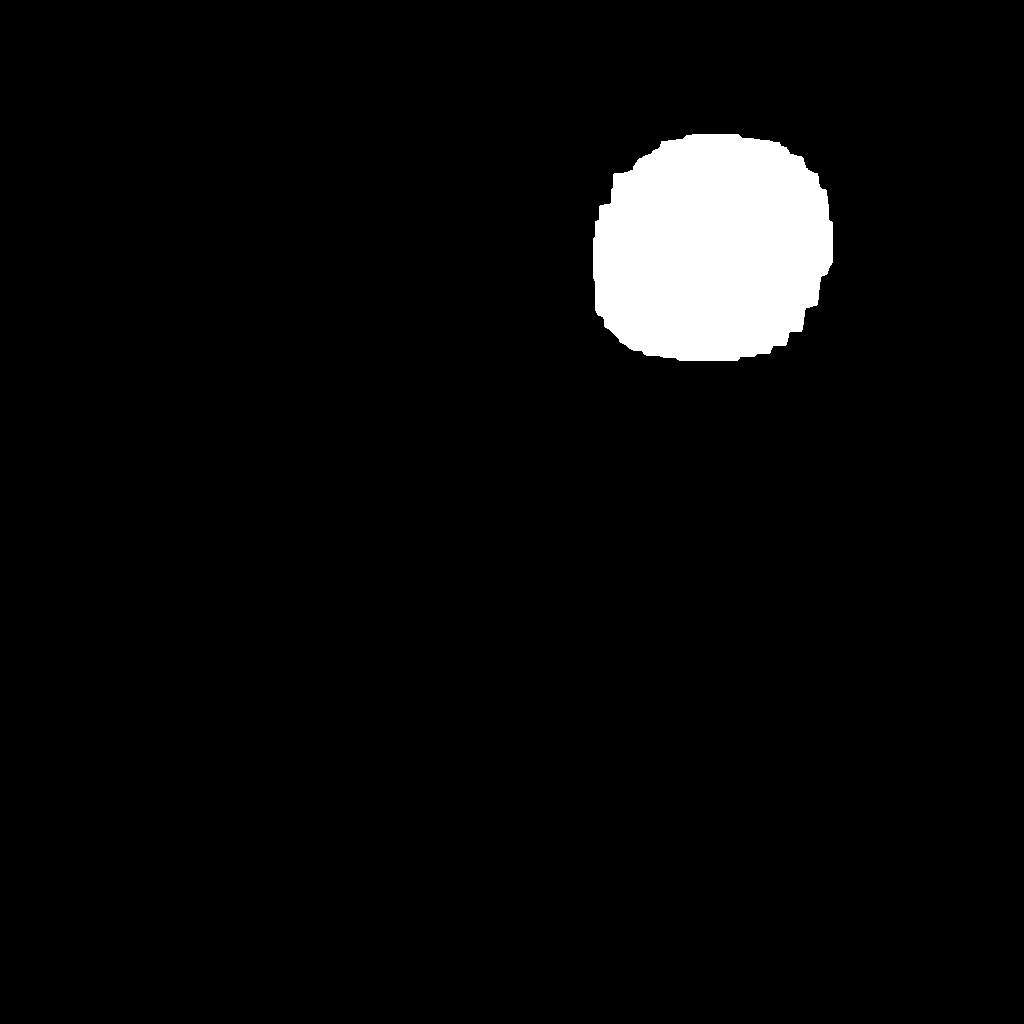}
        \\
        \multicolumn{5}{c}{\large (e) Edit instruction: \emph{``Add a cherry on \textbf{top}."}}
        \\
        
        \includegraphics[width=0.18\textwidth]{figure/Single/62745/62745-input.jpg}&
        \includegraphics[width=0.18\textwidth]{figure/Single/62745/62745_0_foi.jpg}&
        \includegraphics[width=0.18\textwidth]{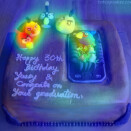}&
        \includegraphics[width=0.18\textwidth]{figure/Single/62745/62745_0_camedit.jpg}&
        \includegraphics[width=0.18\textwidth]{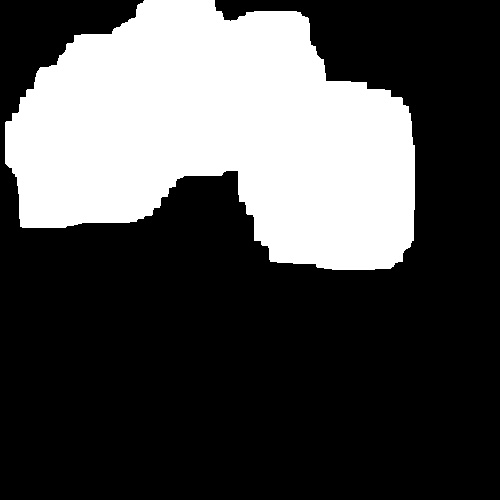}
        \\
        \multicolumn{5}{c}{\large (f) Edit instruction: \emph{``Replace the angry \textbf{birds} with flowers."}}
        \\
        
    \end{tabular}}
    \caption{\textbf{Qualitative comparisons between \name and FoI.} FoI relies on attention maps derived from extracted keywords (highlighted in \emph{\textbf{bold}}), which often lack context-awareness, leading to inaccurate edits. The red regions in the attention maps represent areas with higher attention intensity, indicating where the model focuses during image editing. In contrast, our proposed method, \name, utilizes context-aware \texttt{[MASK]} decoding to generate precise binary masks, enabling more accurate and instruction-compliant image modifications.}

    \label{fig:comparison}
\end{figure*}

\section{Additional Results}

\subsection{Preference-Based Evaluation Results}
\label{sec:pick}
We additionally report PickScore~\cite{kirstain2023pick}, a metric trained to reflect human preferences in image generation tasks. As shown in \cref{tab:pickscore}, our method achieves the highest PickScore under both multi-instruction and context-aware settings, outperforming IP2P, MGIE, SmartEdit, and FoI. These results suggest that our model generates outputs that are more aligned with human preferences compared to the baselines.

\subsection{Comparison with Multi-instruction-based Method}
As illustrated in \cref{fig:comparison}, we compare our method with FoI~\cite{guo2024focus}, one of the state-of-the-art multi-instruction image editing approaches. FoI employs a pretrained GPT model~\cite{gpt4,gpt4v} to extract keywords, which are subsequently used as masks based on the cross-attention maps of the U-Net. While the attention maps in (a) and (b) appear to reflect the objects in the image, they lead to suboptimal results due to a lack of context-awareness. In (a), the attention map focuses on the elephant’s face instead of the head, which the instruction specifies, resulting in an inaccurate edit. Similarly, in (b), while the instruction specifies ``cupcake with frosting," the attention map erroneously attends to all three cupcakes in the image instead of isolating the one on the far left, leading to an incorrect edit. In contrast, the decoded \texttt{[MASK]} generated by our method leverages the contextual understanding of the instruction to accurately identify the specific cupcake requiring modification. In case of (c), it highlights the impact of an incorrect attention map, where the edit fails to meet the instruction. In (d) and (e), the extracted keywords, such as ``background" or ``top", are vague. Even if keywords like ``building" or ``cherry," which represent the objects intended to be added, were selected, they are not present in the given image, meaning the attention maps would still fail to provide meaningful guidance in such cases. In (f), while the attention map partially focuses on the bird, areas with weaker attention intensity remain unedited. By contrast, our proposed method, \name, generates binary masks using the context-aware capabilities of the MLLM to precisely determine which regions need modification. This approach enables more accurate and context-aware image editing compared to existing methods, as demonstrated in the provided examples.

\subsection{Failure Cases}
\label{appendix:failure}
\begin{figure}[ht]

    \centering
    \resizebox{0.5\linewidth}{!}{%
    
    \begin{tabular}{c |cc }
        Input Image  &\textbf{\name (Ours)} & \shortstack[c]{Binary Mask \\ of \texttt{[MASK]}} \\

        \includegraphics[width=0.18\textwidth]{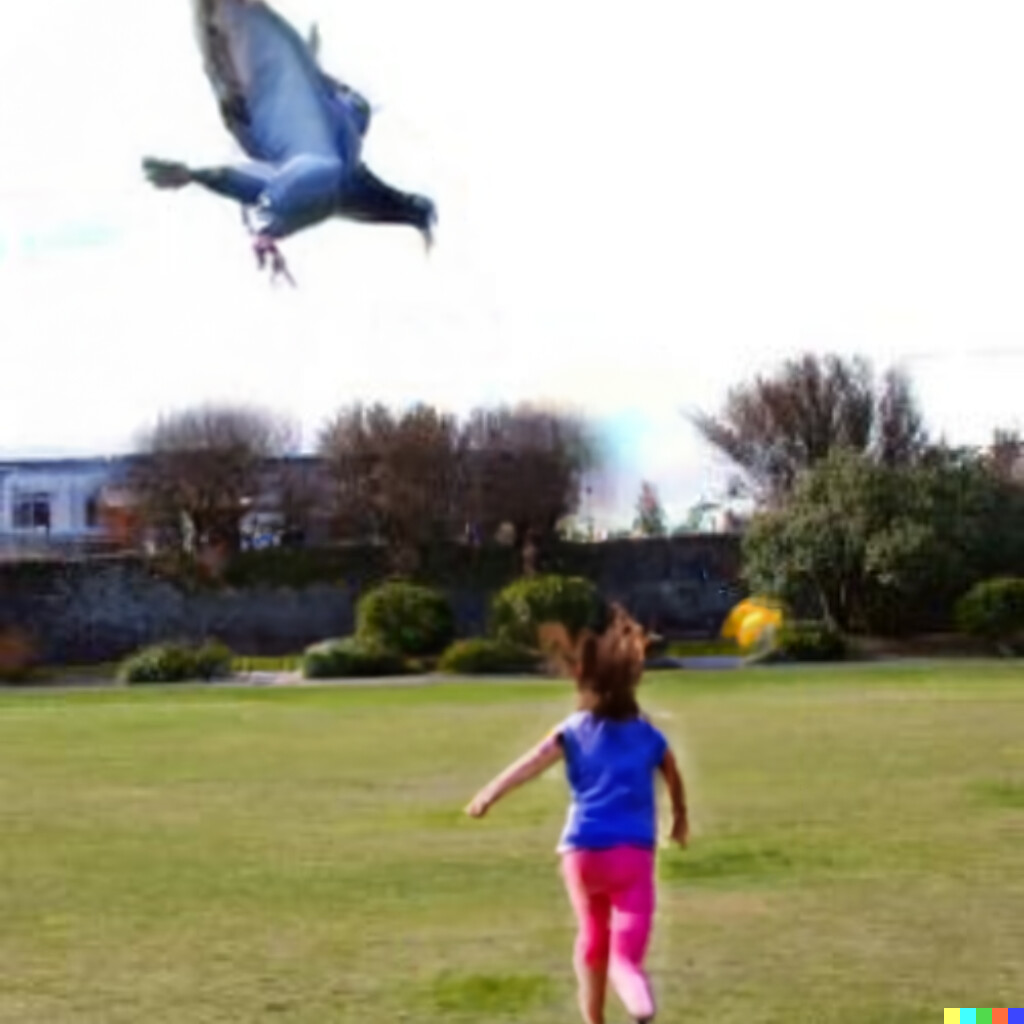}&
        \includegraphics[width=0.18\textwidth]{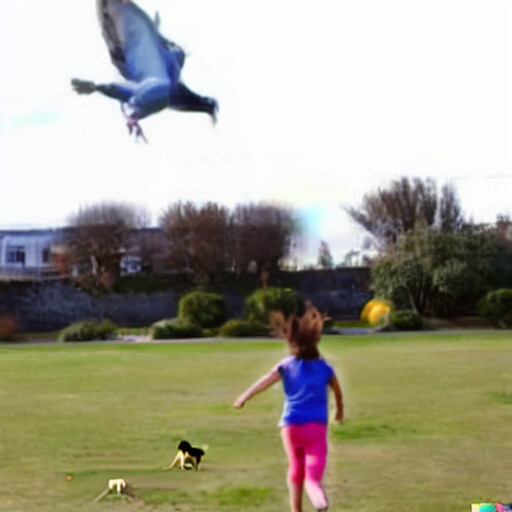}&
        \includegraphics[width=0.18\textwidth]{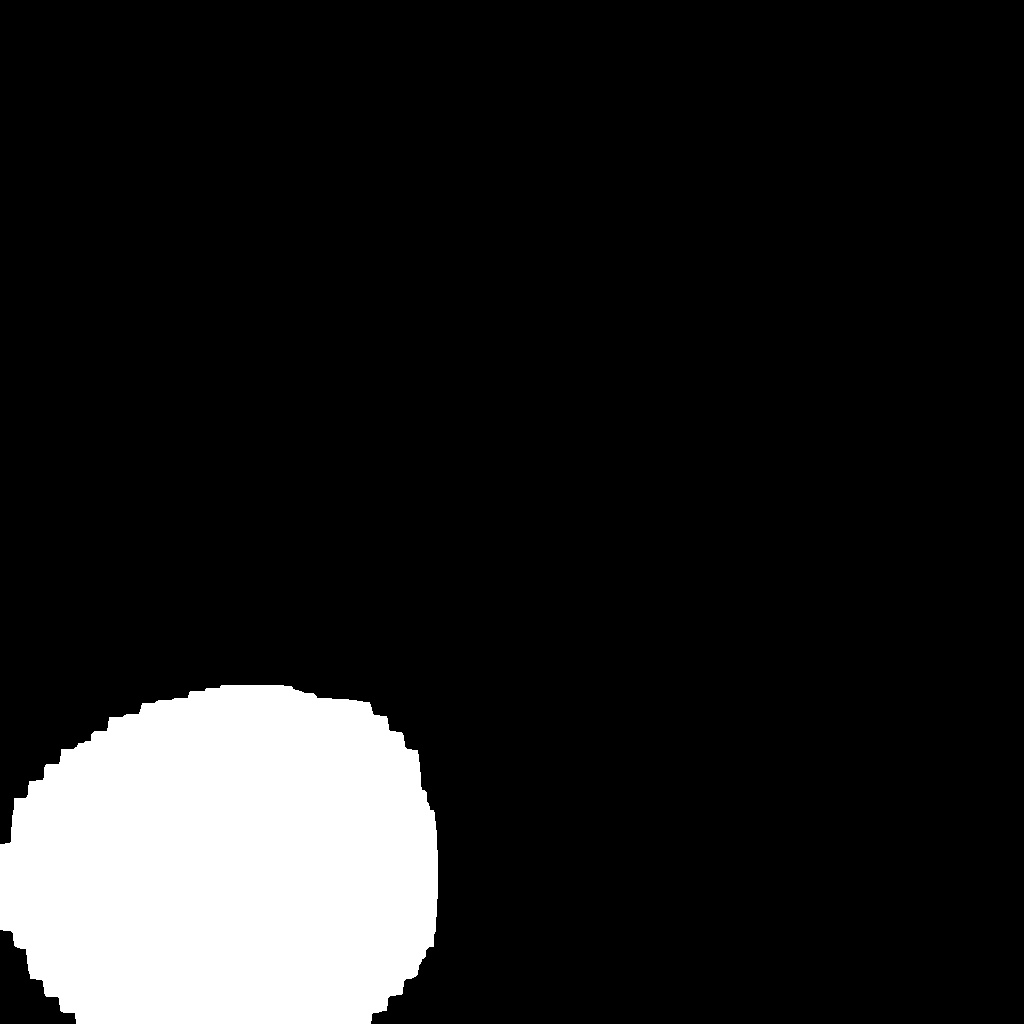}
        \\
        \multicolumn{3}{c}{\large (a) \emph{``Add a dog next to the girl.''}}
        \\
        
        \includegraphics[width=0.18\textwidth]{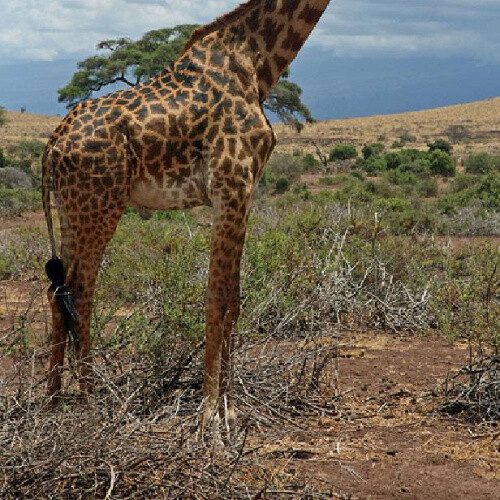}&
        \includegraphics[width=0.18\textwidth]{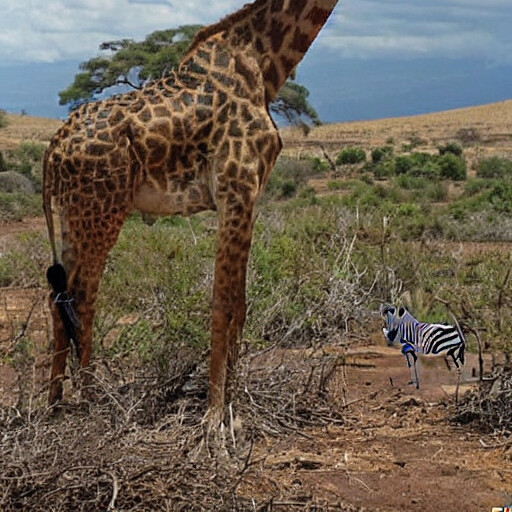}&
        \includegraphics[width=0.18\textwidth]{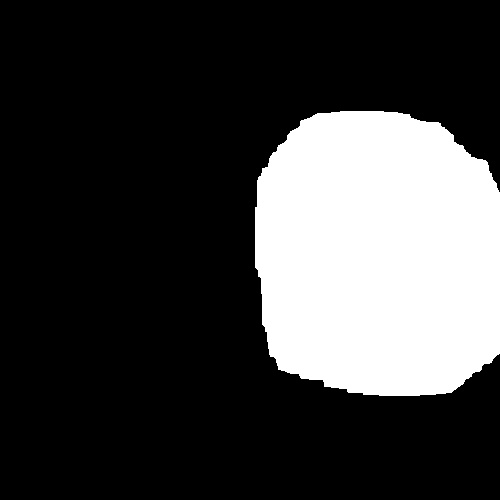}
        \\
        \multicolumn{3}{c}{\large (b) \emph{``Add a zebra.''}}
        \\
        
        \includegraphics[width=0.18\textwidth]{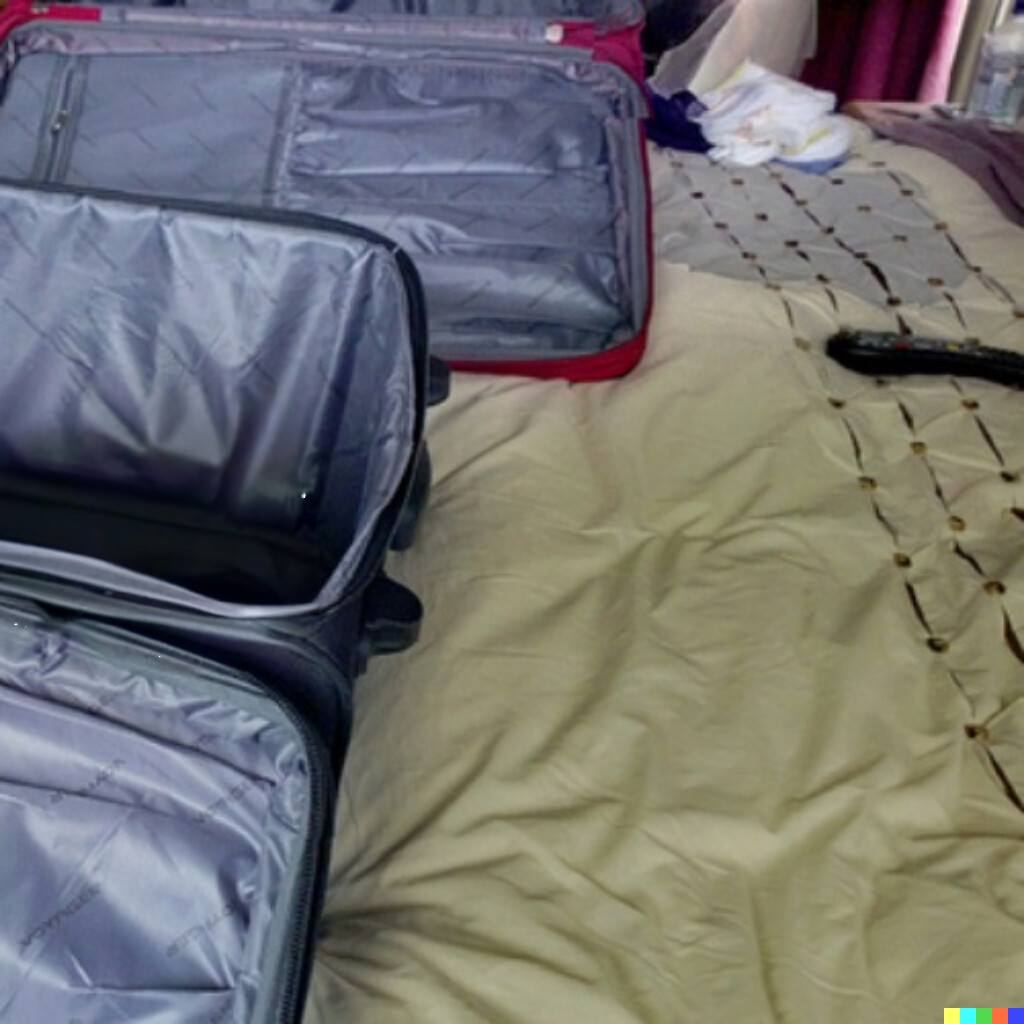}&
        \includegraphics[width=0.18\textwidth]{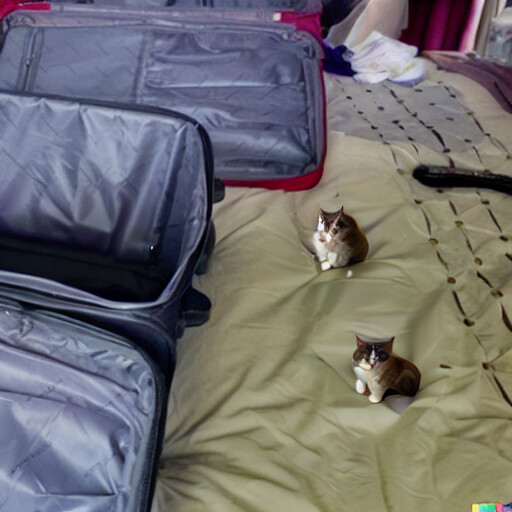}&
        \includegraphics[width=0.18\textwidth]{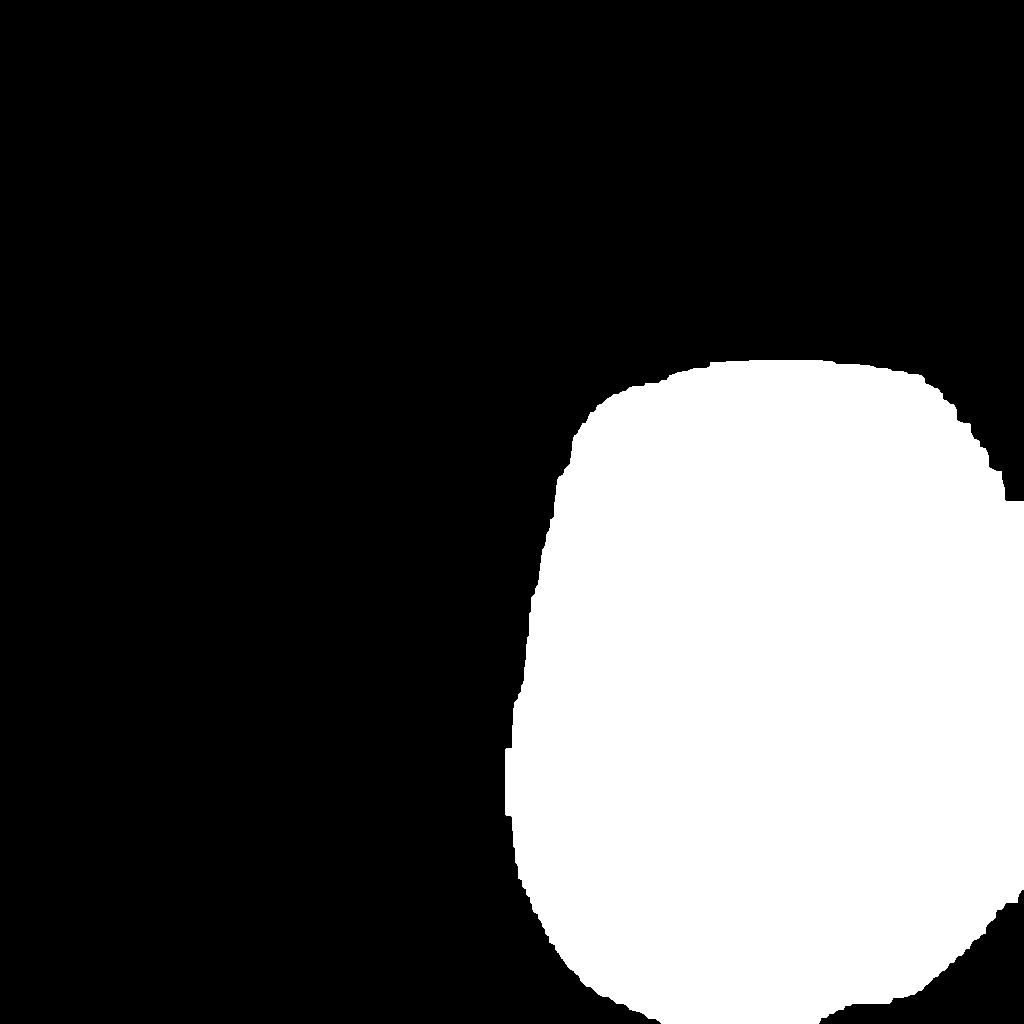}
        \\
        \multicolumn{3}{c}{\large (c) \emph{``Put a cat next to the suitcase.''}}
        \\
        
        \includegraphics[width=0.18\textwidth]{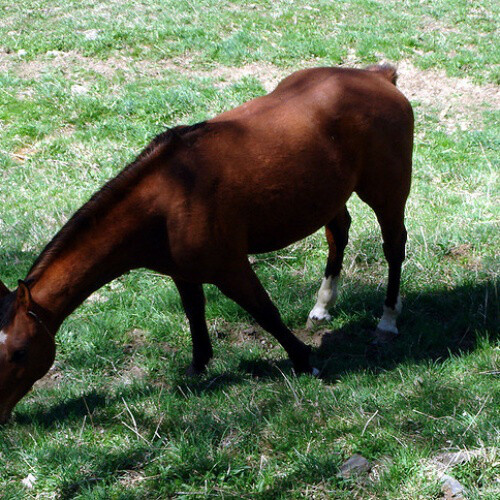}&
        \includegraphics[width=0.18\textwidth]{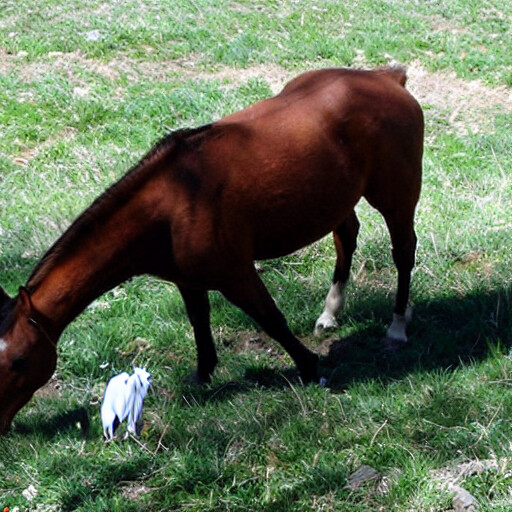}&
        \includegraphics[width=0.18\textwidth]{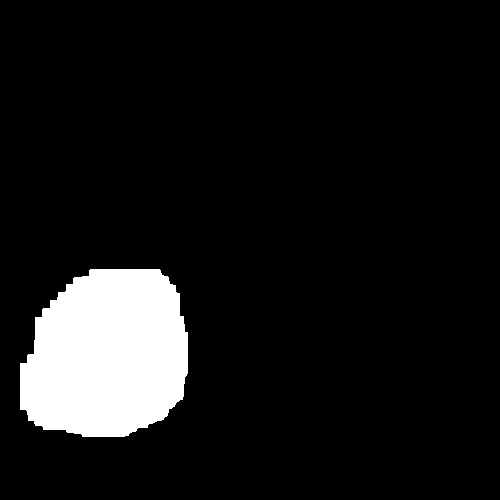}
        \\
        \multicolumn{3}{c}{\large (d) \emph{``Let's add a white goat on the grass.''}}
        \\
        
    \end{tabular}}
    \caption{\textbf{Failure cases of our method \name.} }

    \label{fig:failure}
\end{figure}

We address several cases where our method shows limitations in achieving optimal editing, primarily due to the pretrained diffusion model used in our framework. ~\cref{fig:failure} presents examples of failure cases of \name. While the binary mask identifies the intended editing regions with high accuracy, it occasionally fails to account for their relative sizes. For instance, in cases (a)--(c), although the binary mask captures the correct regions for editing, the resulting additions are smaller than expected relative to the surrounding objects. Similarly, in case (d), the generated binary mask focuses on a less relevant area, despite the presence of more suitable editing regions, such as the grass in the top-left or bottom-right corners. This results in the addition of relatively small objects. These limitations arise because the pretrained diffusion model was originally trained under the assumption that edits would be performed without masks. Incorporating more advanced diffusion models or fine-tuning the diffusion model  to better integrate with mask-based editing could further improve the performance of our proposed method, which we leave as a direction for future work.

\subsection{Further Qualitative Results}
\label{appendix:results}
\begin{figure*}[t]
    \setlength\tabcolsep{2.5pt}
    \centering
    \resizebox{\linewidth}{!}{%
    \normalsize
    \begin{tabular}{ccccccc}
        Input Image & {\footnotesize \shortstack{IP2P\\\cite{brooks2023instructpix2pix}}} & {\footnotesize \shortstack{MGIE\\\cite{fu2024guiding}}} & {\footnotesize \shortstack{SmartEdit\\\cite{huang2023smartedit}}} & {\footnotesize \shortstack{FoI\\\cite{guo2024focus}}} & \multicolumn{2}{c}{\name~\textbf{(Ours)}} \\
\includegraphics[width=0.14\textwidth]{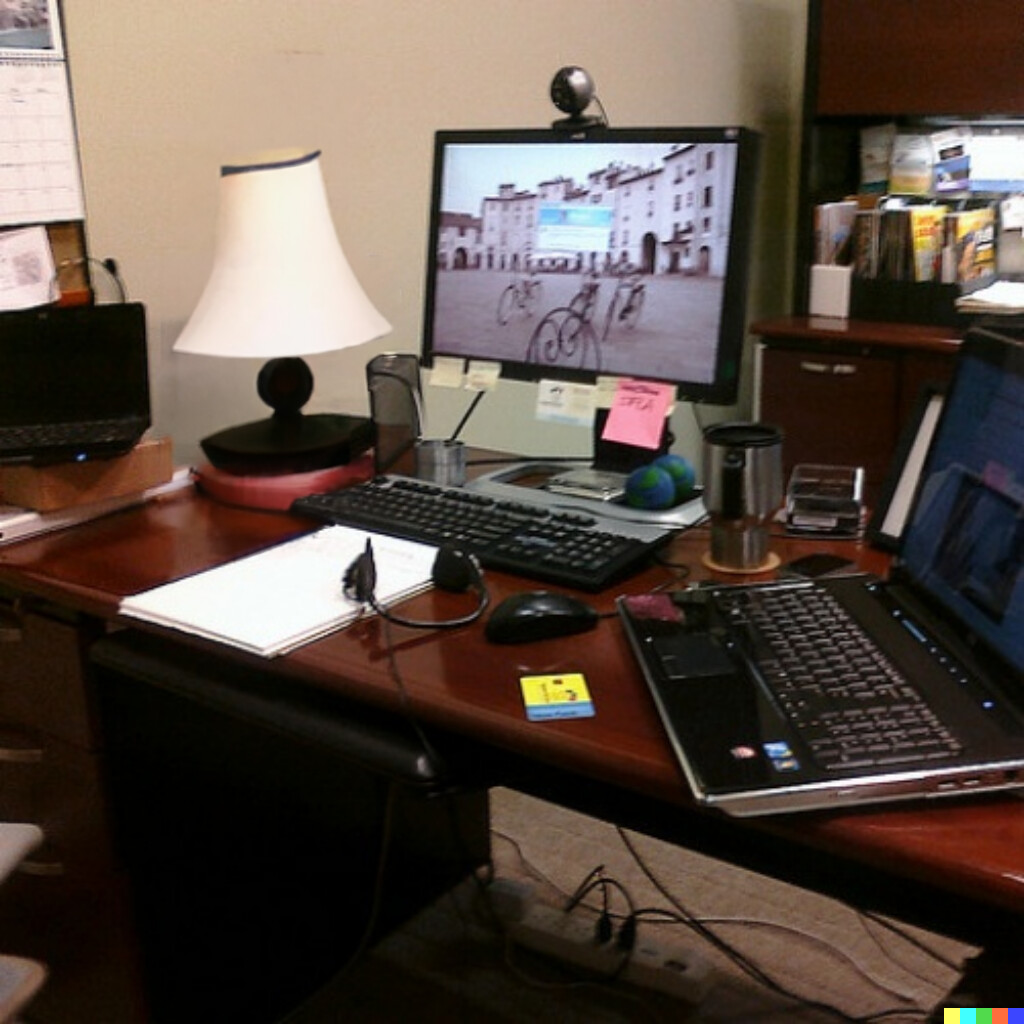}&
\includegraphics[width=0.14\textwidth]{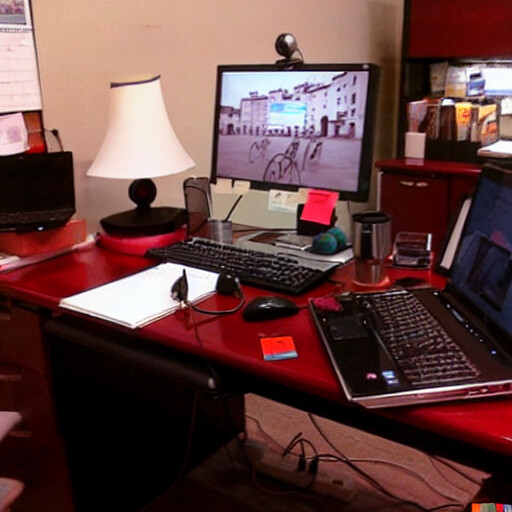}&
\includegraphics[width=0.14\textwidth]{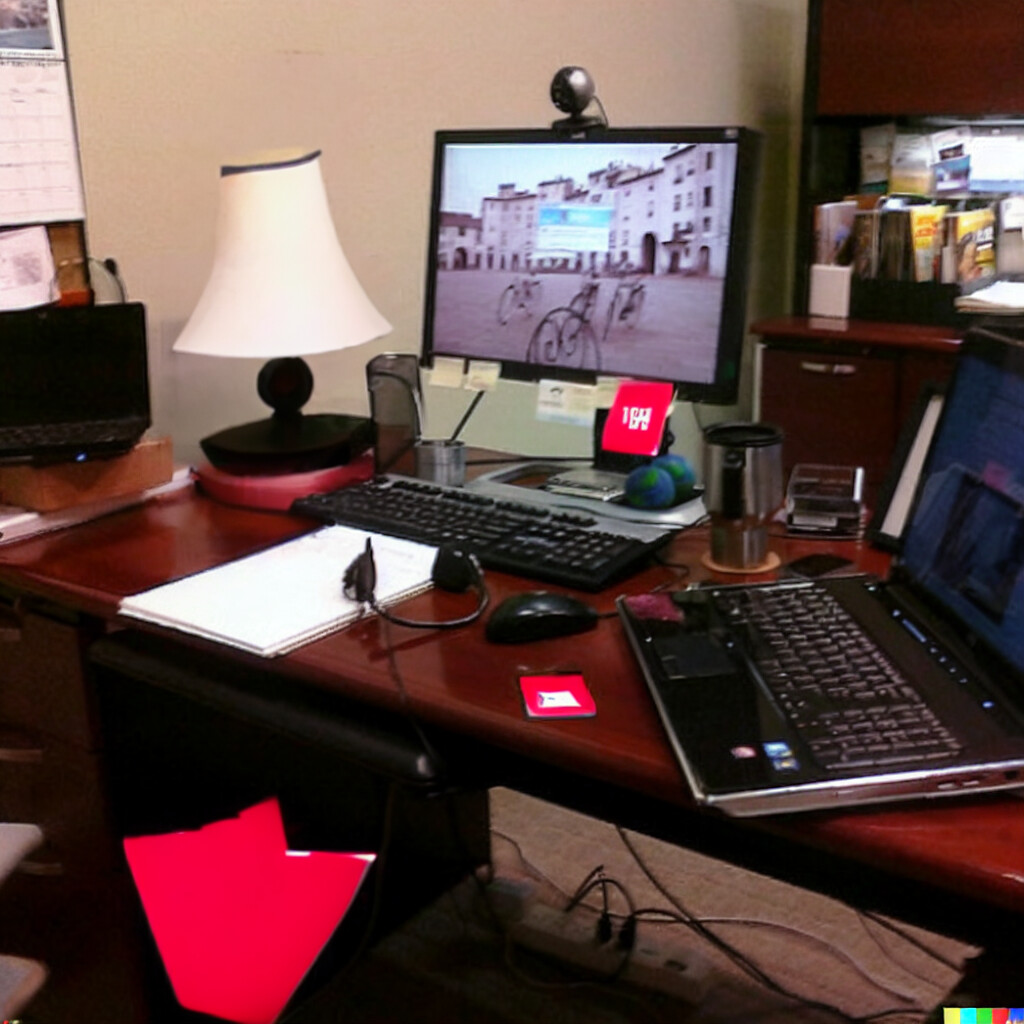}&
\includegraphics[width=0.14\textwidth]{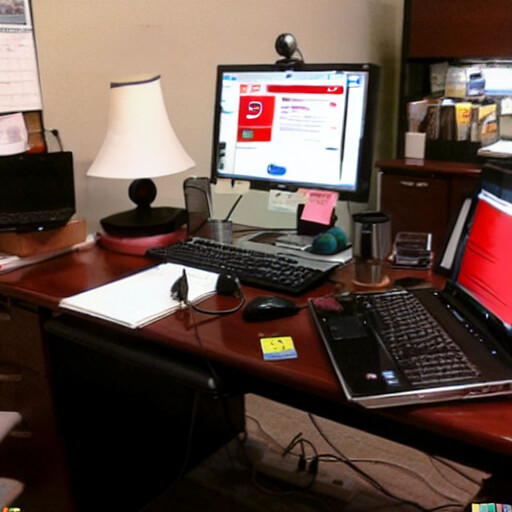}&
\includegraphics[width=0.14\textwidth]{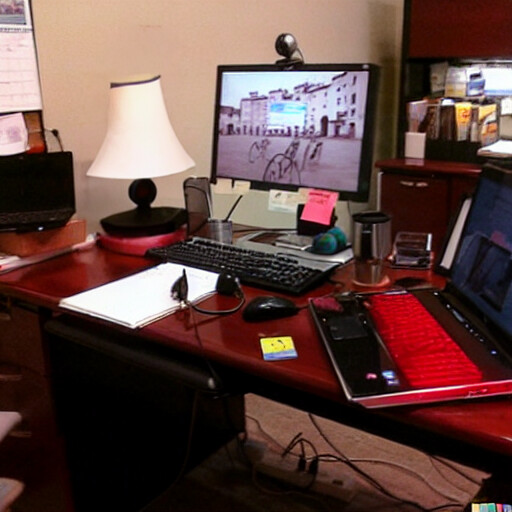}&
\includegraphics[width=0.14\textwidth]{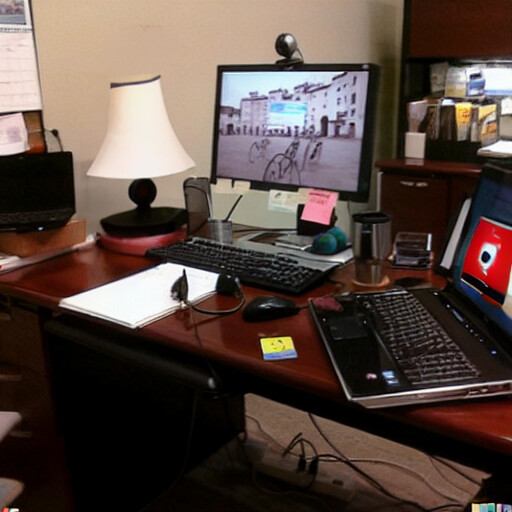}&
\shortstack{
\includegraphics[width=0.06\textwidth]{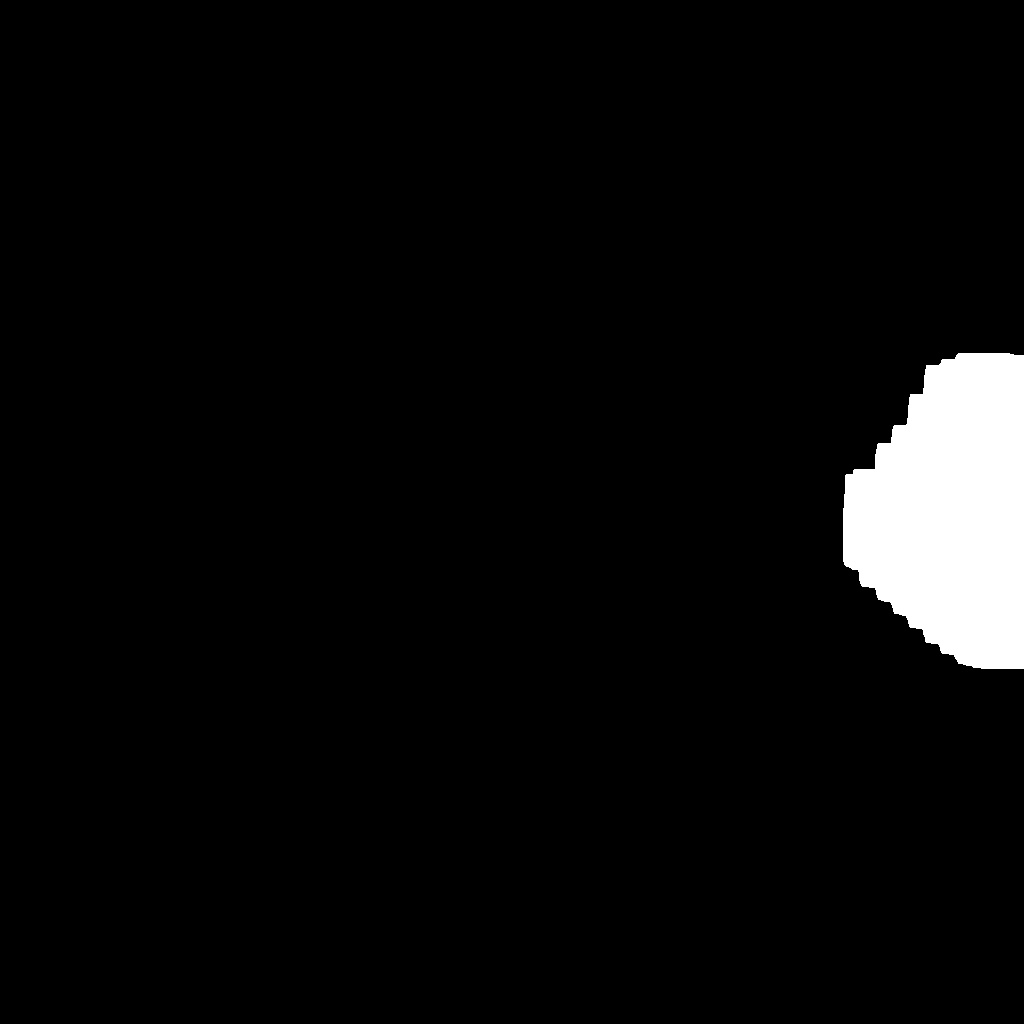} \\
\texttt{[MASK]}
}
 \\

  \multicolumn{7}{c}{(a) Edit instruction: \emph{``Let the laptop have a red webpage.''}} \\
\includegraphics[width=0.14\textwidth]{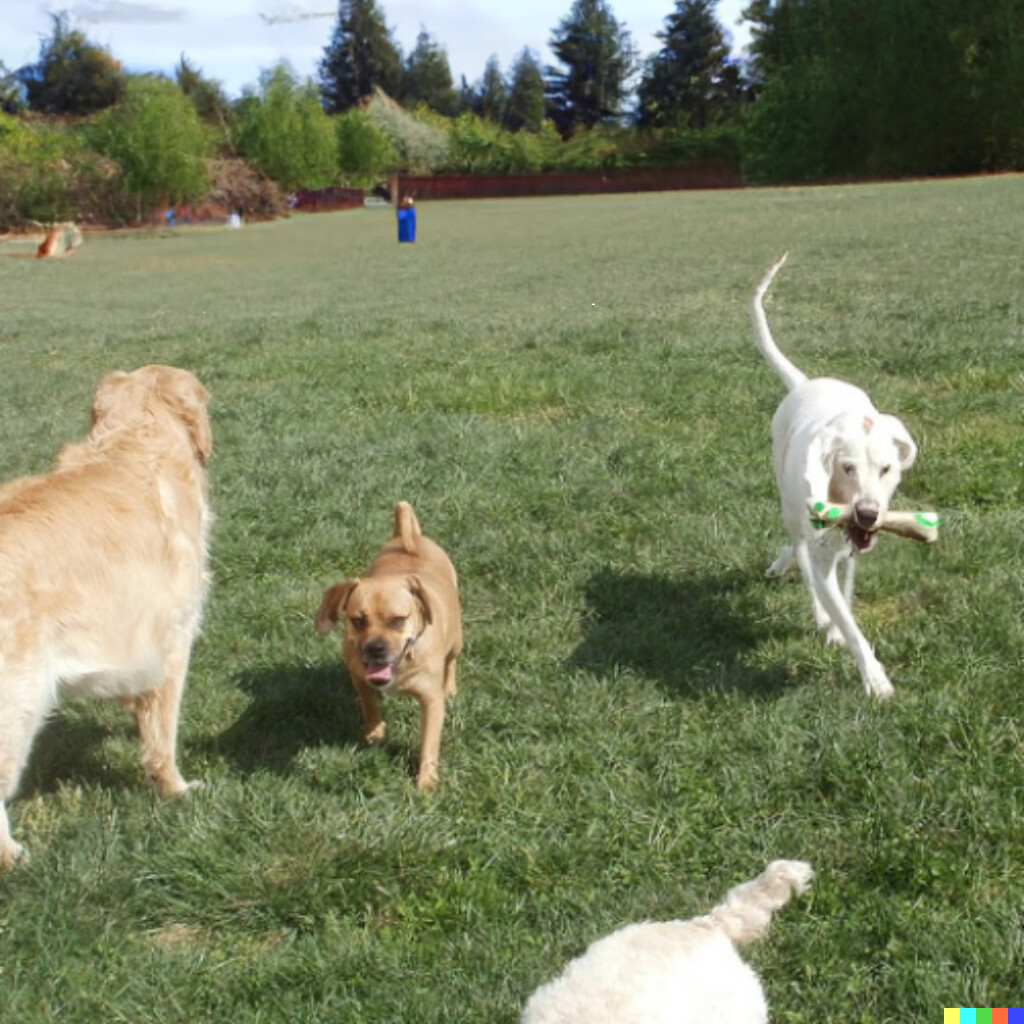}&
\includegraphics[width=0.14\textwidth]{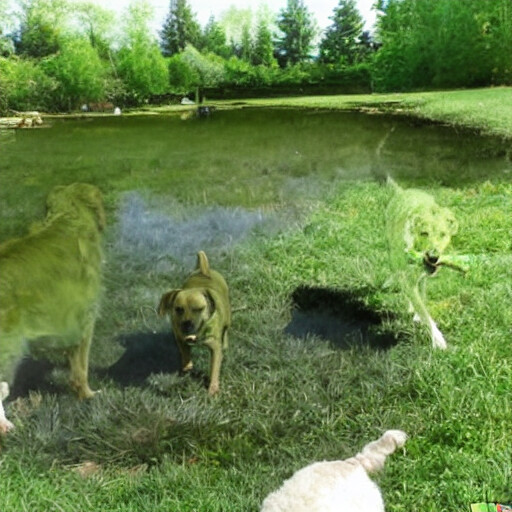}&
\includegraphics[width=0.14\textwidth]{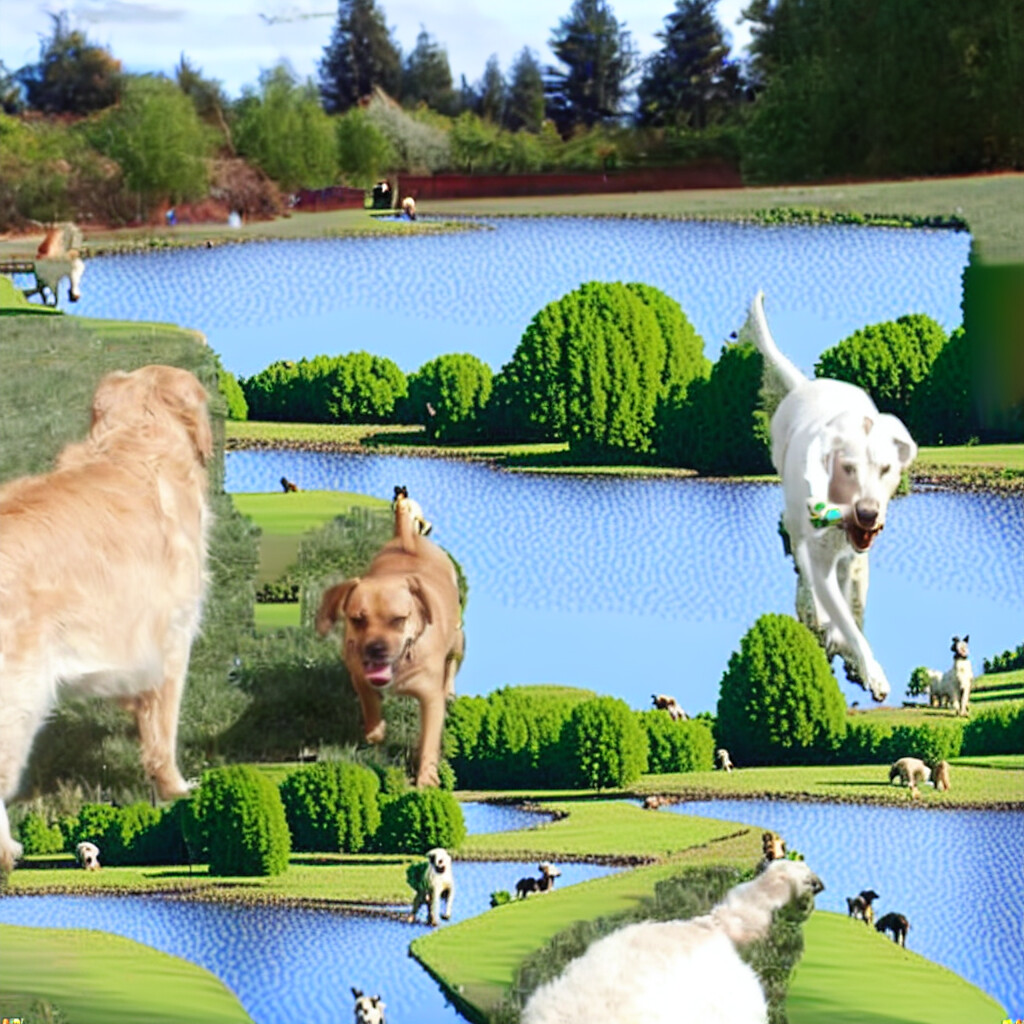}&
\includegraphics[width=0.14\textwidth]{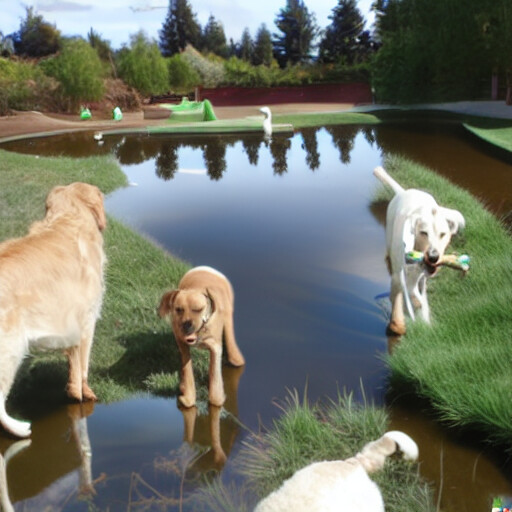}&
\includegraphics[width=0.14\textwidth]{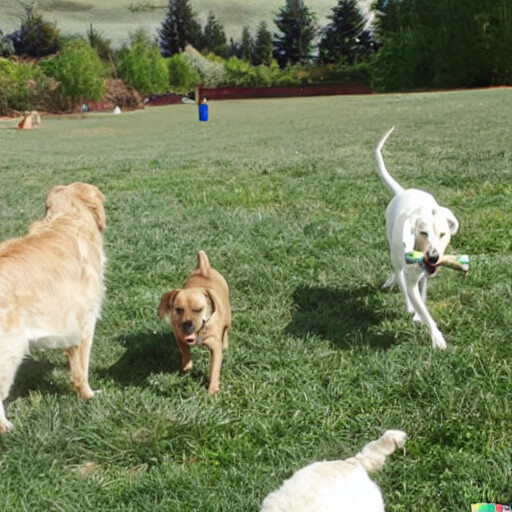}&
\includegraphics[width=0.14\textwidth]{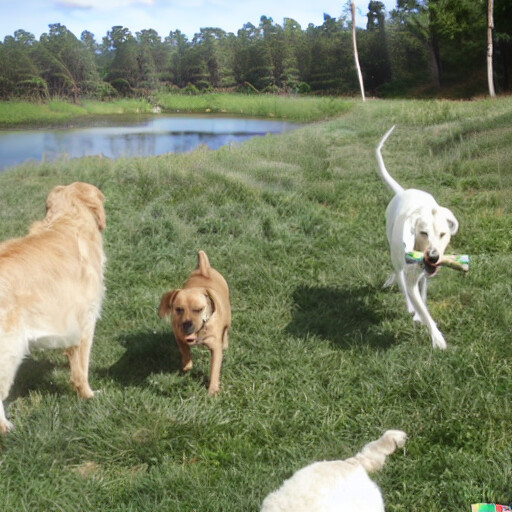}&
\shortstack{
\includegraphics[width=0.06\textwidth]{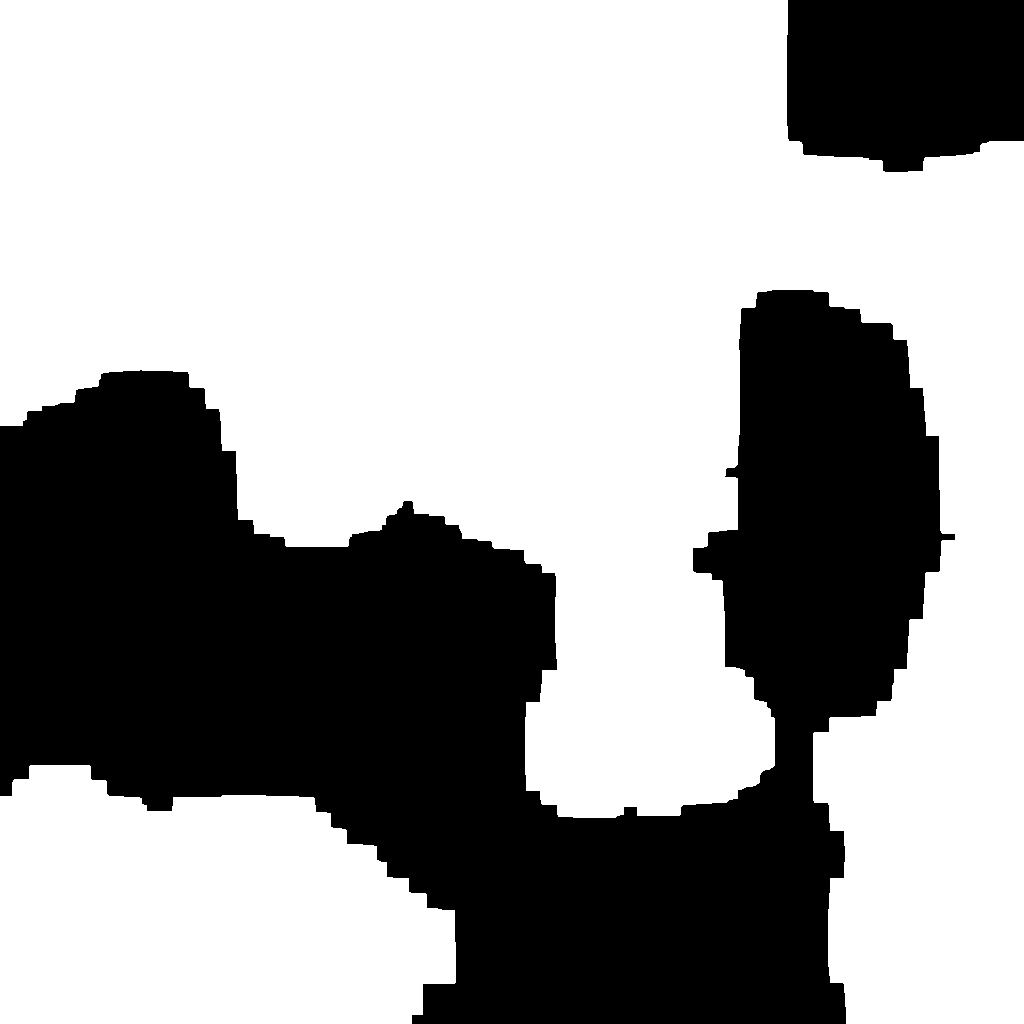} \\
\texttt{[MASK]} 
}
 \\

   \multicolumn{7}{c}{(b) Edit instruction: \emph{``Could we have a pond next to the forest?''}} \\ 
\includegraphics[width=0.14\textwidth]{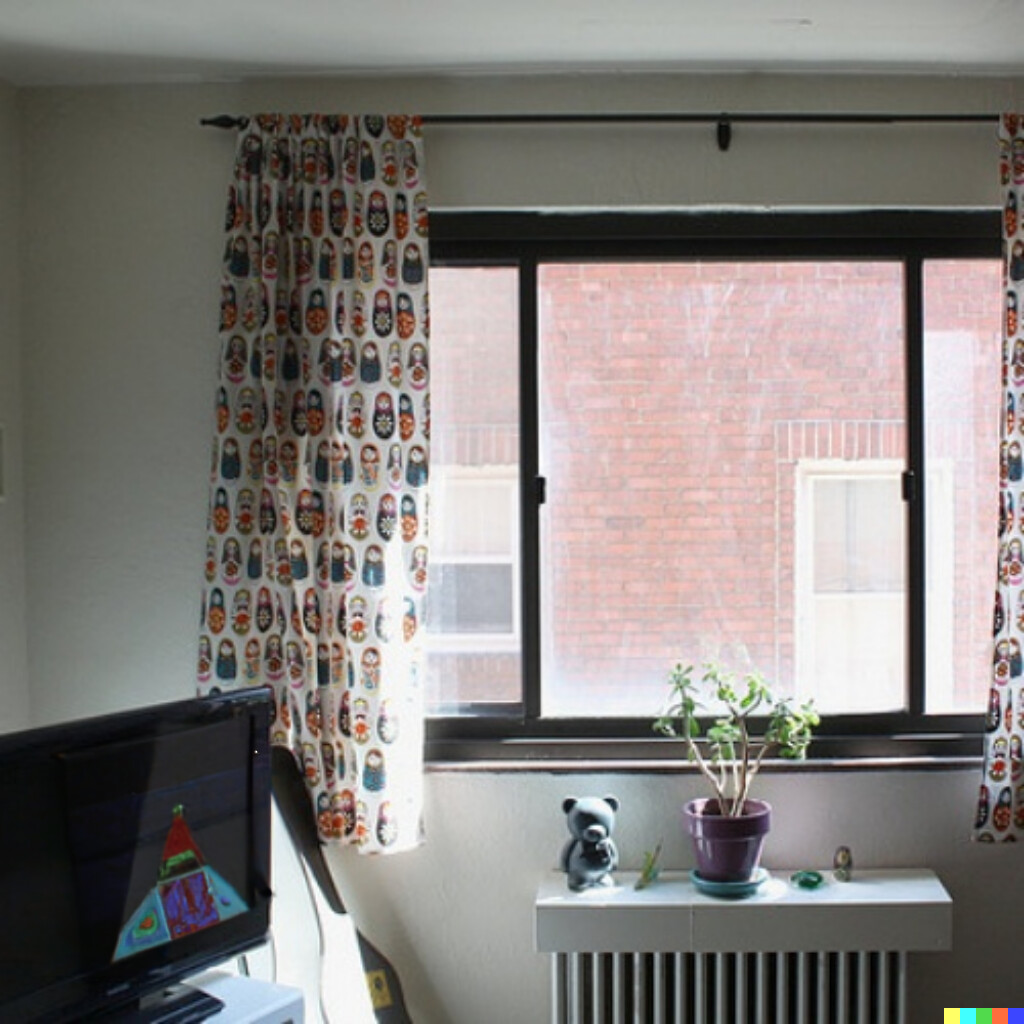}&
\includegraphics[width=0.14\textwidth]{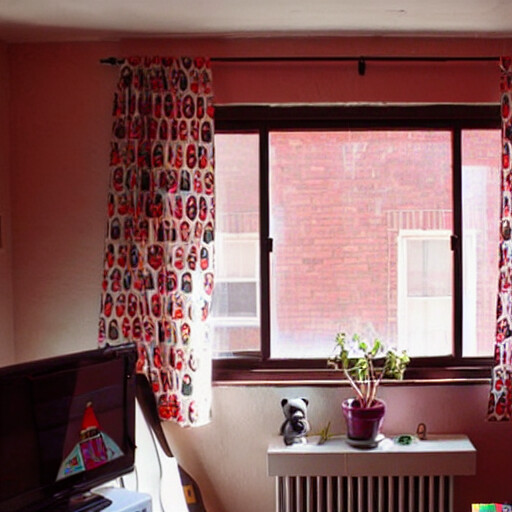}&
\includegraphics[width=0.14\textwidth]{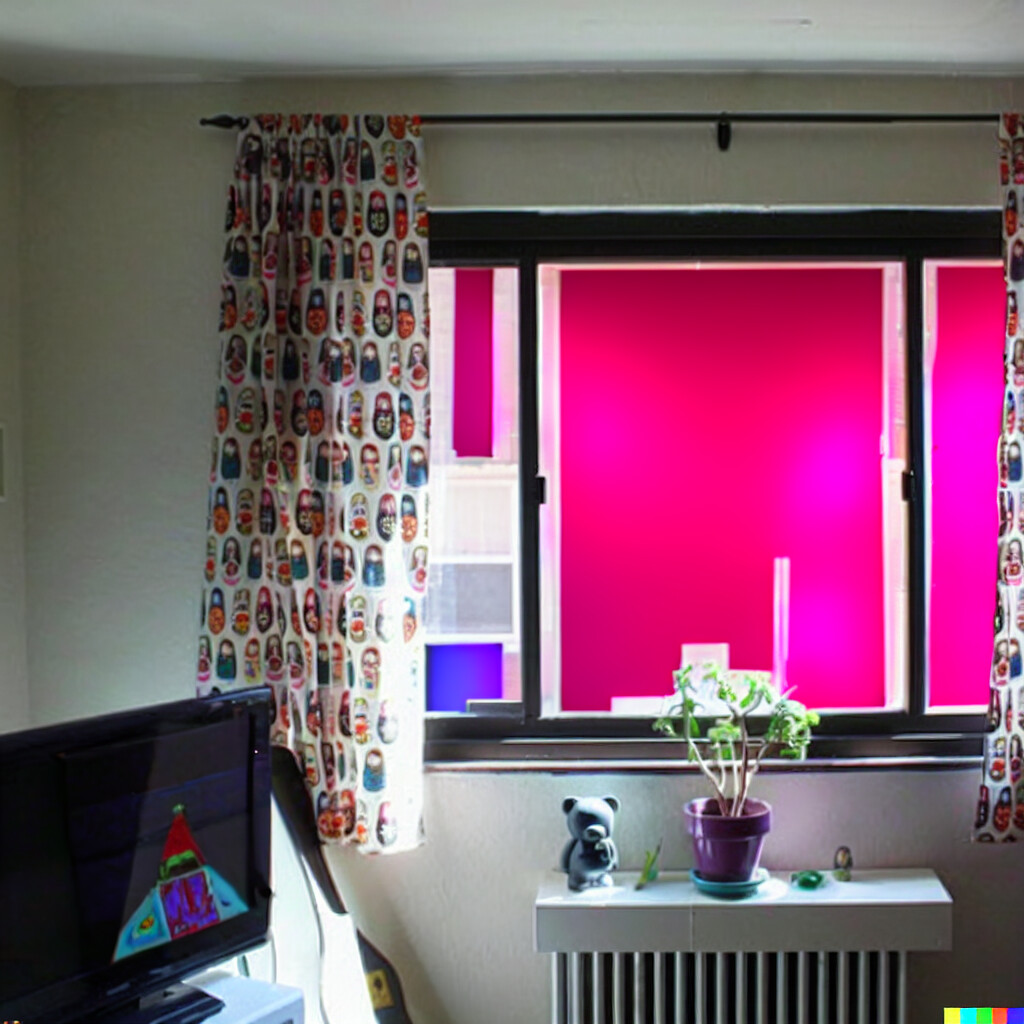}&
\includegraphics[width=0.14\textwidth]{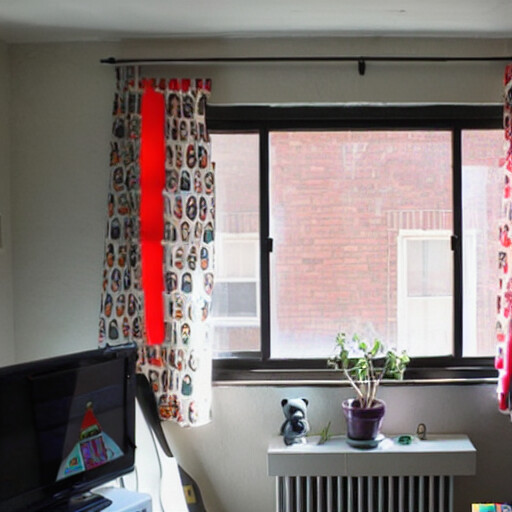}&
\includegraphics[width=0.14\textwidth]{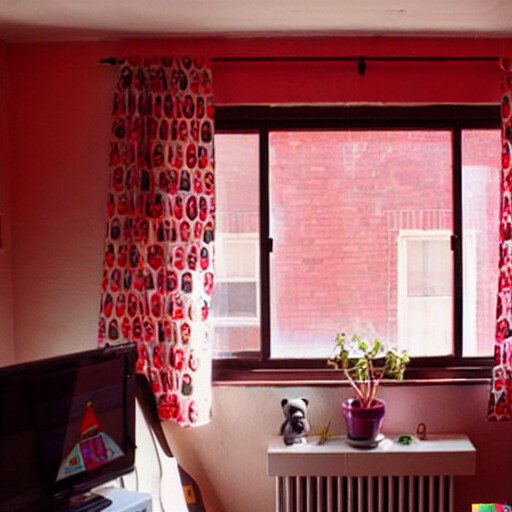}&
\includegraphics[width=0.14\textwidth]{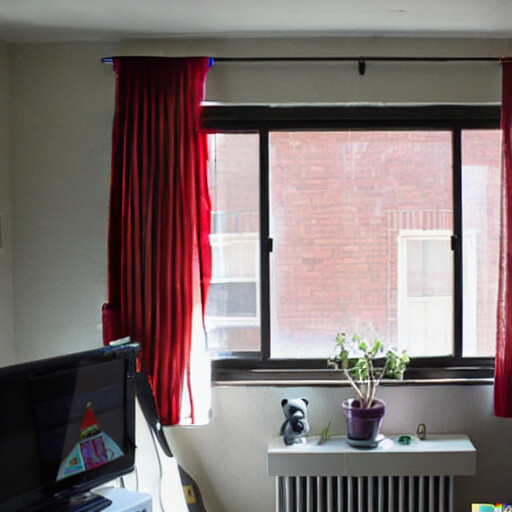}&
\shortstack{
\includegraphics[width=0.06\textwidth]{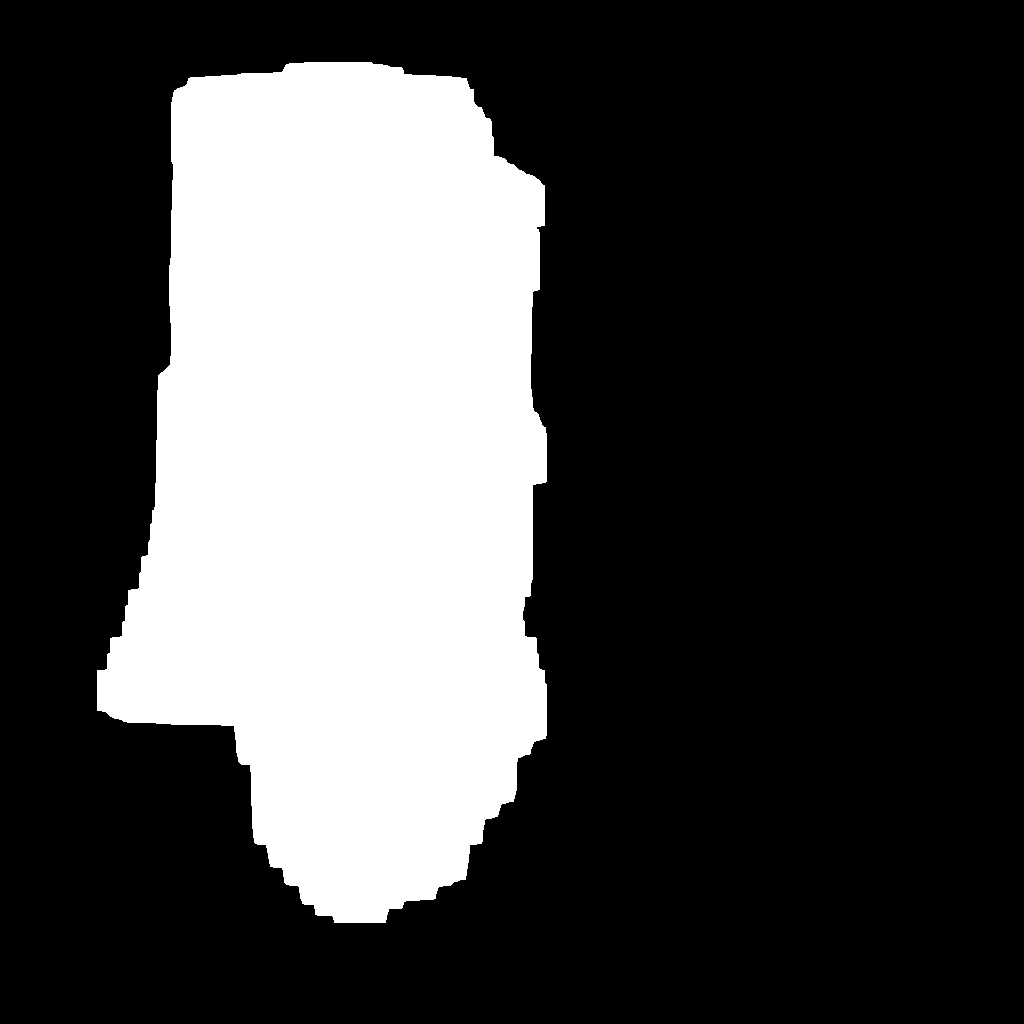} \\
\texttt{[MASK]}}\\

   \multicolumn{7}{c}{(c) Edit instruction: \emph{``Let the curtains be plain red.''}} \\
\includegraphics[width=0.14\textwidth]{figure/Single/182496/182496-output1.jpg}&
\includegraphics[width=0.14\textwidth]{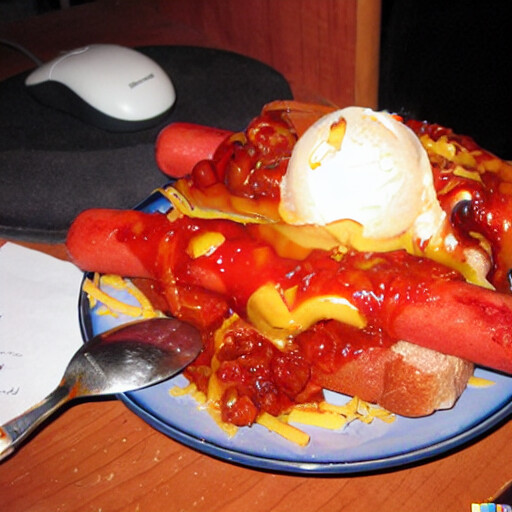}&
\includegraphics[width=0.14\textwidth]{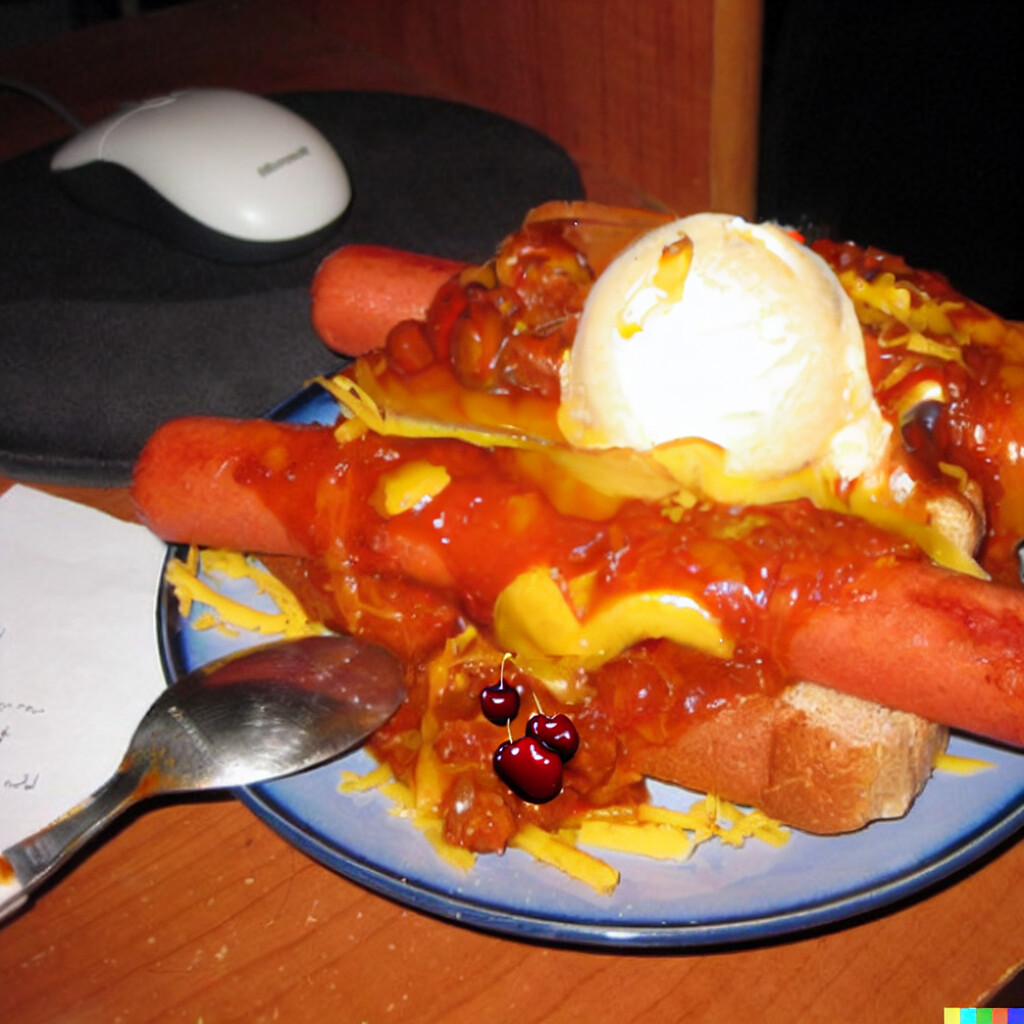}&
\includegraphics[width=0.14\textwidth]{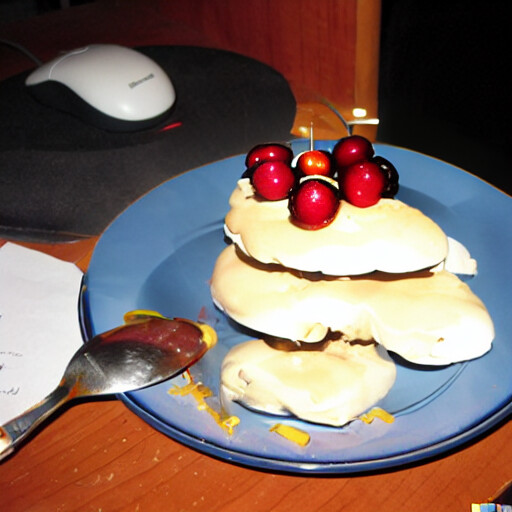}&
\includegraphics[width=0.14\textwidth]{figure/Single/182496/182496_1_foi.jpg}&
\includegraphics[width=0.14\textwidth]{figure/Single/182496/182496_1_camedit.jpg}&
\shortstack{
\includegraphics[width=0.06\textwidth]{figure/Single/182496/182496-output1_mask_0_0.jpg} \\
\texttt{[MASK]}}\\

   \multicolumn{7}{c}{(d) Edit instruction: \emph{``Add a cherry on top.''}} \\
\includegraphics[width=0.14\textwidth]{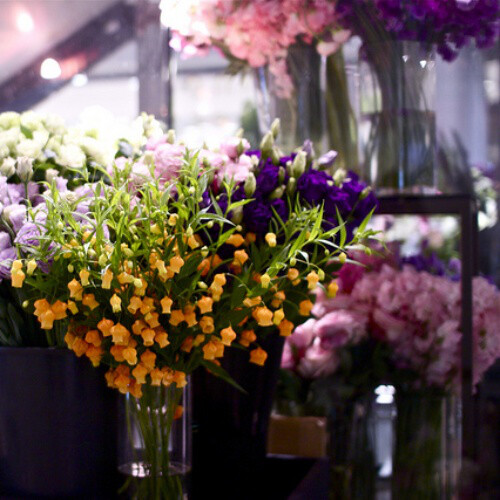}&
\includegraphics[width=0.14\textwidth]{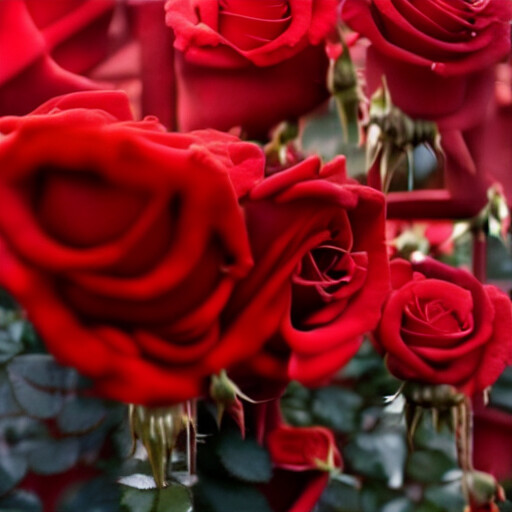}&
\includegraphics[width=0.14\textwidth]{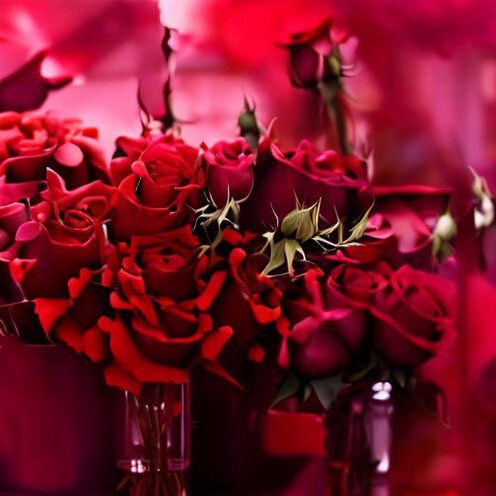}&
\includegraphics[width=0.14\textwidth]{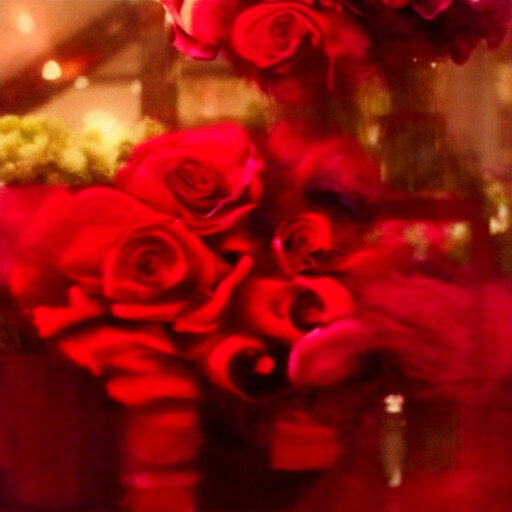}&
\includegraphics[width=0.14\textwidth]{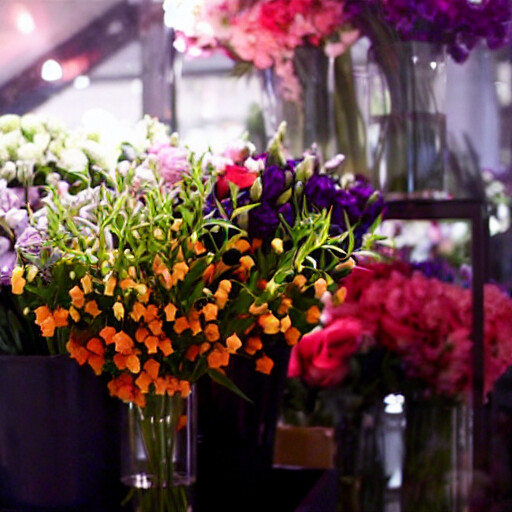}&
\includegraphics[width=0.14\textwidth]{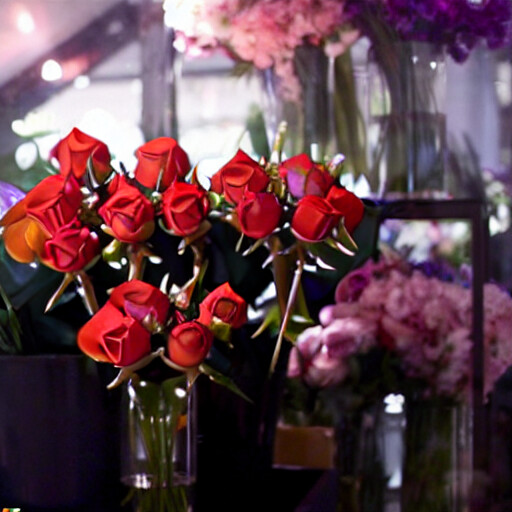}&
\shortstack{
\includegraphics[width=0.06\textwidth]{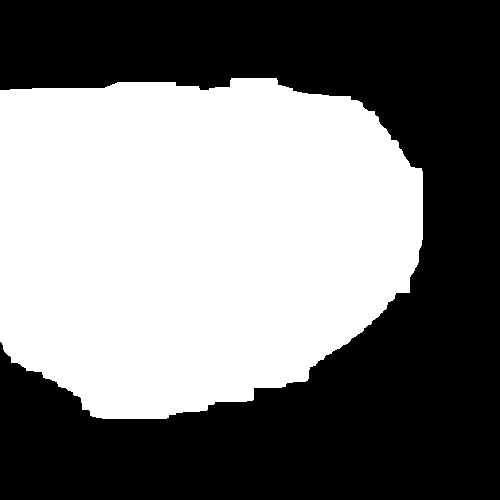} \\
\texttt{[MASK]}}\\

   \multicolumn{7}{c}{(e) Edit instruction: \emph{``Make the flowers into red roses.''}} \\
\includegraphics[width=0.14\textwidth]{figure/Single/360449/360449-input.jpg}&
\includegraphics[width=0.14\textwidth]{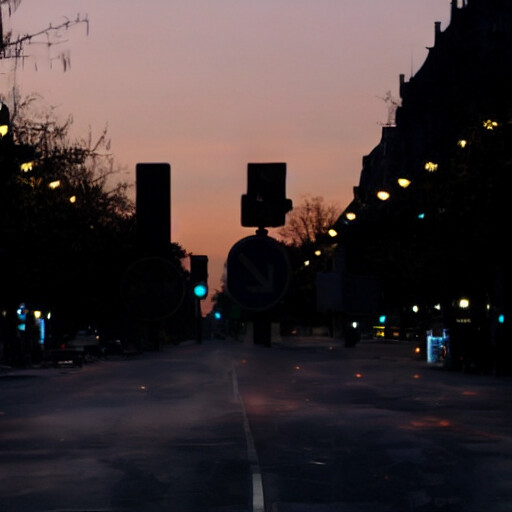}&
\includegraphics[width=0.14\textwidth]{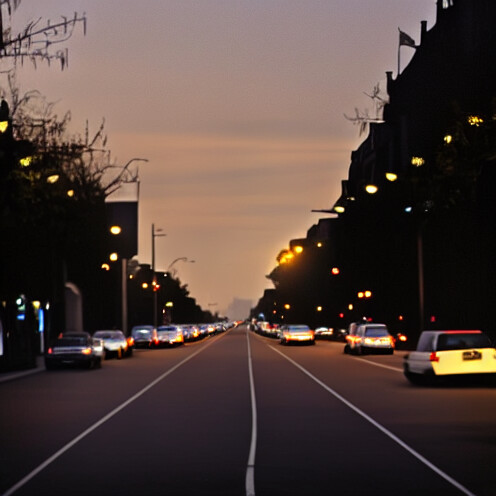}&
\includegraphics[width=0.14\textwidth]{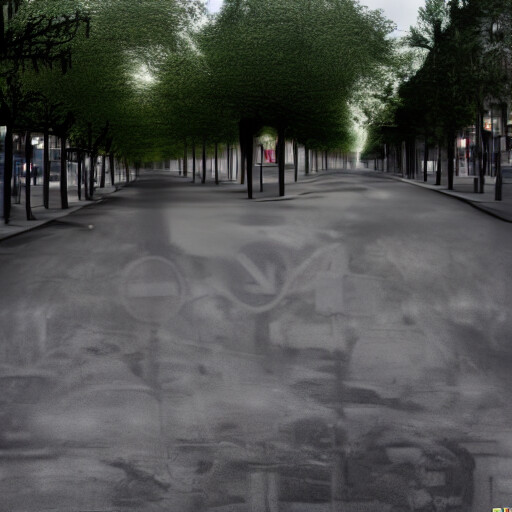}&
\includegraphics[width=0.14\textwidth]{figure/Single/360449/360449_0_foi.jpg}&
\includegraphics[width=0.14\textwidth]{figure/Single/360449/360449_0_camedit.jpg}&
\shortstack{
\includegraphics[width=0.06\textwidth]{figure/Single/360449/360449-input_mask_0_0.jpg} \\
\texttt{[MASK]}}\\

   \multicolumn{7}{c}{(f) Edit instruction: \emph{``Make the street empty.''}} \\
\includegraphics[width=0.14\textwidth]{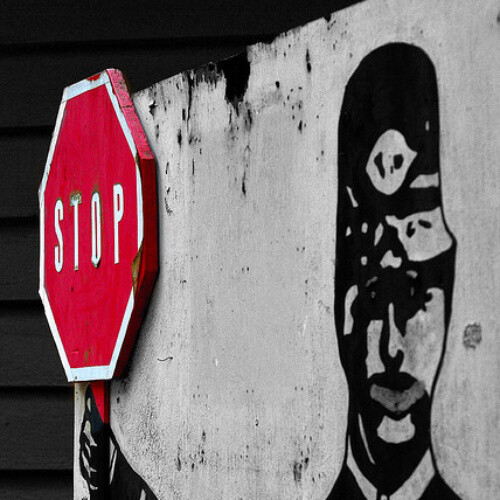}&
\includegraphics[width=0.14\textwidth]{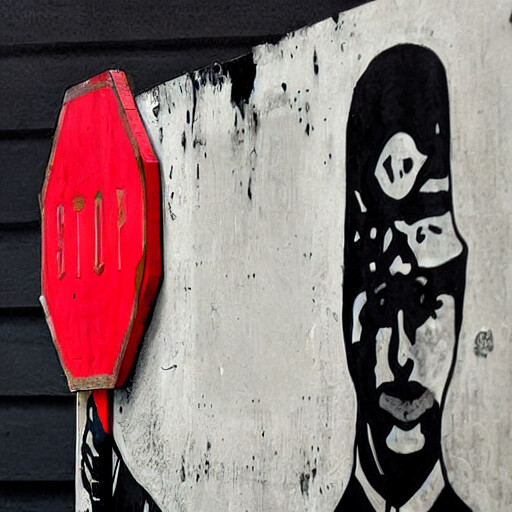}&
\includegraphics[width=0.14\textwidth]{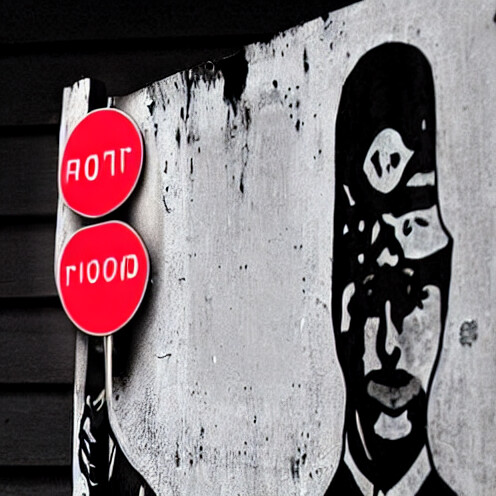}&
\includegraphics[width=0.14\textwidth]{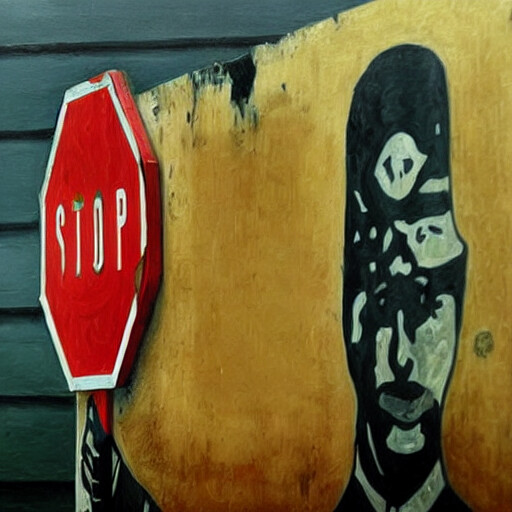}&
\includegraphics[width=0.14\textwidth]{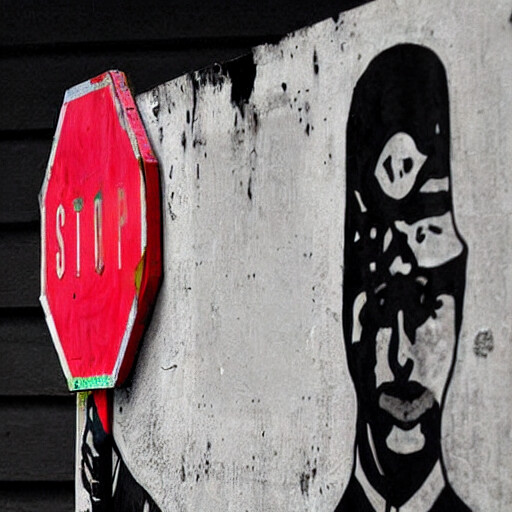}&
\includegraphics[width=0.14\textwidth]{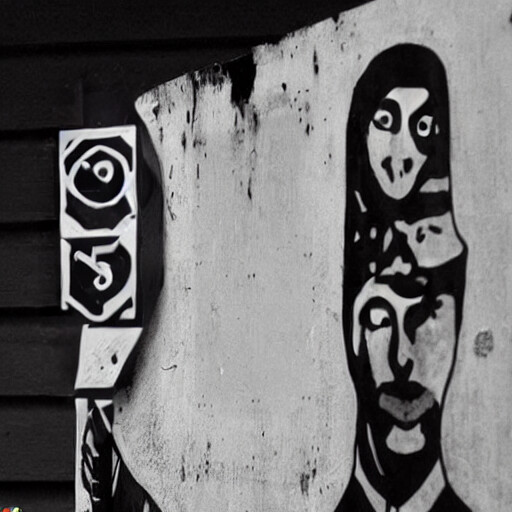}&
\shortstack{
\includegraphics[width=0.06\textwidth]{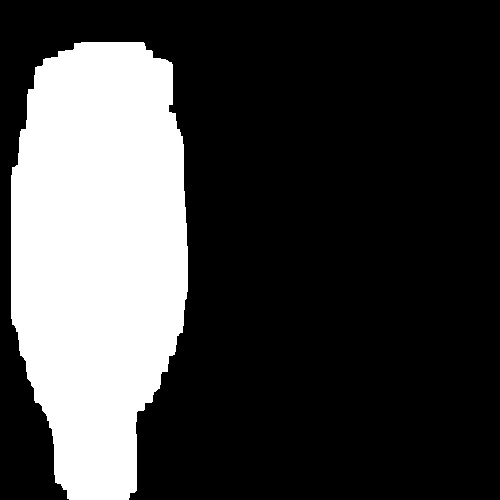} \\
\texttt{[MASK]}
}\\

   \multicolumn{7}{c}{(g) Edit instruction: \emph{``Replace the stop sign with a painting.''}} \\
\includegraphics[width=0.14\textwidth]{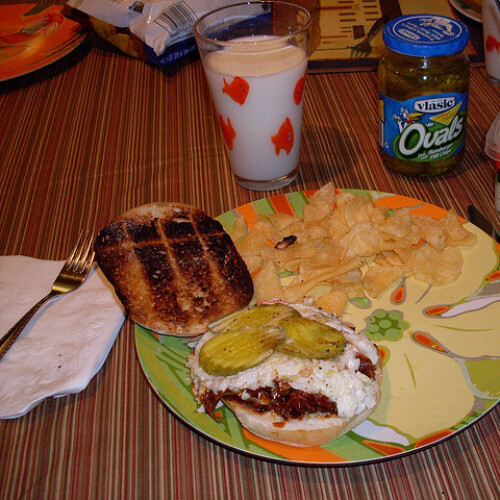}&
\includegraphics[width=0.14\textwidth]{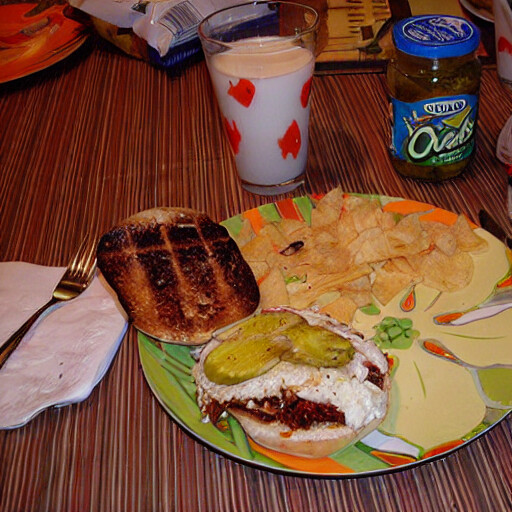}&
\includegraphics[width=0.14\textwidth]{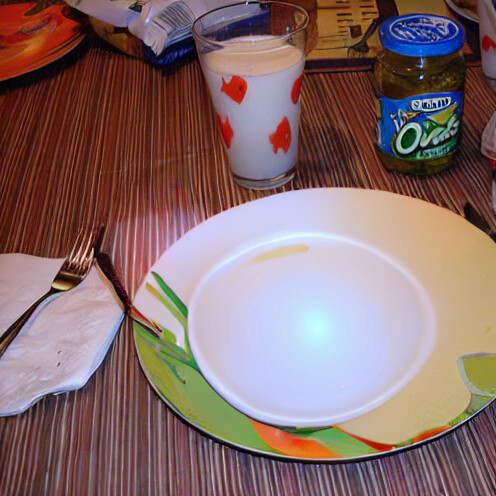}&
\includegraphics[width=0.14\textwidth]{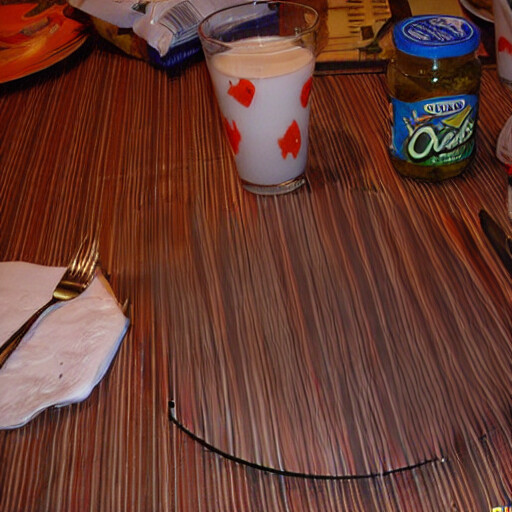}&
\includegraphics[width=0.14\textwidth]{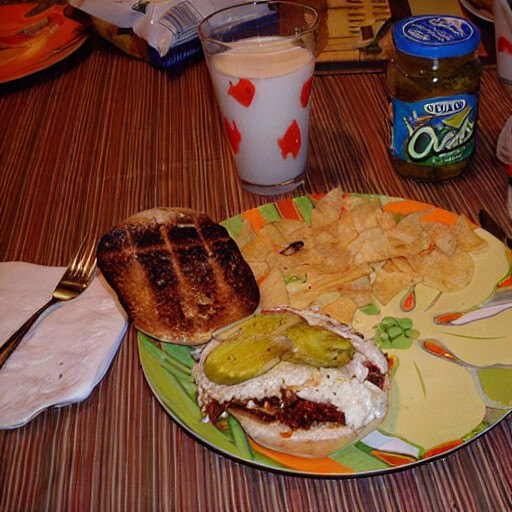}&
\includegraphics[width=0.14\textwidth]{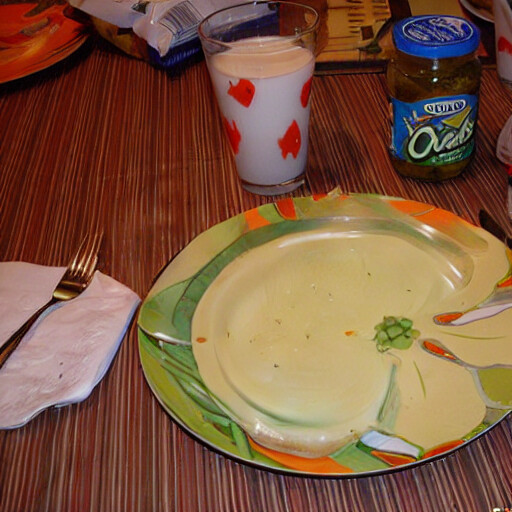}&
\shortstack{
\includegraphics[width=0.06\textwidth]{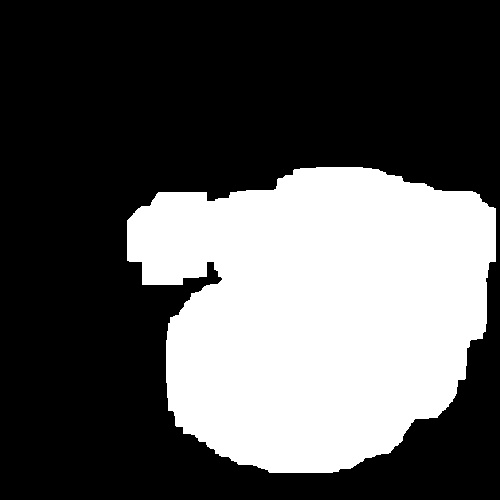} \\
\texttt{[MASK]}
}
 \\

   \multicolumn{7}{c}{(h) Edit instruction: \emph{``Remove all the food from the plate.''}} \\ 
    \end{tabular}}
    \vspace{-3mm}
    \caption{\textbf{Qualitative comparisons for single instruction task}}
    \label{fig:appendix_qualitative_single}
    \vspace{-1.5em}
\end{figure*}

\begin{figure*}[t]
    \setlength\tabcolsep{2.5pt}
    \centering
    \resizebox{\linewidth}{!}{%
    \normalsize
    \begin{tabular}{ccccccc}
        Input Image & {\footnotesize \shortstack{IP2P\\\cite{brooks2023instructpix2pix}}} & {\footnotesize \shortstack{MGIE\\\cite{fu2024guiding}}} & {\footnotesize \shortstack{SmartEdit\\\cite{huang2023smartedit}}} & {\footnotesize \shortstack{FoI\\\cite{guo2024focus}}} & \multicolumn{2}{c}{\name~\textbf{(Ours)}} \\

        \includegraphics[width=0.14\textwidth]{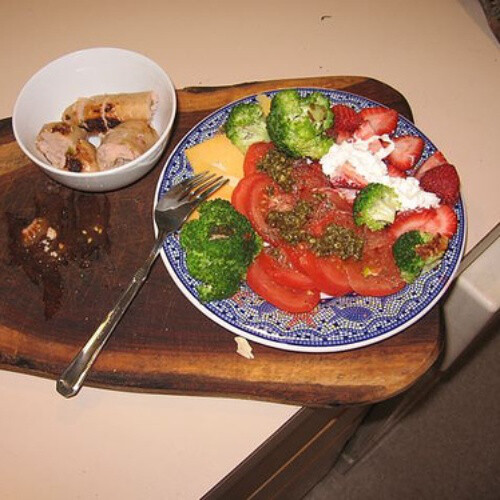}&
        \includegraphics[width=0.14\textwidth]{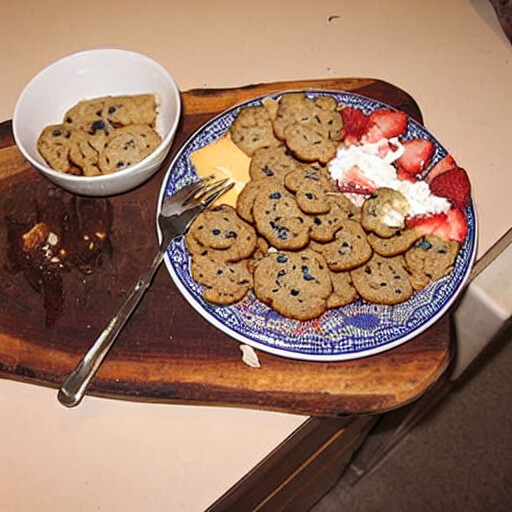}&
        \includegraphics[width=0.14\textwidth]{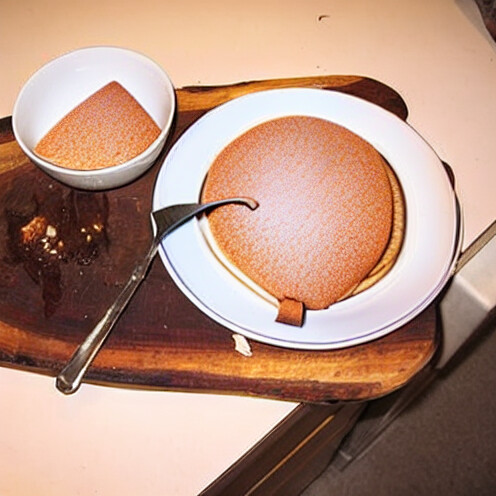}&
        \includegraphics[width=0.14\textwidth]{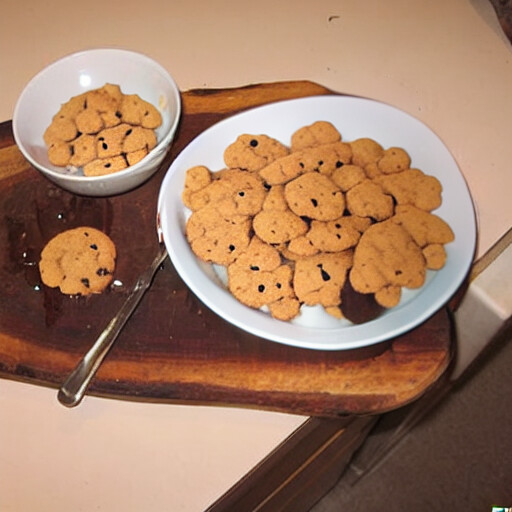}&
        \includegraphics[width=0.14\textwidth]{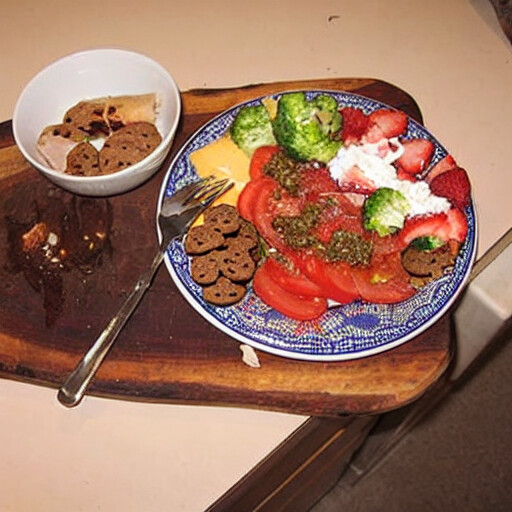}&
        \includegraphics[width=0.14\textwidth]{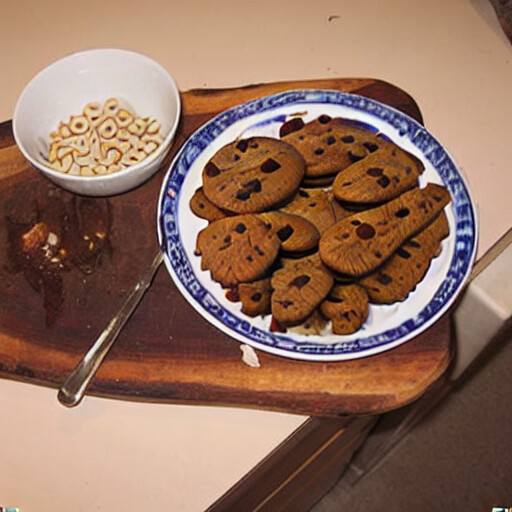}&
        \shortstack{
        \includegraphics[width=0.06\textwidth]{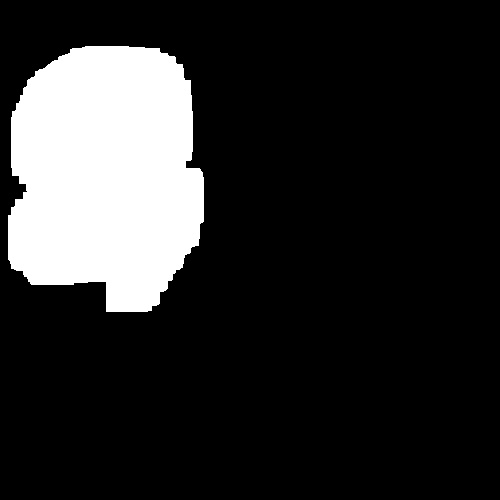}\quad\includegraphics[width=0.06\textwidth]{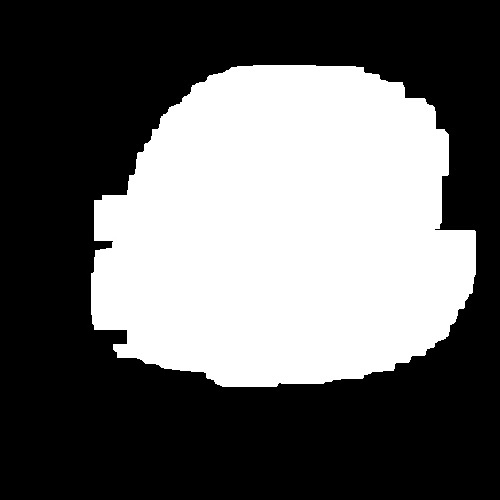}\\
        \texttt{[MASK]}~\texttt{[MASK]}}
        \\
        \multicolumn{7}{c}{(a) Edit instruction: \emph{``Turn the meat into cereal, and fill the plate with cookies.''}} \\

        \includegraphics[width=0.14\textwidth]{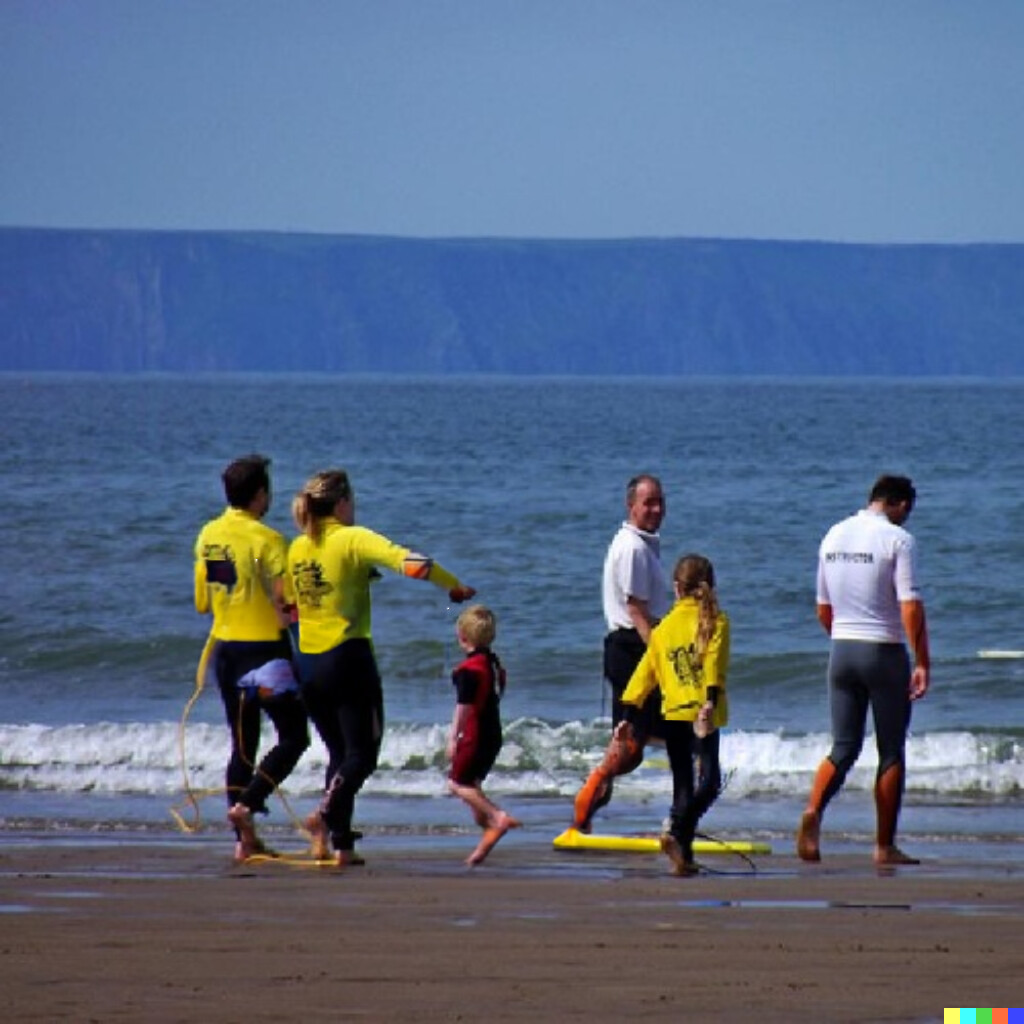}&
        \includegraphics[width=0.14\textwidth]{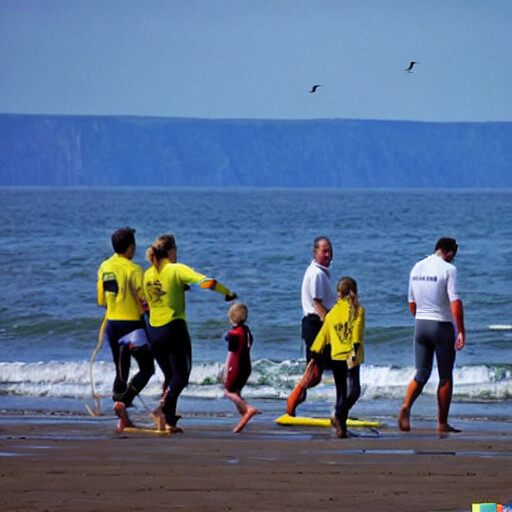}&
        \includegraphics[width=0.14\textwidth]{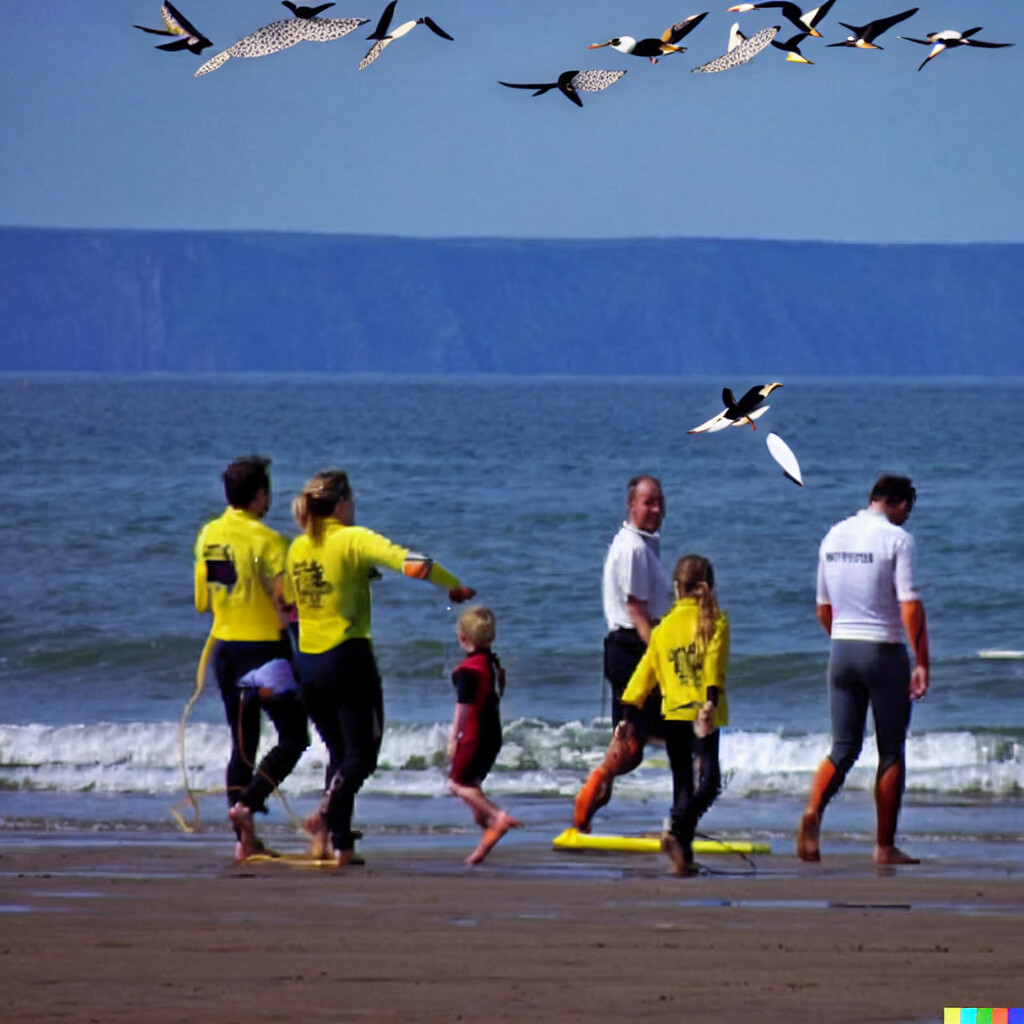}&
        \includegraphics[width=0.14\textwidth]{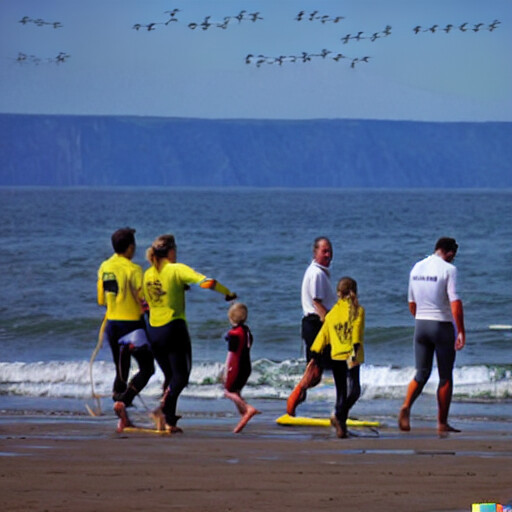}&
        \includegraphics[width=0.14\textwidth]{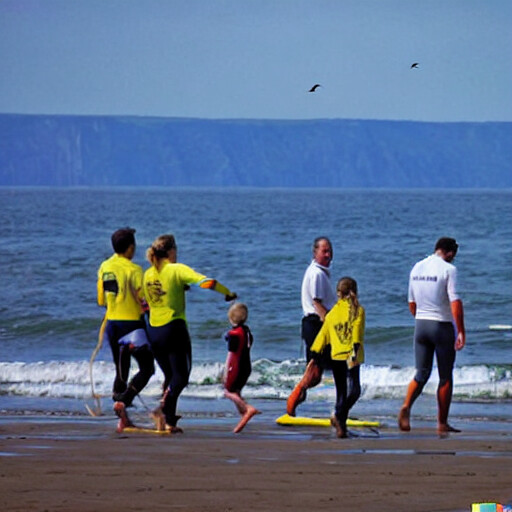}&
        \includegraphics[width=0.14\textwidth]{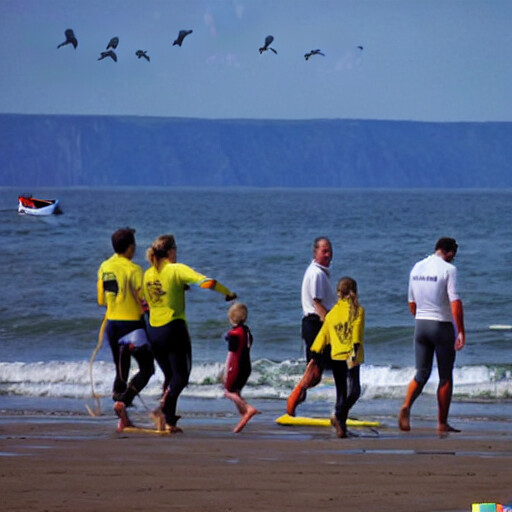}&
        \shortstack{
        \includegraphics[width=0.06\textwidth]{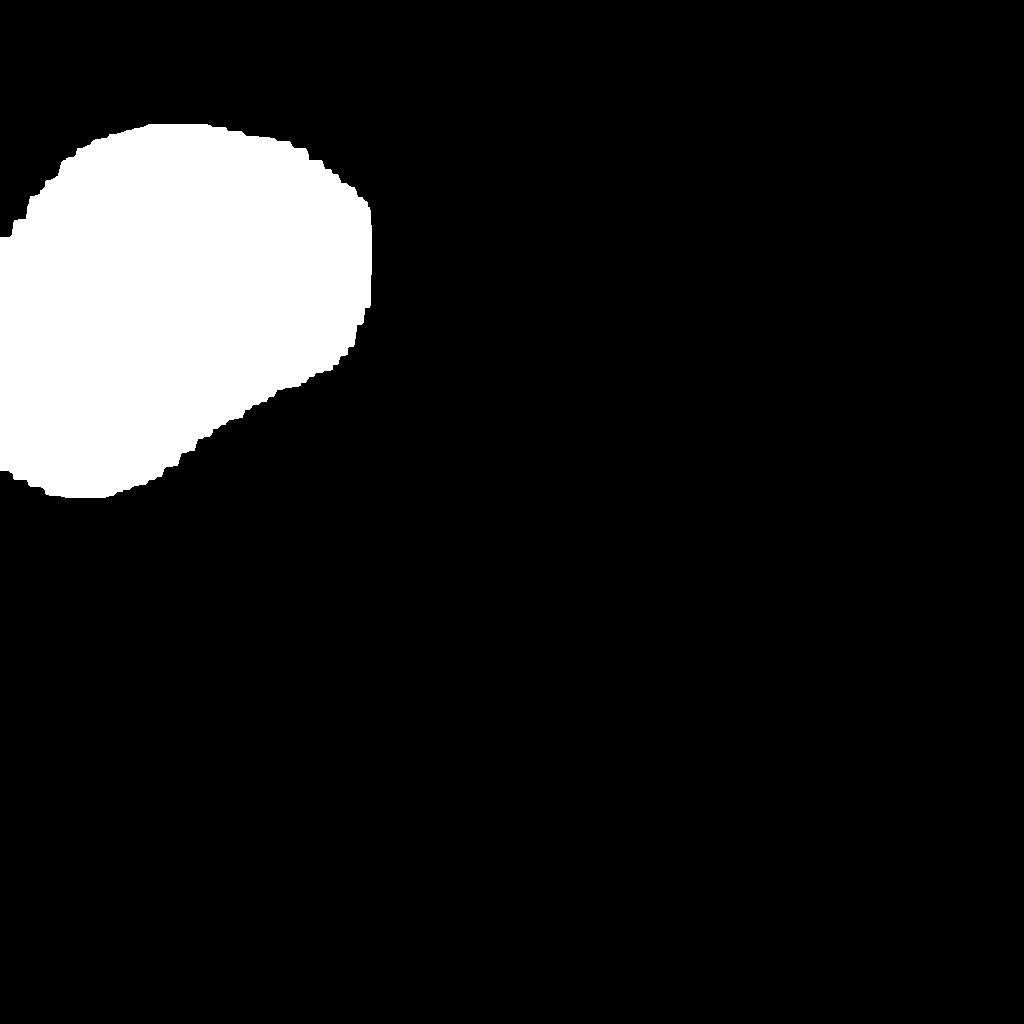}\quad\includegraphics[width=0.06\textwidth]{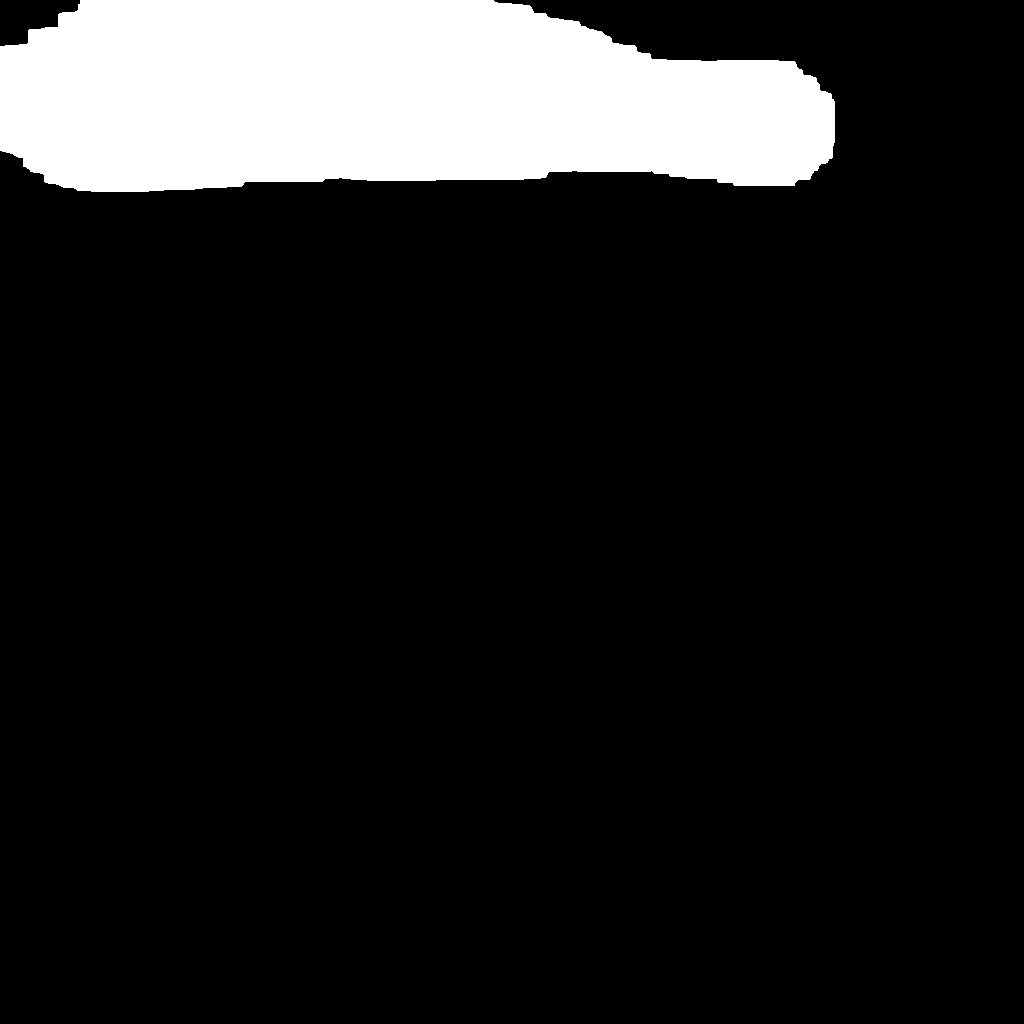}\\
        \texttt{[MASK]}~\texttt{[MASK]}}
        \\
        \multicolumn{7}{c}{(b) Edit instruction: \emph{``Add a boat, and add birds to the sky.''}} \\
        
        \includegraphics[width=0.14\textwidth]{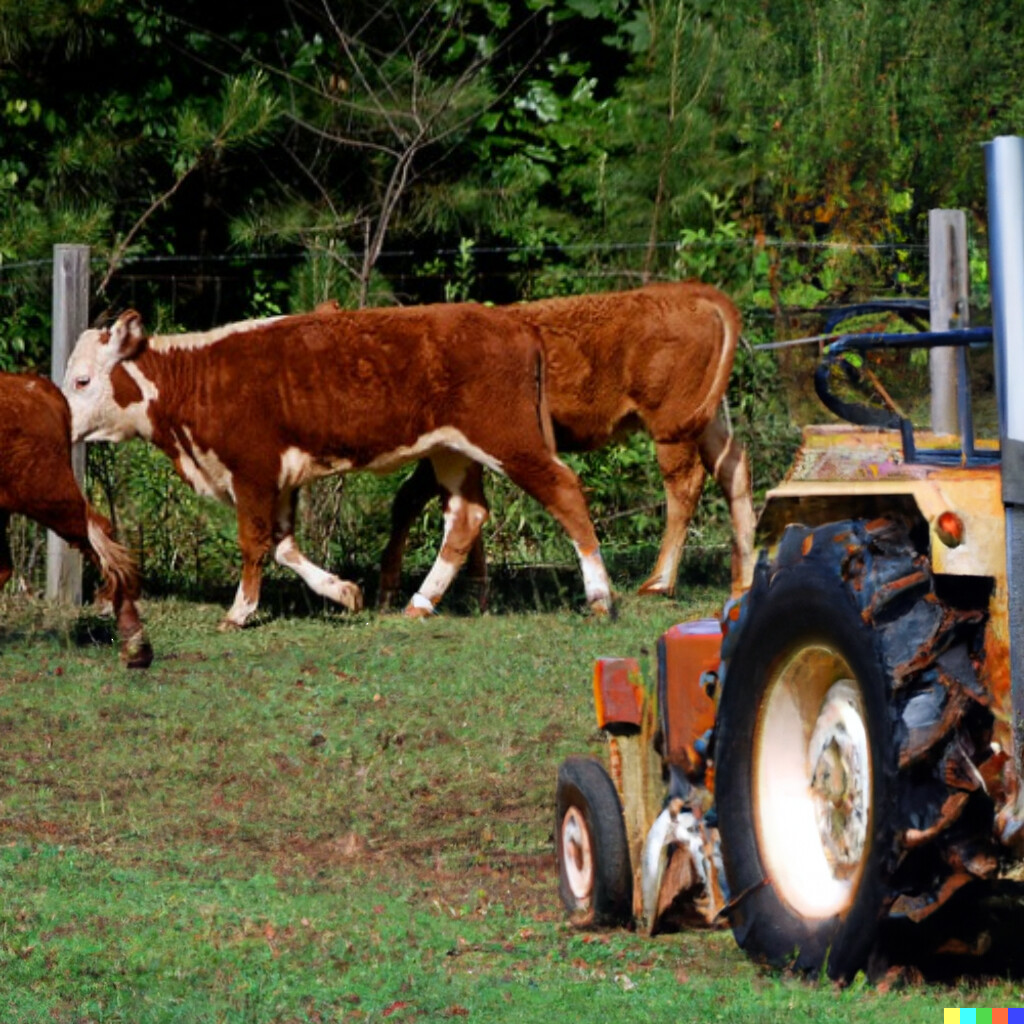}&
        \includegraphics[width=0.14\textwidth]{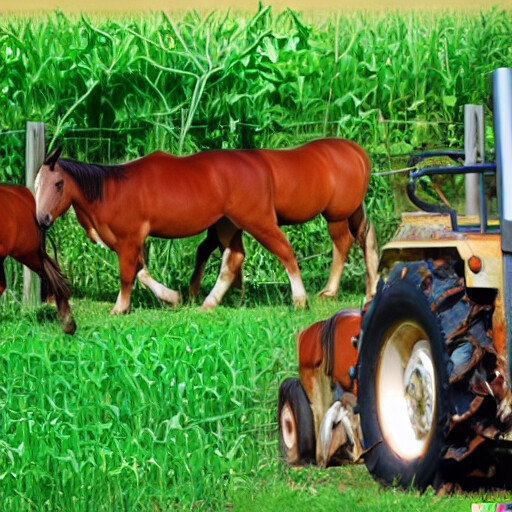}&
        \includegraphics[width=0.14\textwidth]{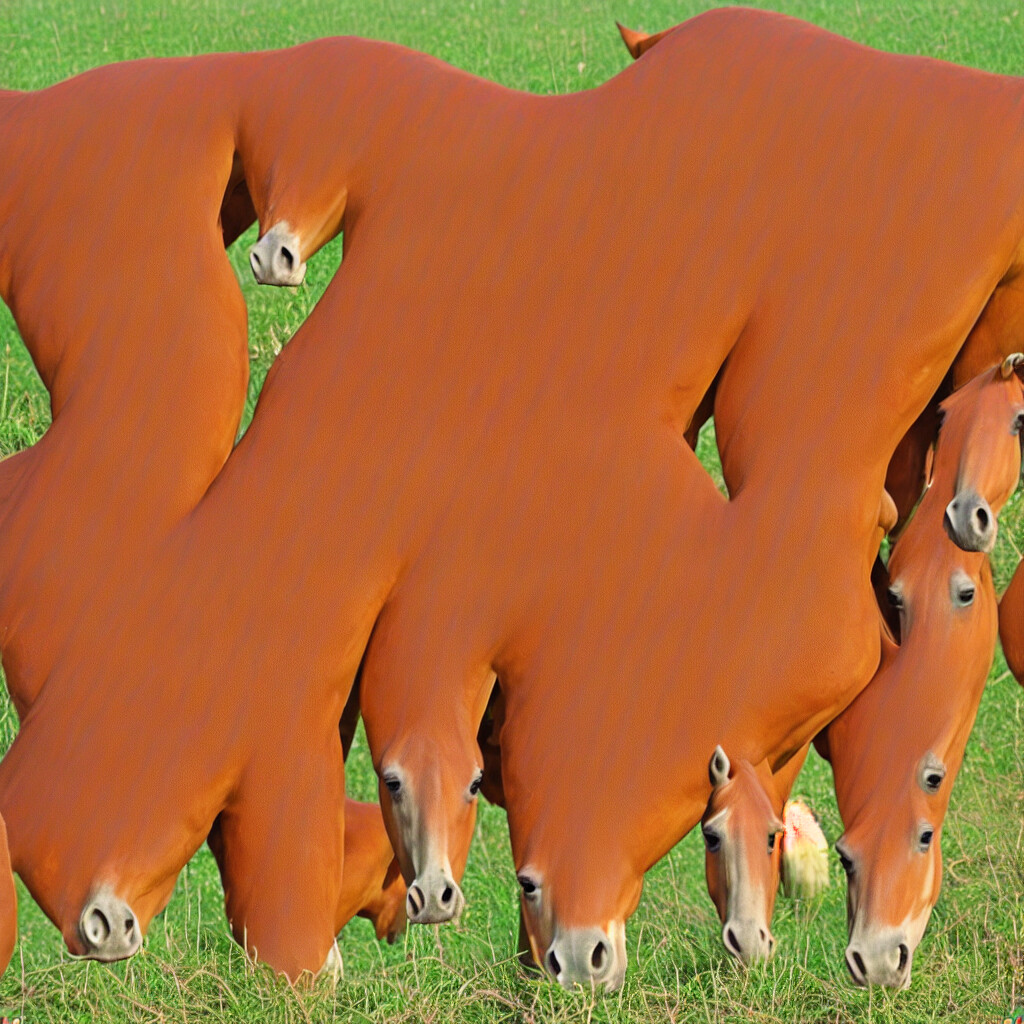}&
        \includegraphics[width=0.14\textwidth]{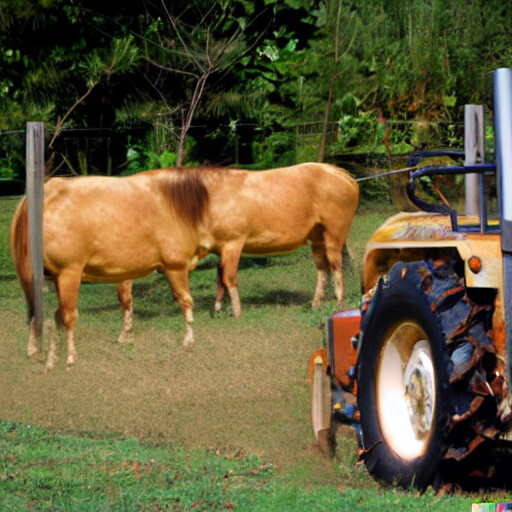}&
        \includegraphics[width=0.14\textwidth]{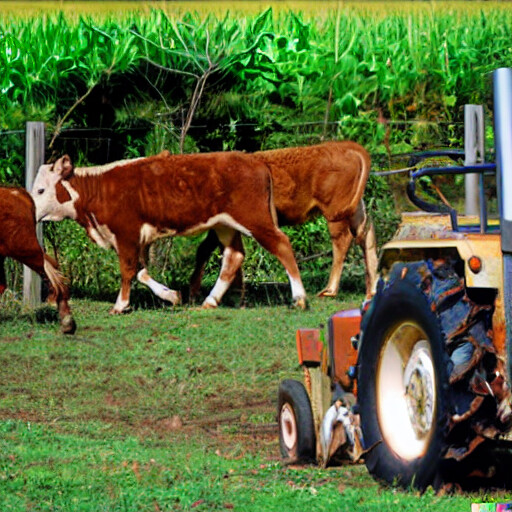}&
        \includegraphics[width=0.14\textwidth]{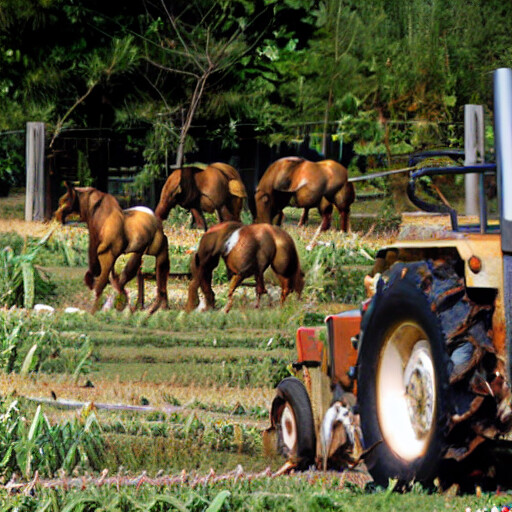}&
        \shortstack{
        \includegraphics[width=0.06\textwidth]{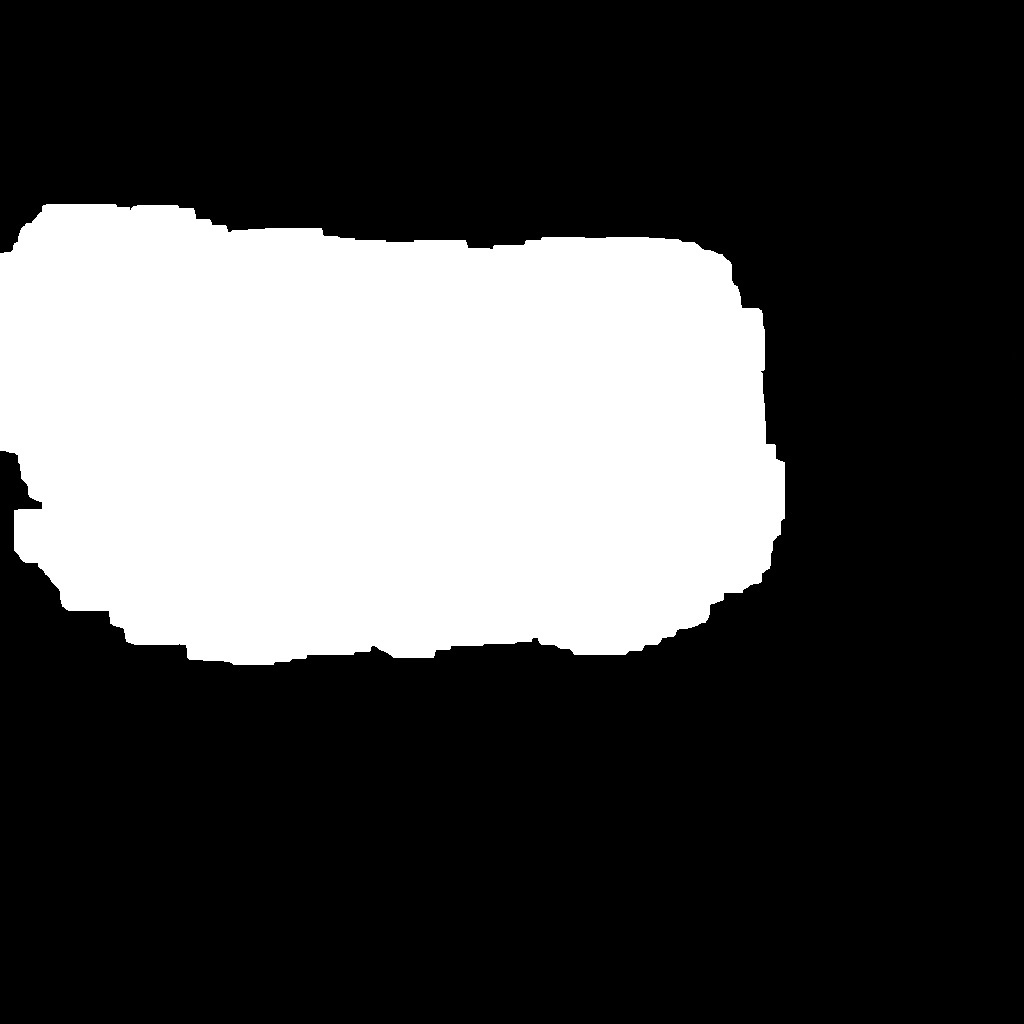}\quad\includegraphics[width=0.06\textwidth]{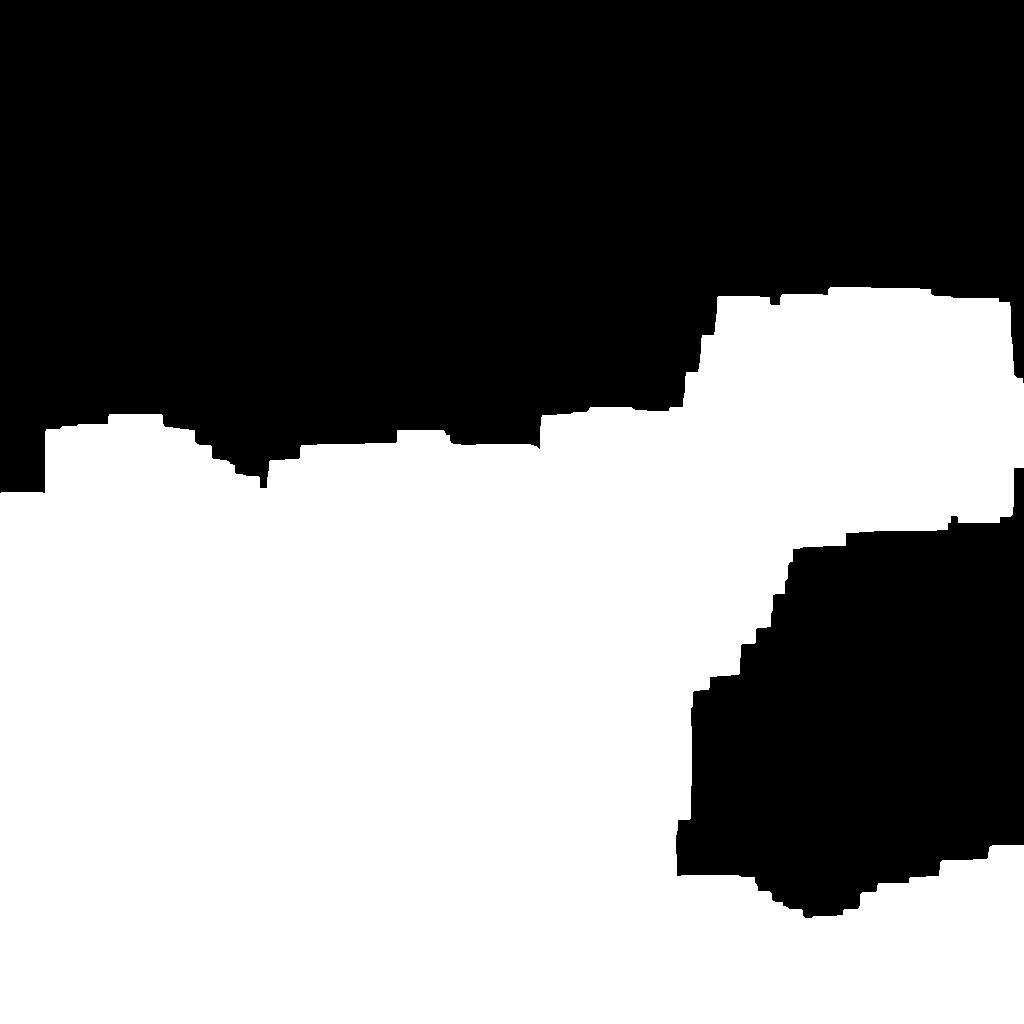}\\
        \texttt{[MASK]}~\texttt{[MASK]}}
        \\
        \multicolumn{7}{c}{(c) Edit instruction: \emph{``Change the cows to horses, and make the field a cornfield.''}} \\

        \includegraphics[width=0.14\textwidth]{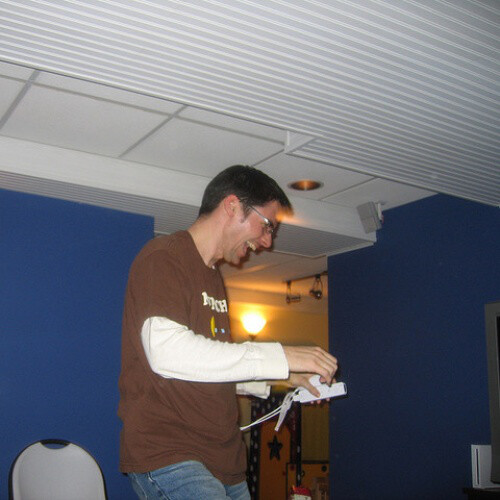}&
        \includegraphics[width=0.14\textwidth]{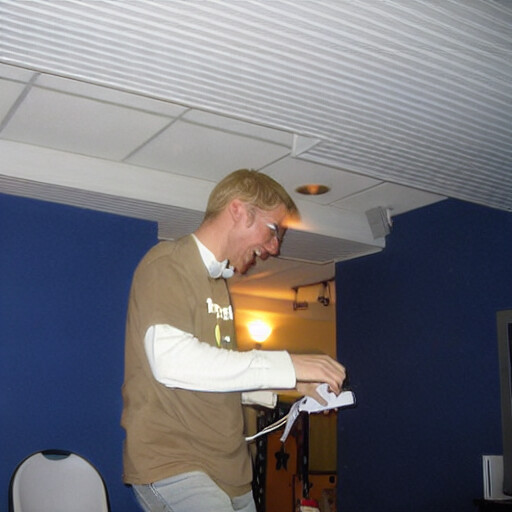}&
        \includegraphics[width=0.14\textwidth]{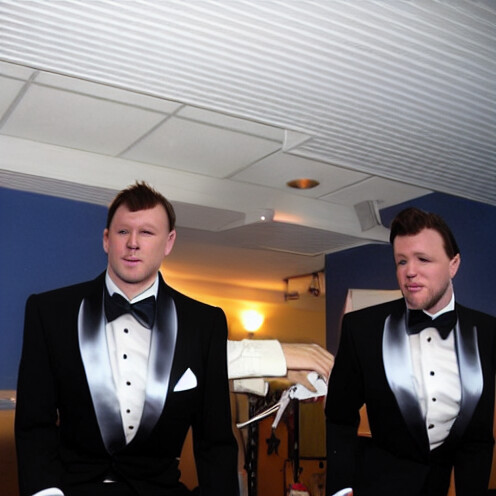}&
        \includegraphics[width=0.14\textwidth]{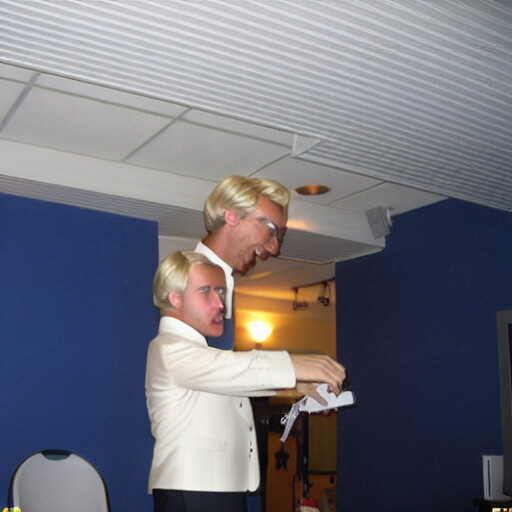}&
        \includegraphics[width=0.14\textwidth]{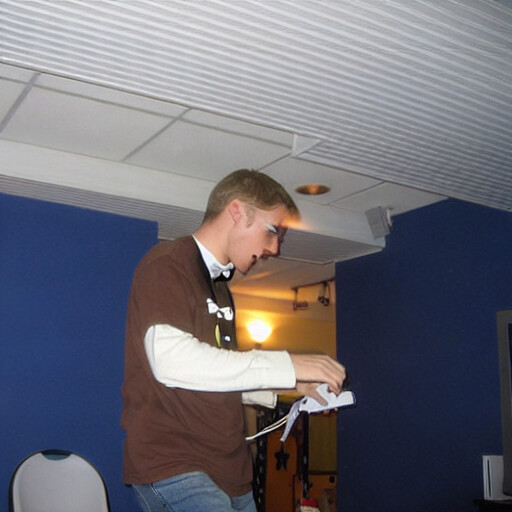}&
        \includegraphics[width=0.14\textwidth]{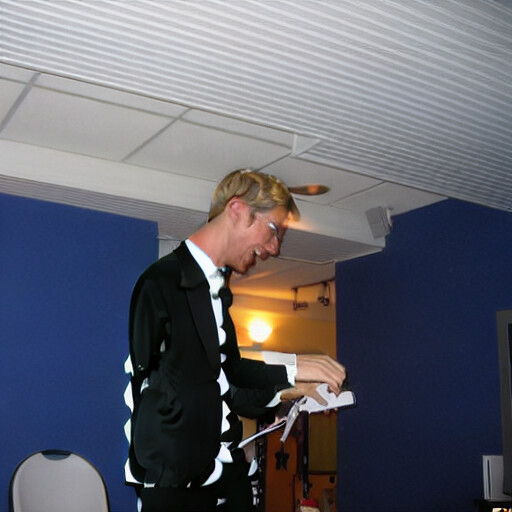}&
        \shortstack{
        \includegraphics[width=0.06\textwidth]{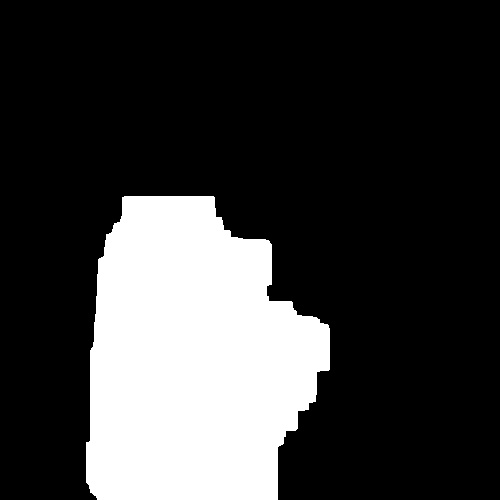}\quad\includegraphics[width=0.06\textwidth]{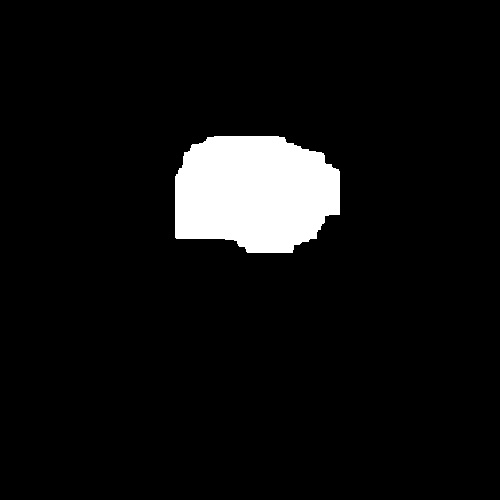}\\
        \texttt{[MASK]}~\texttt{[MASK]}}
        \\
        \multicolumn{7}{c}{(d) Edit instruction: \emph{``Change his outfit into a tuxedo. Furthermore, change his hair to blonde.''}} \\

        \includegraphics[width=0.14\textwidth]{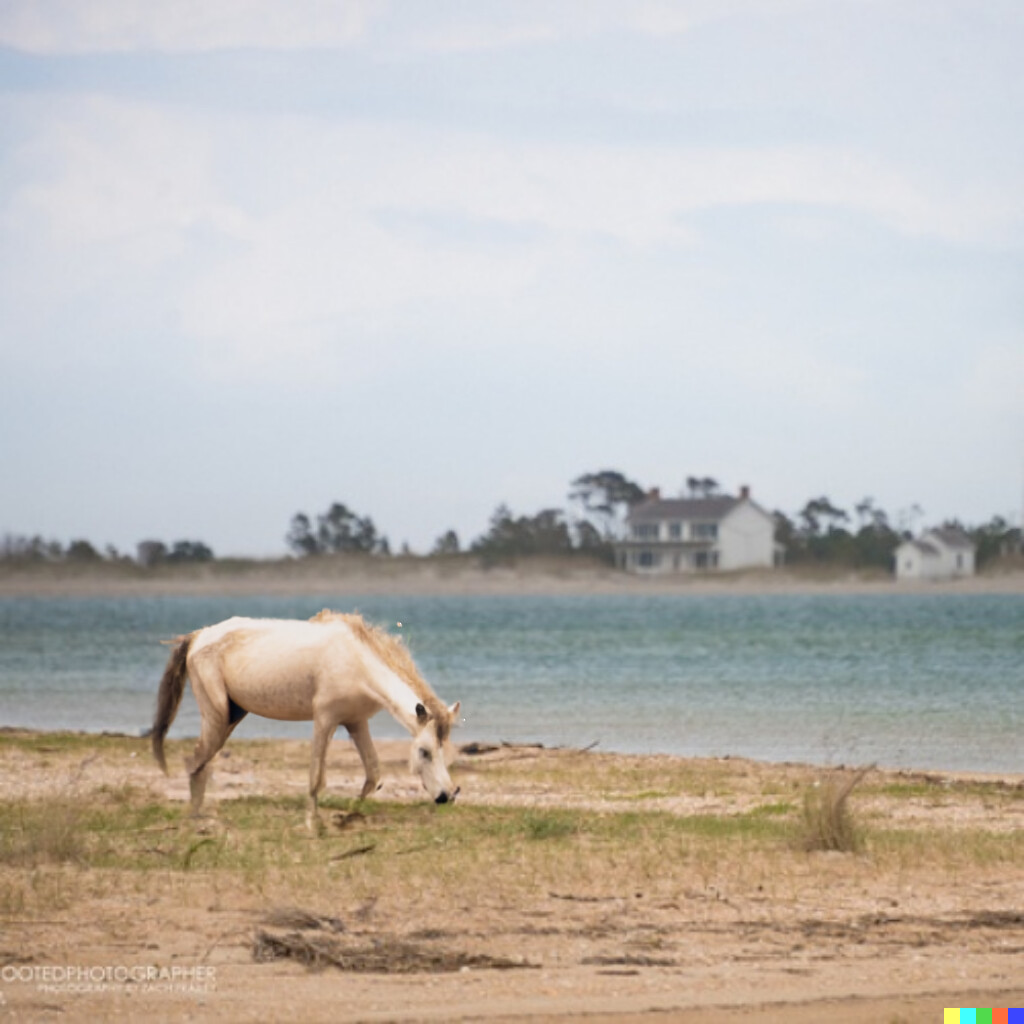}&
        \includegraphics[width=0.14\textwidth]{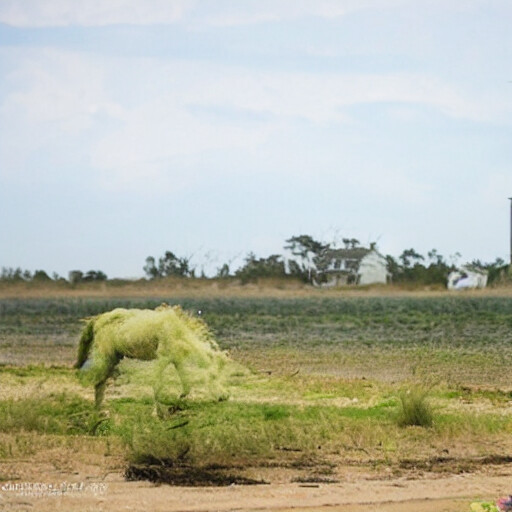}&
        \includegraphics[width=0.14\textwidth]{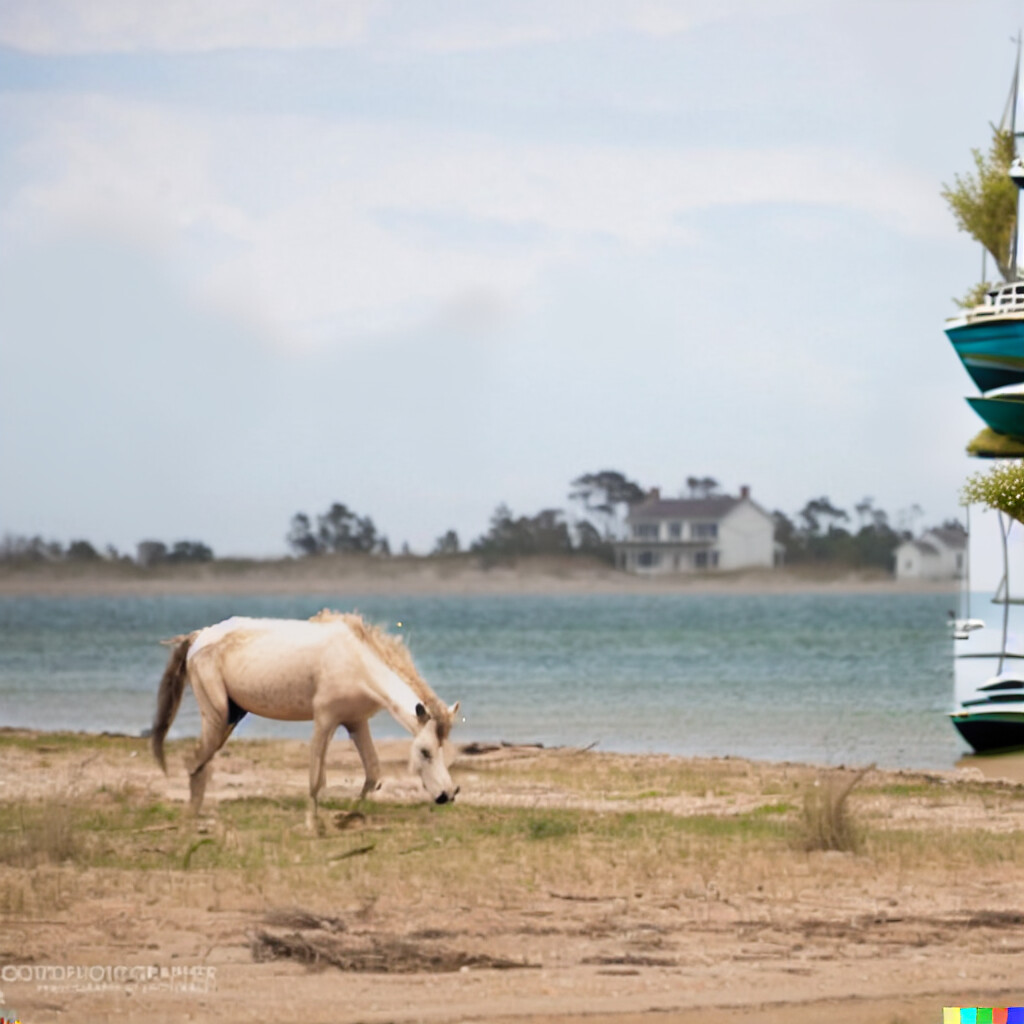}&
        \includegraphics[width=0.14\textwidth]{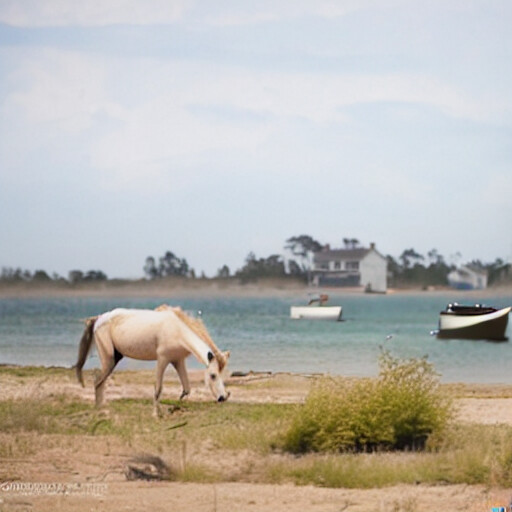}&
        \includegraphics[width=0.14\textwidth]{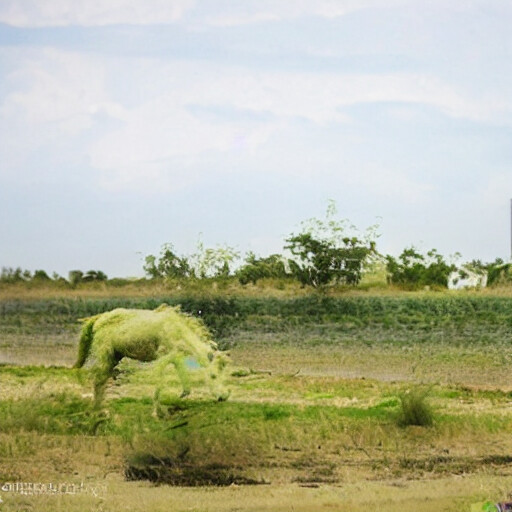}&
        \includegraphics[width=0.14\textwidth]{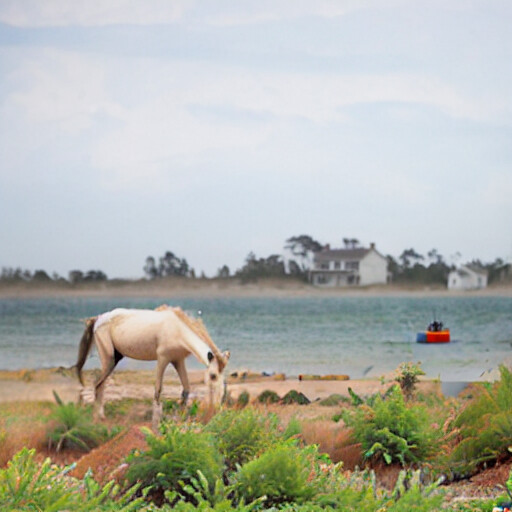}&
        \shortstack{
        \includegraphics[width=0.06\textwidth]{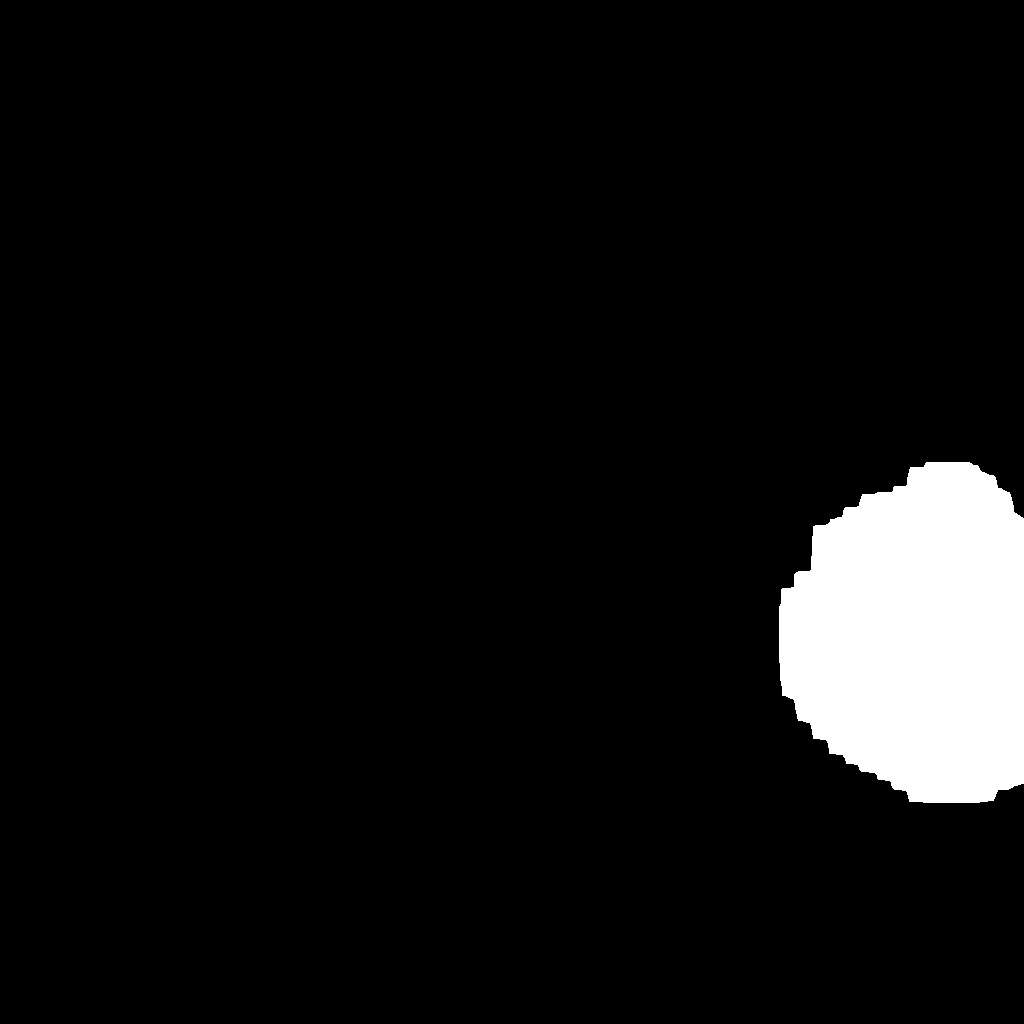 }\quad\includegraphics[width=0.06\textwidth]{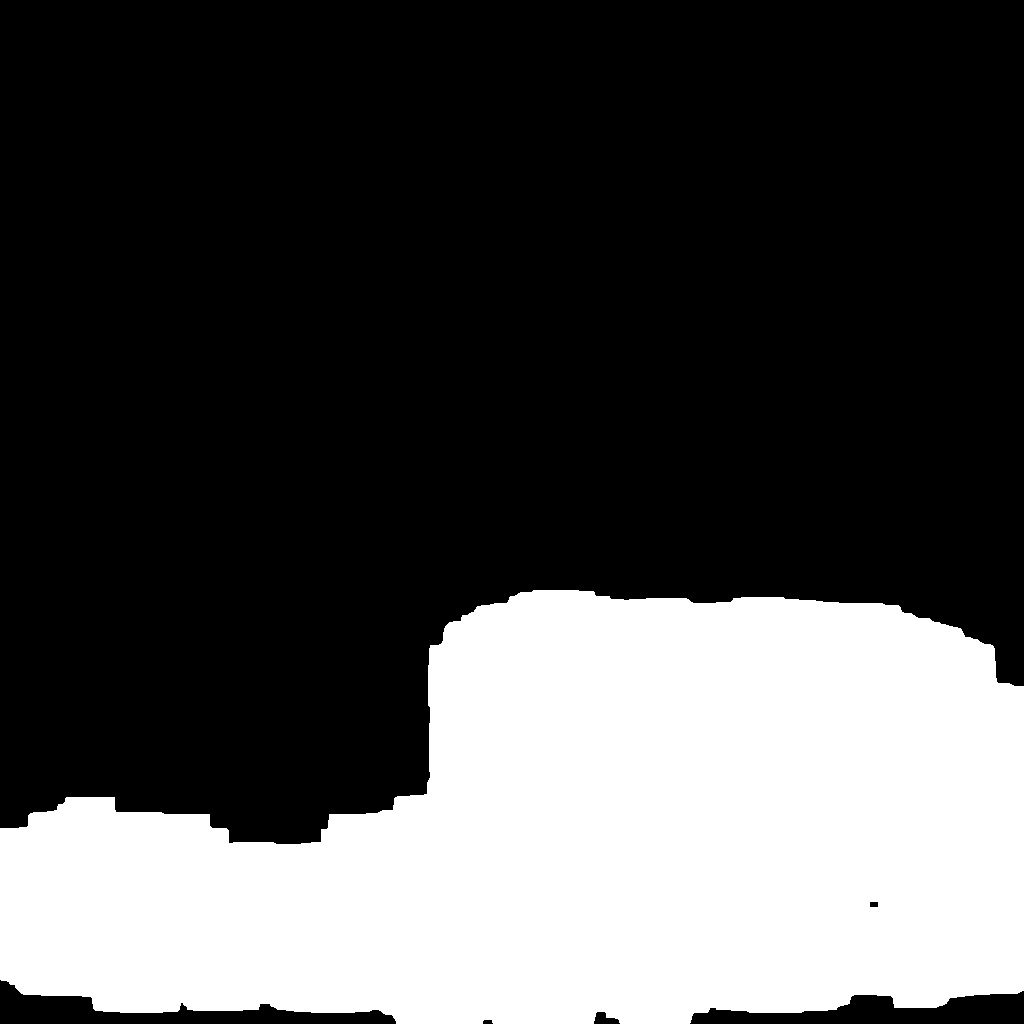}\\
        \texttt{[MASK]}~\texttt{[MASK]}}
        \\
        \multicolumn{7}{c}{(e) Edit instruction: \emph{``Add a boat, and add tall shrubs.''}} \\
        
        \includegraphics[width=0.14\textwidth]{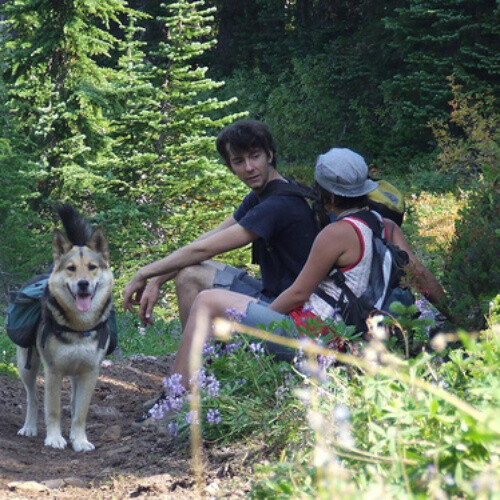}&
        \includegraphics[width=0.14\textwidth]{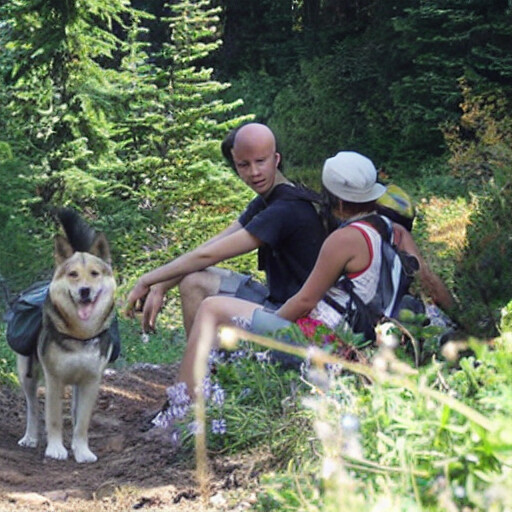}&
        \includegraphics[width=0.14\textwidth]{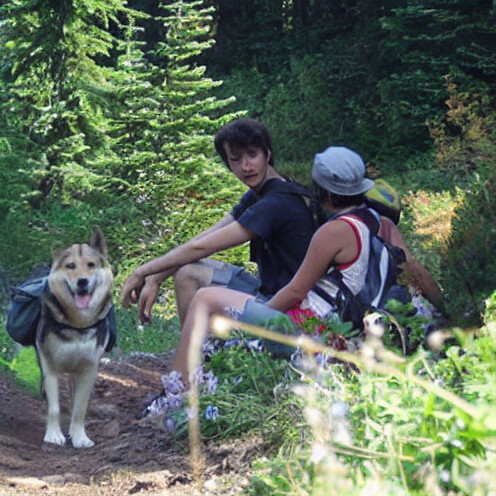}&
        \includegraphics[width=0.14\textwidth]{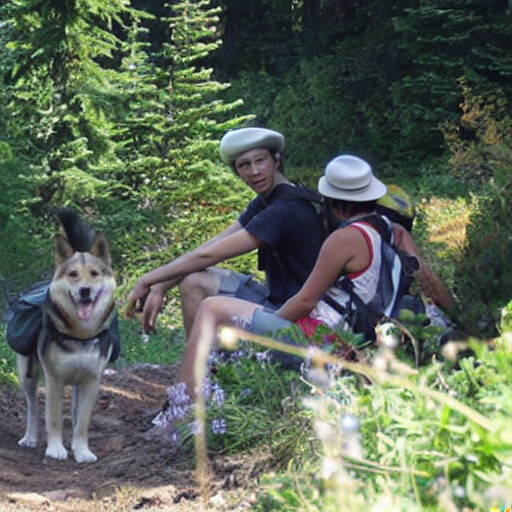}&
        \includegraphics[width=0.14\textwidth]{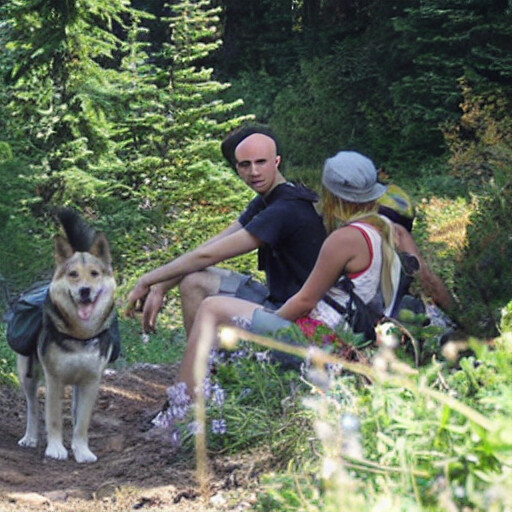}&
        \includegraphics[width=0.14\textwidth]{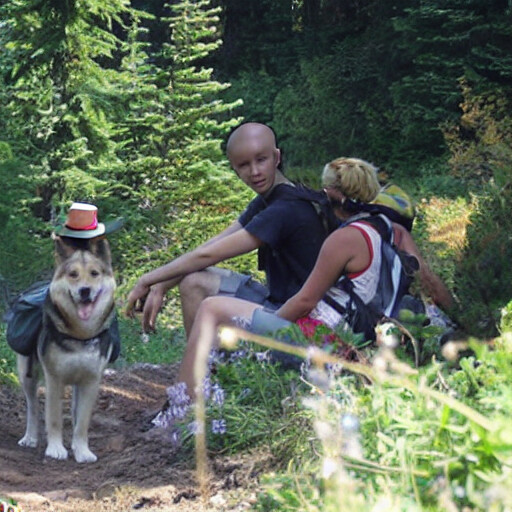}&
        \shortstack{
        \includegraphics[width=0.06\textwidth]{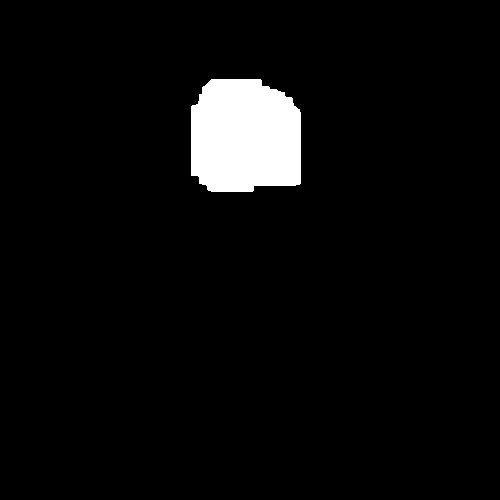}\quad\includegraphics[width=0.06\textwidth]{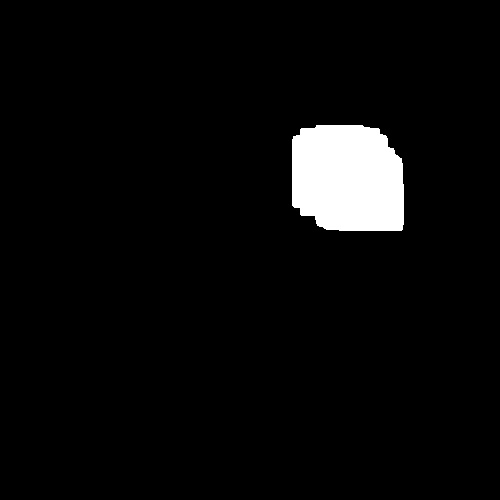}
        \quad\includegraphics[width=0.06\textwidth]{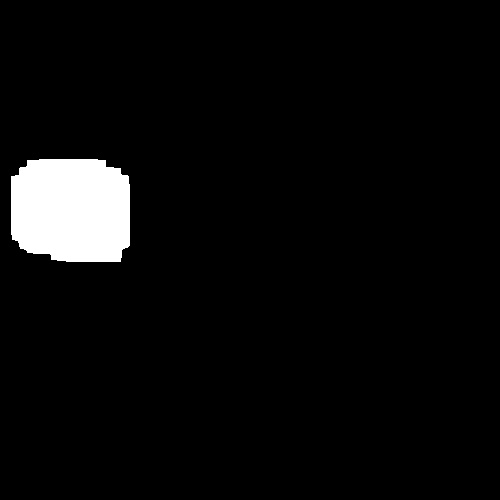}\\
        \texttt{[MASK]}~\texttt{[MASK]}~\texttt{[MASK]}}
        \\
        \multicolumn{7}{c}{(f) Edit instruction: \emph{``What if the man was bald, what if the woman was blonde, and what if the dog had a hat?''}} \\
        
        \includegraphics[width=0.14\textwidth]{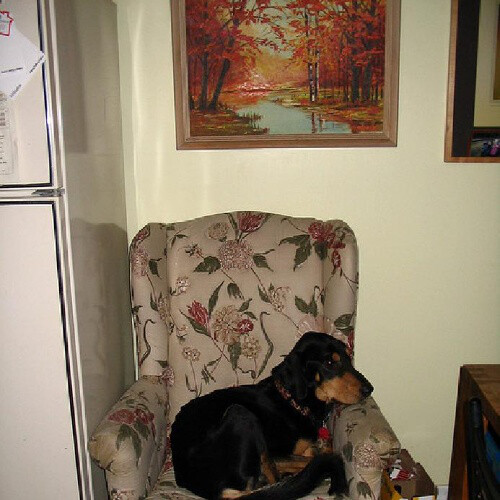}&
        \includegraphics[width=0.14\textwidth]{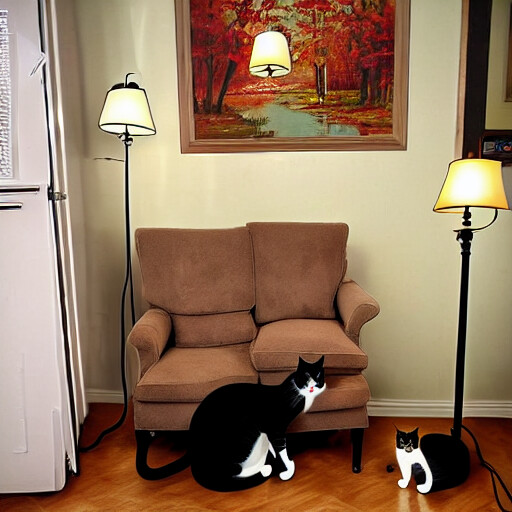}&
        \includegraphics[width=0.14\textwidth]{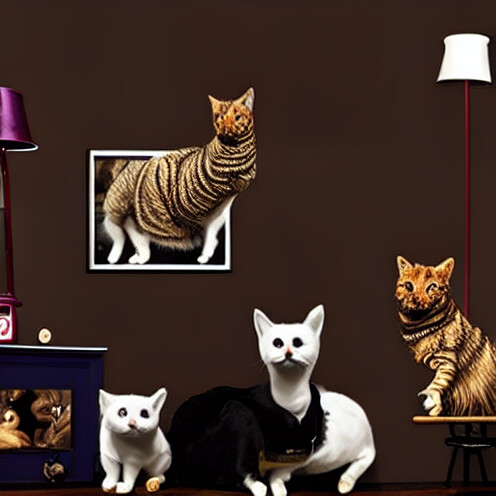}&
        \includegraphics[width=0.14\textwidth]{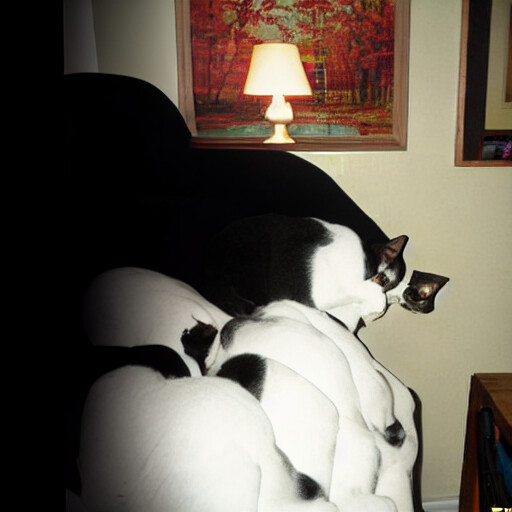}&
        \includegraphics[width=0.14\textwidth]{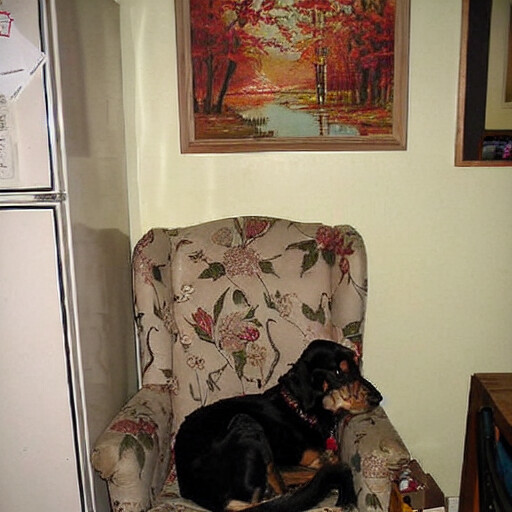}&
        \includegraphics[width=0.14\textwidth]{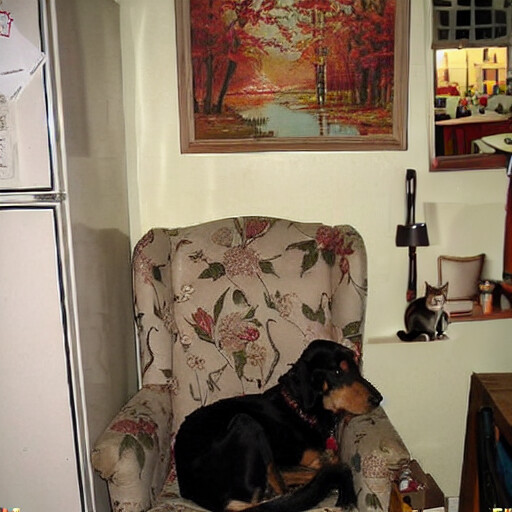}&
        \shortstack{
        \includegraphics[width=0.06\textwidth]{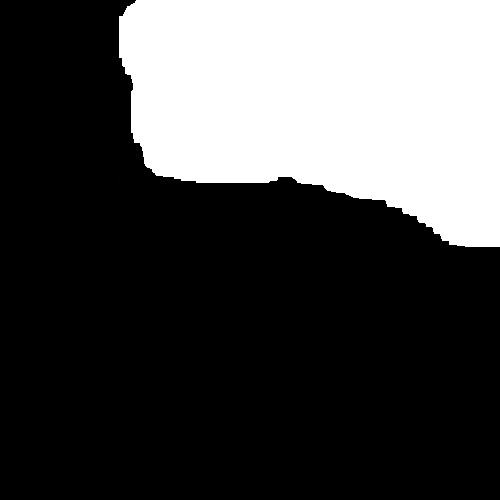}\quad\includegraphics[width=0.06\textwidth]{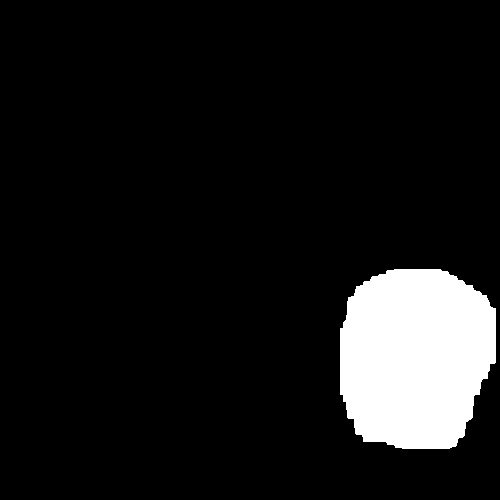}
        \quad\includegraphics[width=0.06\textwidth]{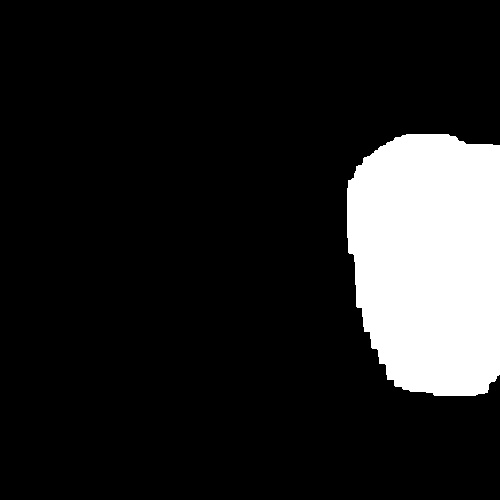}\\
        \texttt{[MASK]}~\texttt{[MASK]}~\texttt{[MASK]}}
        \\
        \multicolumn{7}{c}{(g) Edit instruction: \emph{``Add the dinning room in the background.}} \\
        \multicolumn{7}{c}{\emph{Also, add a cat beside the dog. Then, place a floor lamp next to the chair.''}} \\
        
        \includegraphics[width=0.14\textwidth]{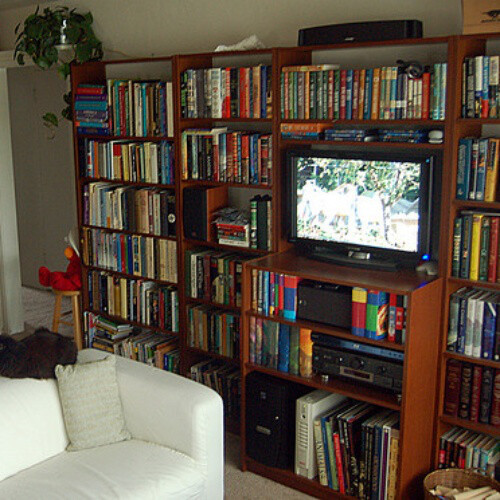}&
        \includegraphics[width=0.14\textwidth]{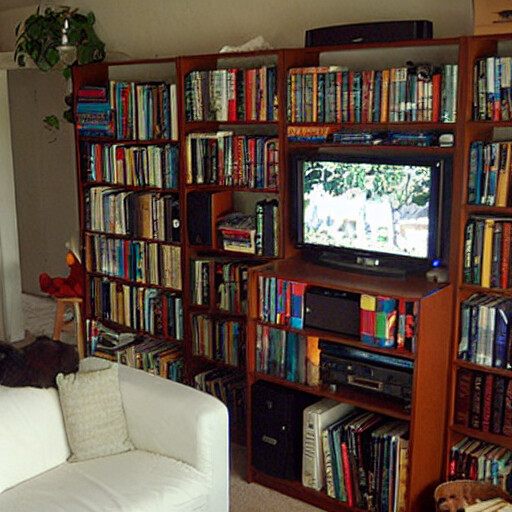}&
        \includegraphics[width=0.14\textwidth]{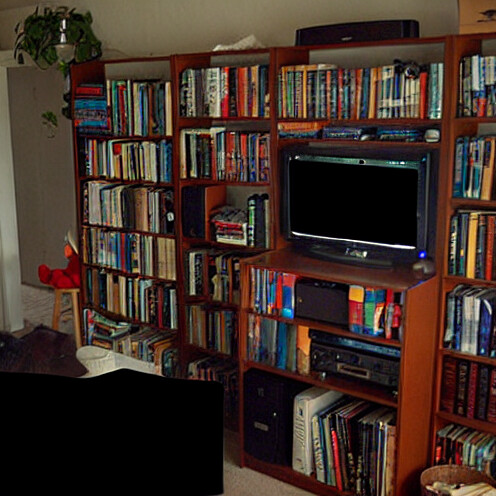}&
        \includegraphics[width=0.14\textwidth]{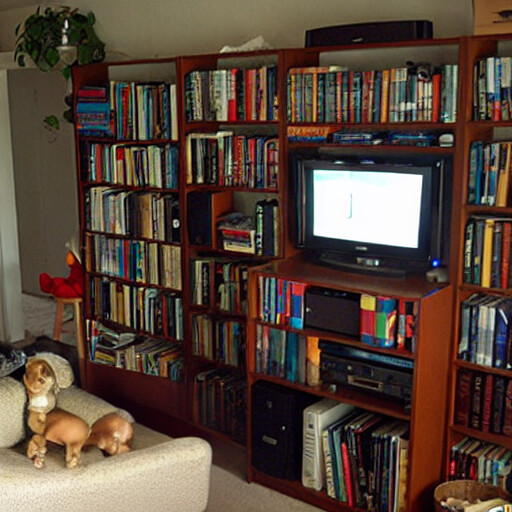}&
        \includegraphics[width=0.14\textwidth]{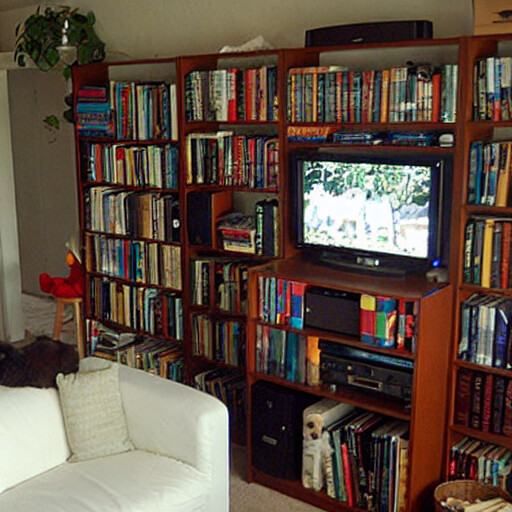}&
        \includegraphics[width=0.14\textwidth]{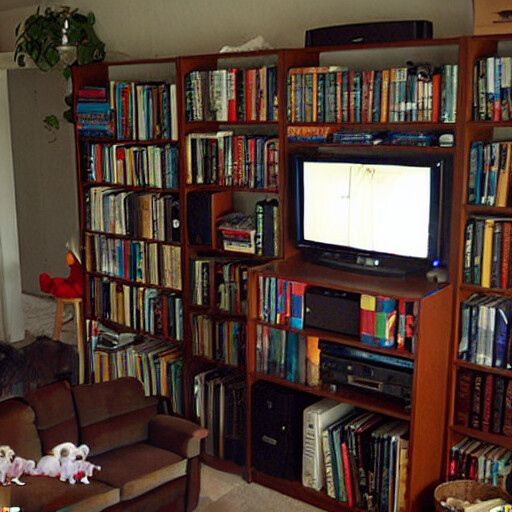}&
        \shortstack{
        \includegraphics[width=0.06\textwidth]{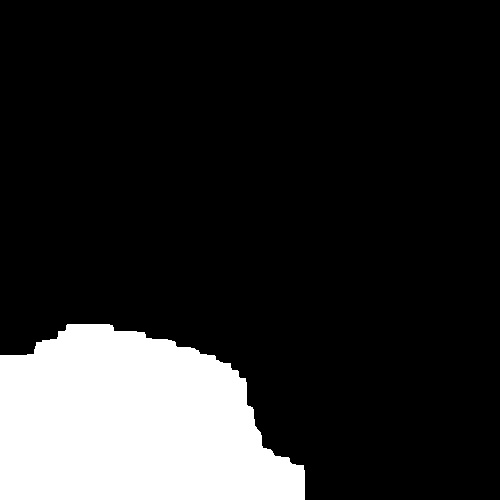}\quad\includegraphics[width=0.06\textwidth]{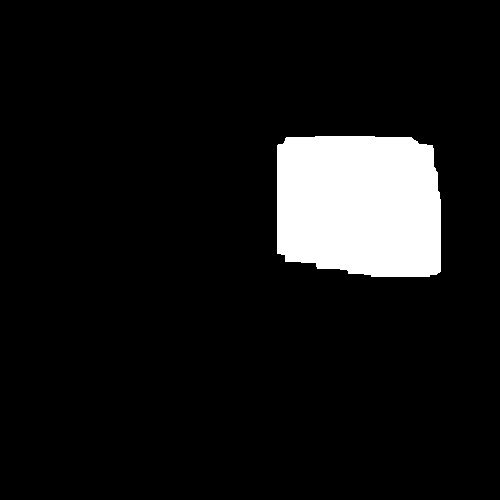}
        \quad\includegraphics[width=0.06\textwidth]{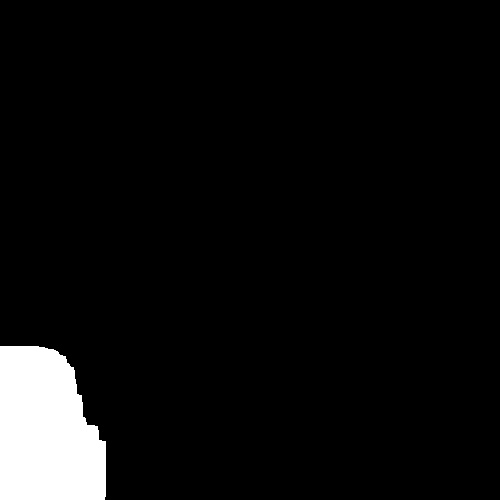}\\
        \texttt{[MASK]}~\texttt{[MASK]}~\texttt{[MASK]}}
        \\
        \multicolumn{7}{c}{(h) Edit instruction: \emph{``Change the white couch to a brown couch,}} \\
        \multicolumn{7}{c}{\emph{and let the TV have a blank screen. Also, add a puppy near the couch.''}} \\
        
    \end{tabular}}
    \vspace{-3mm}
    \caption{\textbf{Qualitative comparisons for multi-instruction task}}
    \label{fig:appendix_qualitative_multi}
    \vspace{-1.5em}
\end{figure*}

\begin{figure*}[t]
    \setlength\tabcolsep{2.5pt}
    \centering
    \resizebox{\linewidth}{!}{%
    \normalsize
    \begin{tabular}{ccccccc}
        Input Image & {\footnotesize \shortstack{IP2P\\\cite{brooks2023instructpix2pix}}} & {\footnotesize \shortstack{MGIE\\\cite{fu2024guiding}}} & {\footnotesize \shortstack{SmartEdit\\\cite{huang2023smartedit}}} & {\footnotesize \shortstack{FoI\\\cite{guo2024focus}}} & \multicolumn{2}{c}{\name~\textbf{(Ours)}} \\
             
        \includegraphics[width=0.14\textwidth]{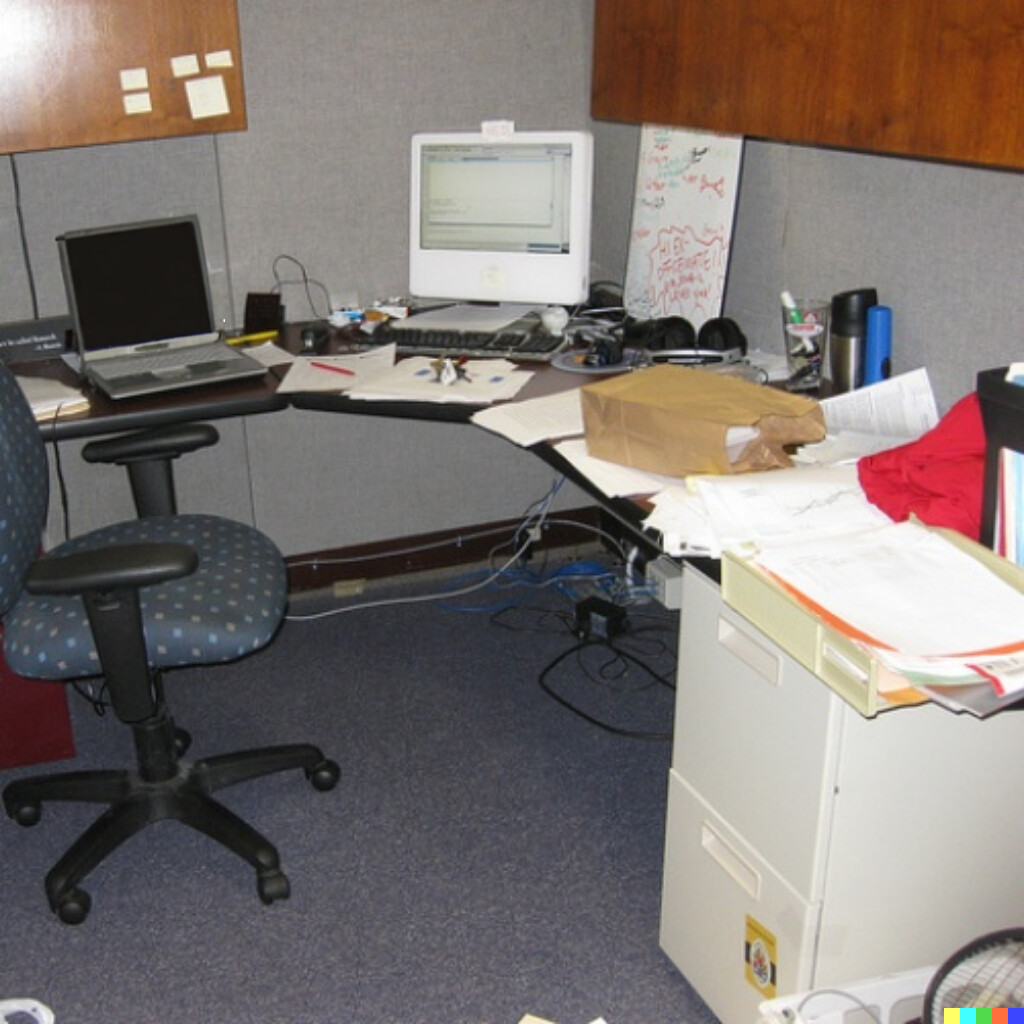}&
        \includegraphics[width=0.14\textwidth]{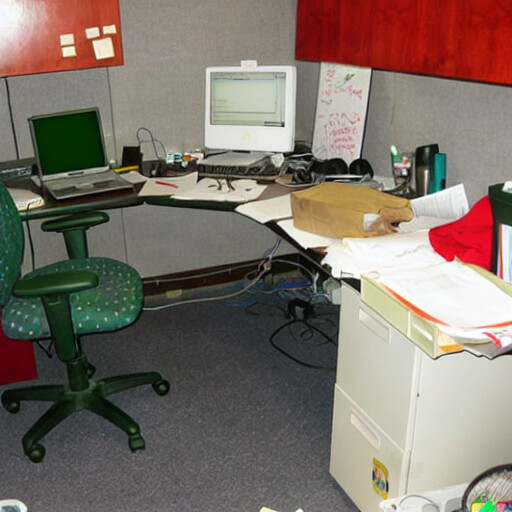}&
        \includegraphics[width=0.14\textwidth]{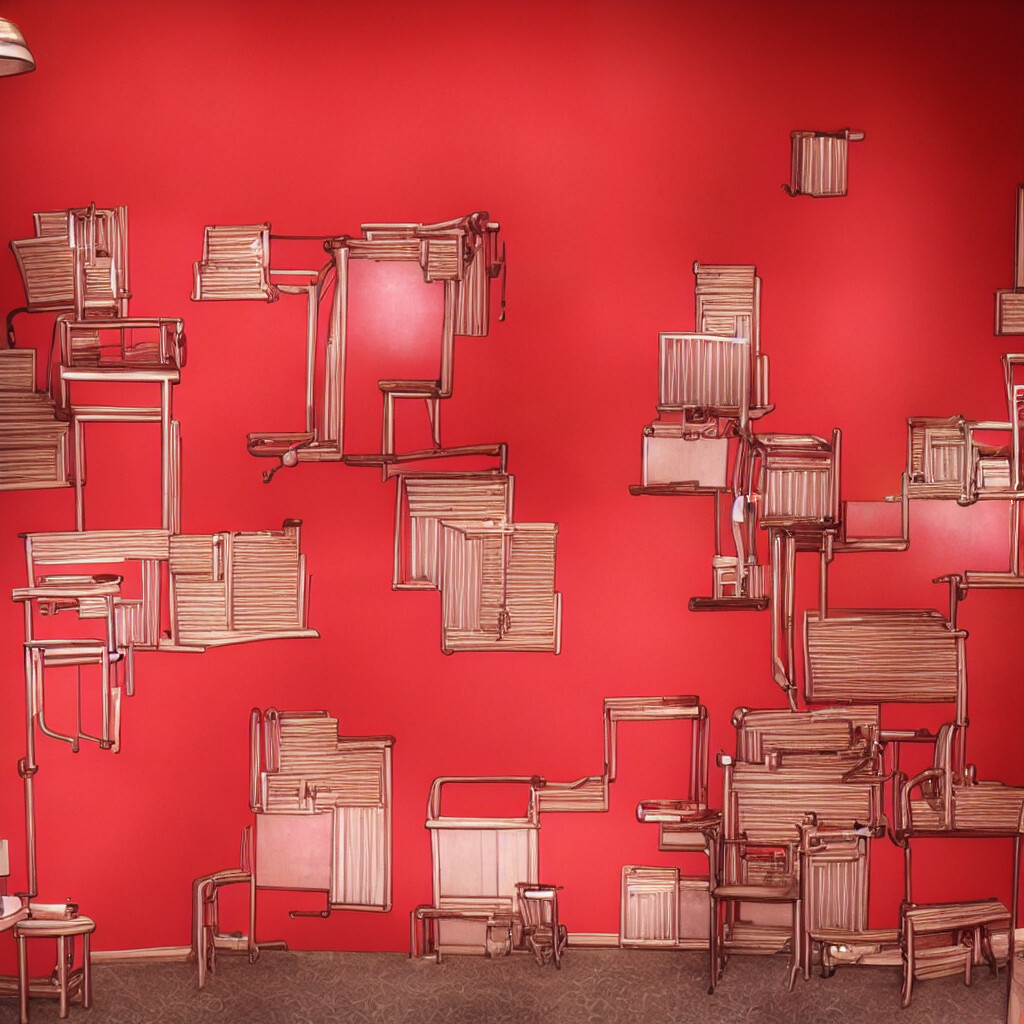}&
        \includegraphics[width=0.14\textwidth]{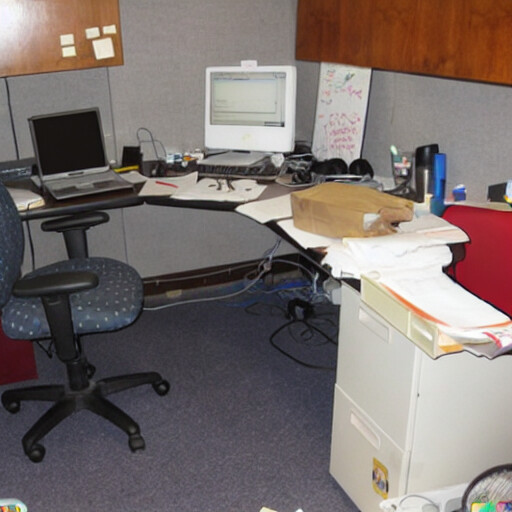}&
        \includegraphics[width=0.14\textwidth]{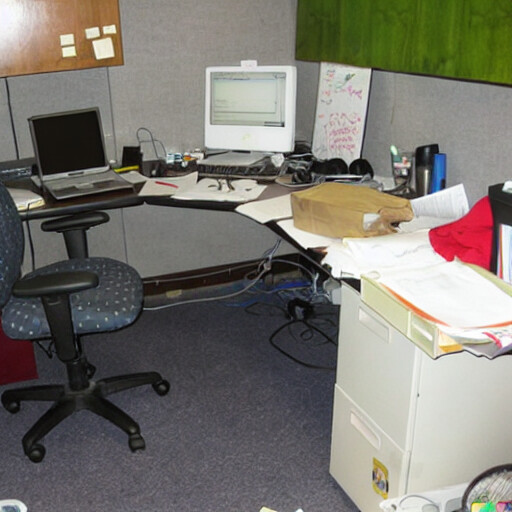}&
        \includegraphics[width=0.14\textwidth]{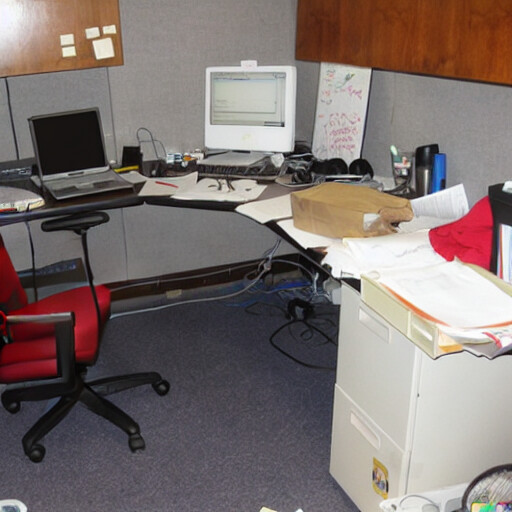}&
        \shortstack{
        \includegraphics[width=0.06\textwidth]{figure/negmask.jpg}\quad\includegraphics[width=0.06\textwidth]{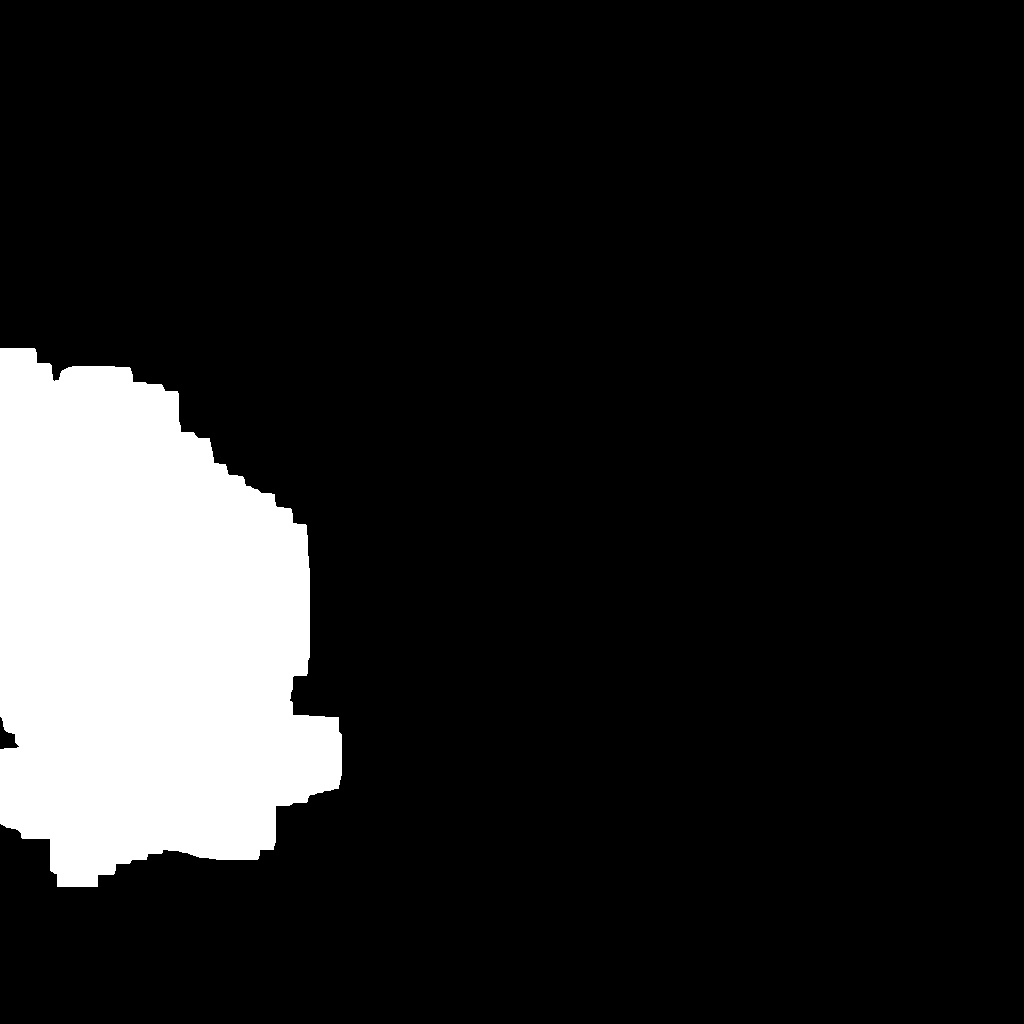}\\
        \texttt{[NEG]}~\texttt{[MASK]}}
        \\
        \multicolumn{7}{c}{(a) Edit instruction: \emph{``Change the color of the curtains to green and let the chair be red.''}} \\
        
        \includegraphics[width=0.14\textwidth]{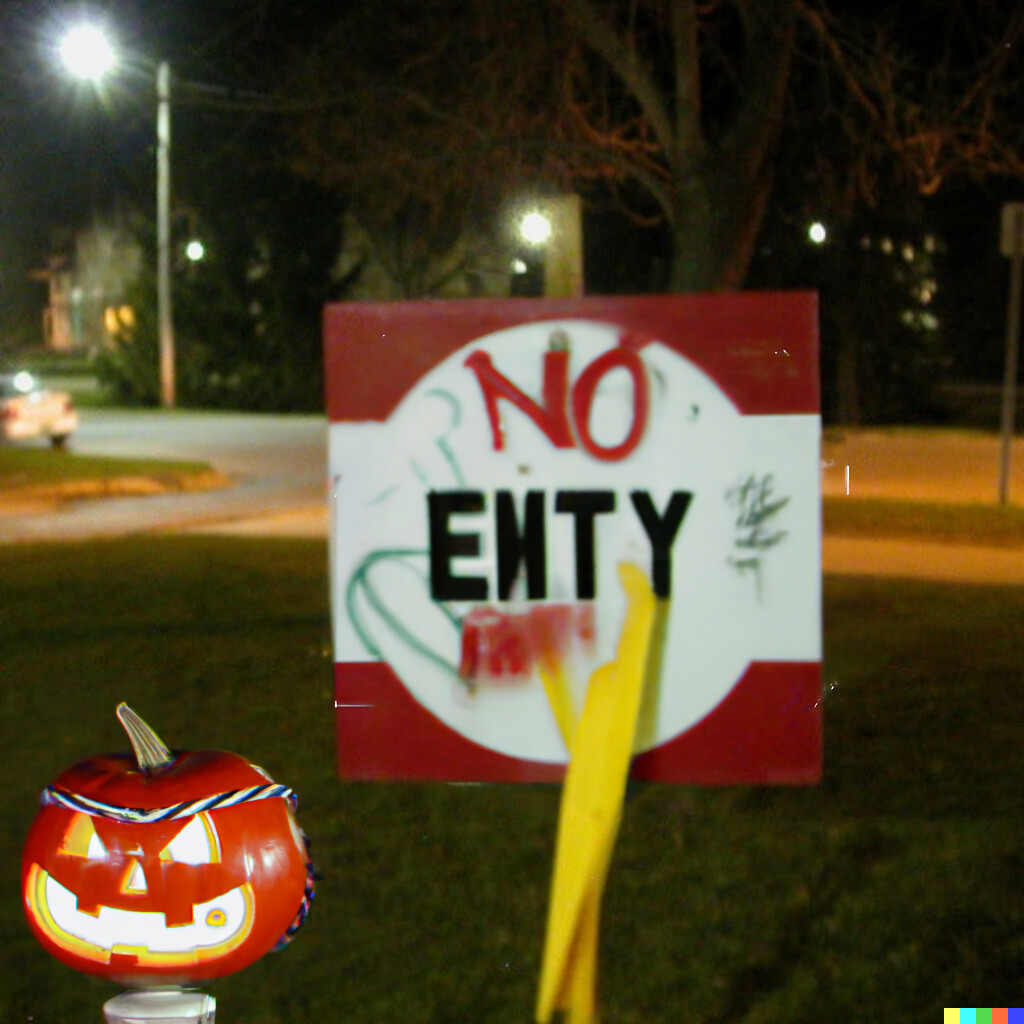}&
        \includegraphics[width=0.14\textwidth]{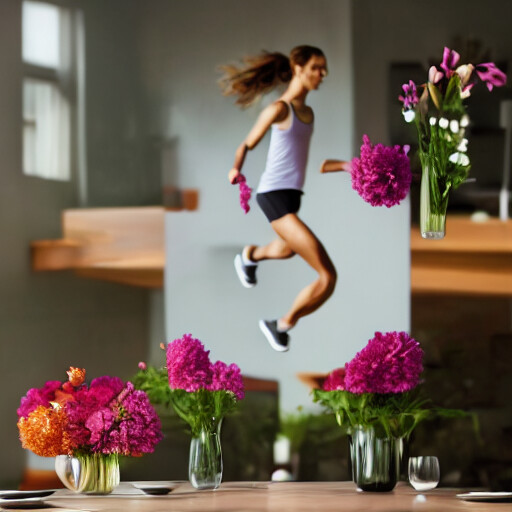}&
        \includegraphics[width=0.14\textwidth]{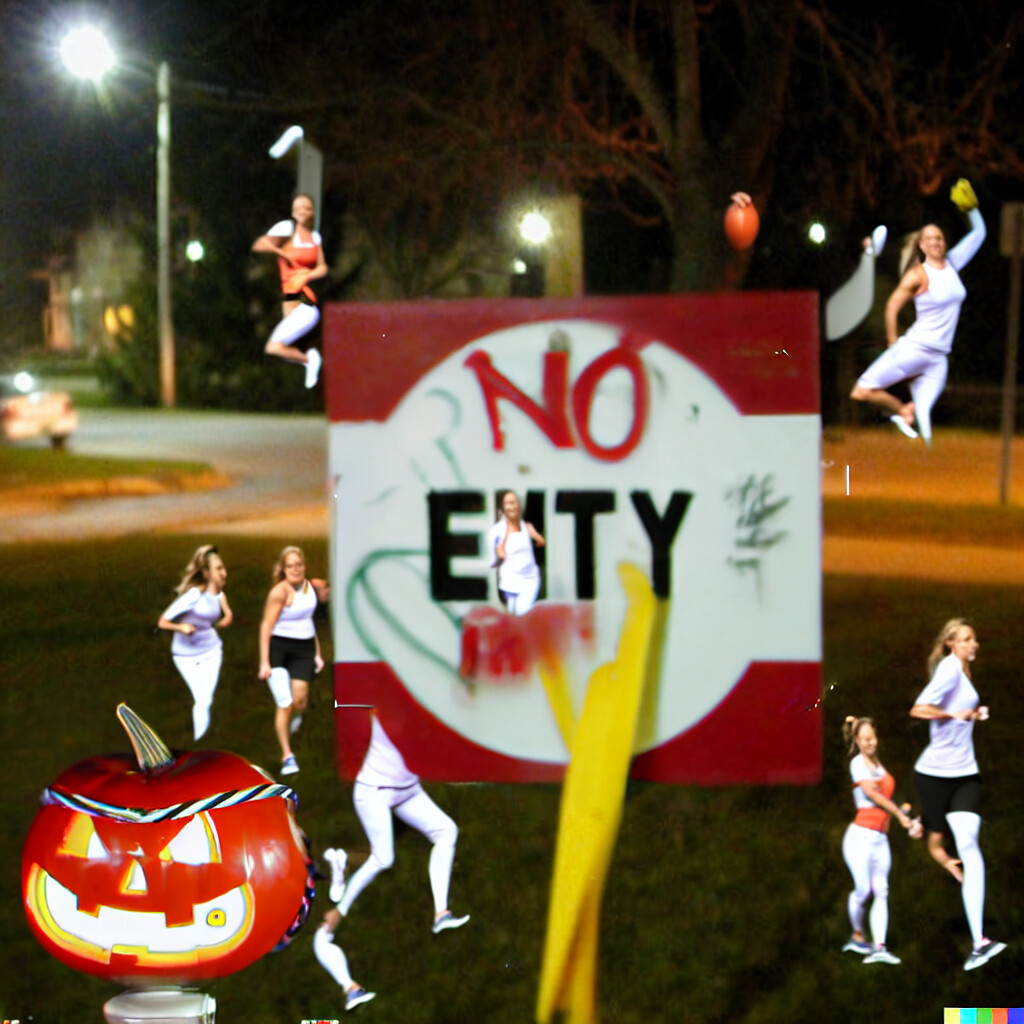}&
        \includegraphics[width=0.14\textwidth]{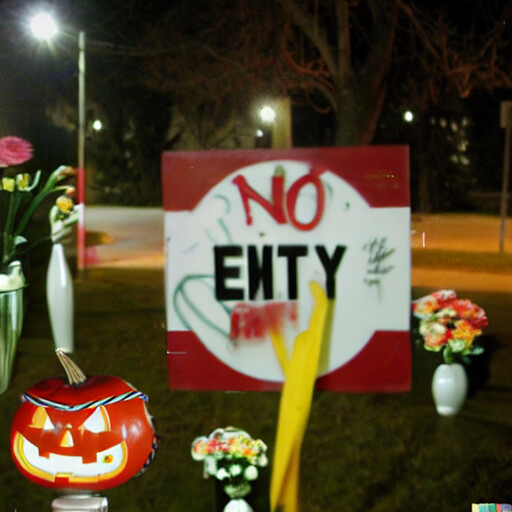}&
        \includegraphics[width=0.14\textwidth]{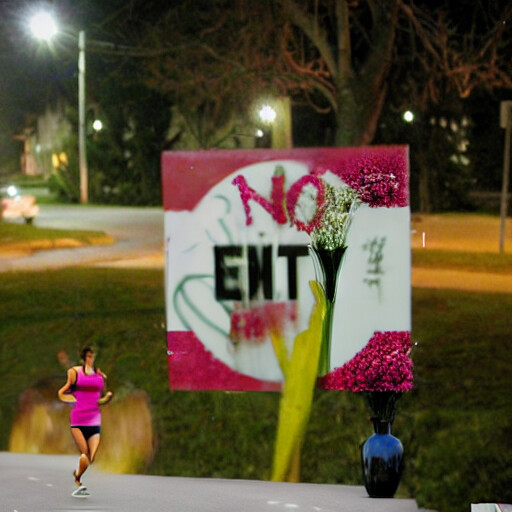}&
        \includegraphics[width=0.14\textwidth]{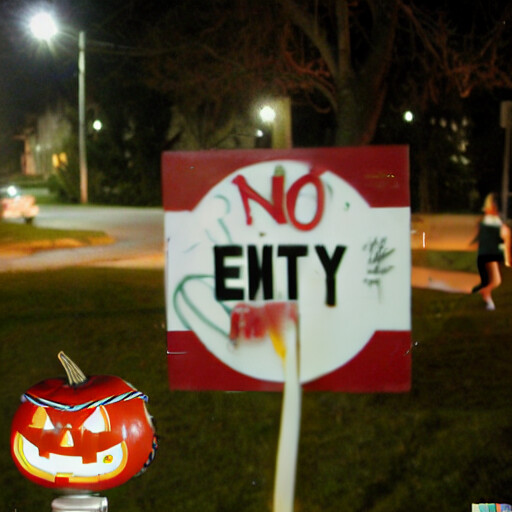}&
        \shortstack{
        \includegraphics[width=0.06\textwidth]{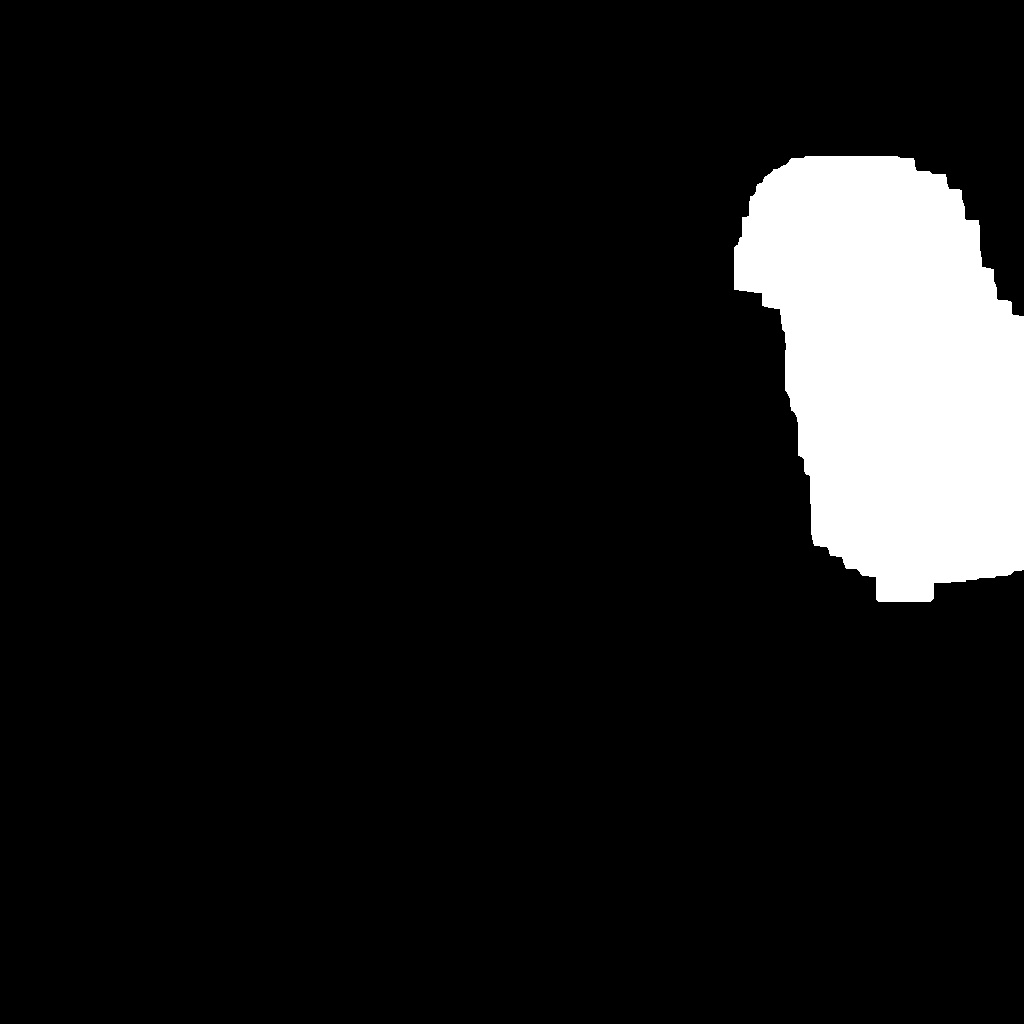}\quad\includegraphics[width=0.06\textwidth]{figure/negmask.jpg}\\
        \texttt{[MASK]~}~\texttt{[NEG]}}
        \\
        \multicolumn{7}{c}{(b) Edit instruction: \emph{``Add a woman jogging in the back.}} \\
        \multicolumn{7}{c}{\emph{Furthermore, place a flower vase on the dining table.''}} \\
        
        \includegraphics[width=0.14\textwidth]{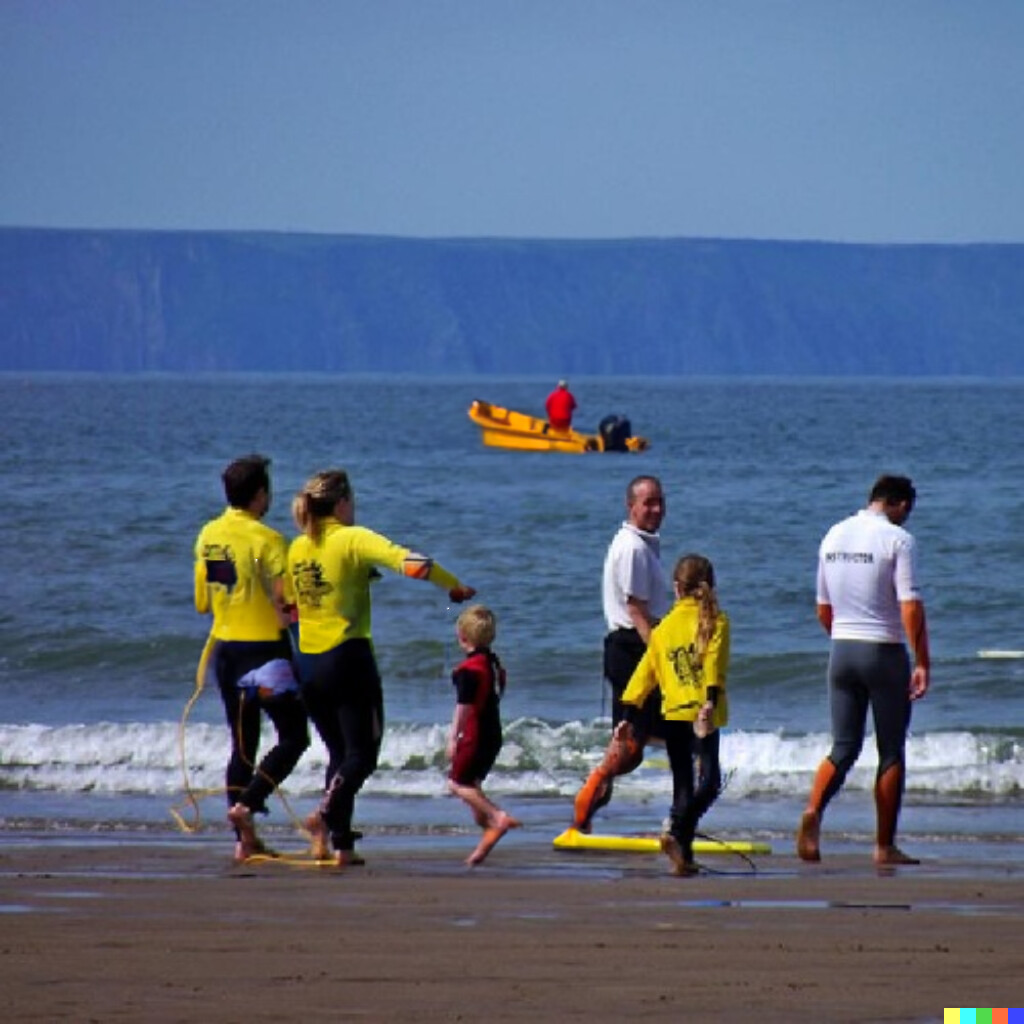}&
        \includegraphics[width=0.14\textwidth]{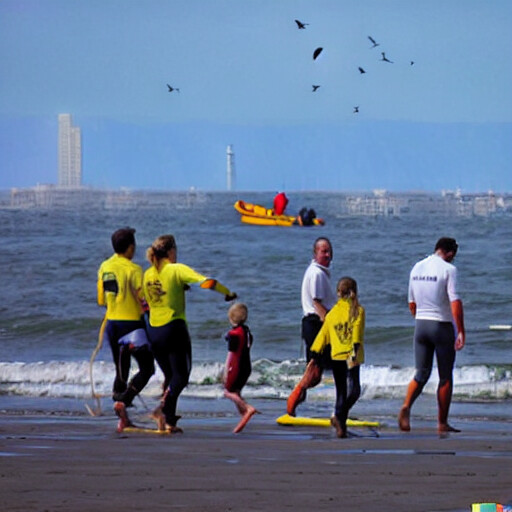}&
        \includegraphics[width=0.14\textwidth]{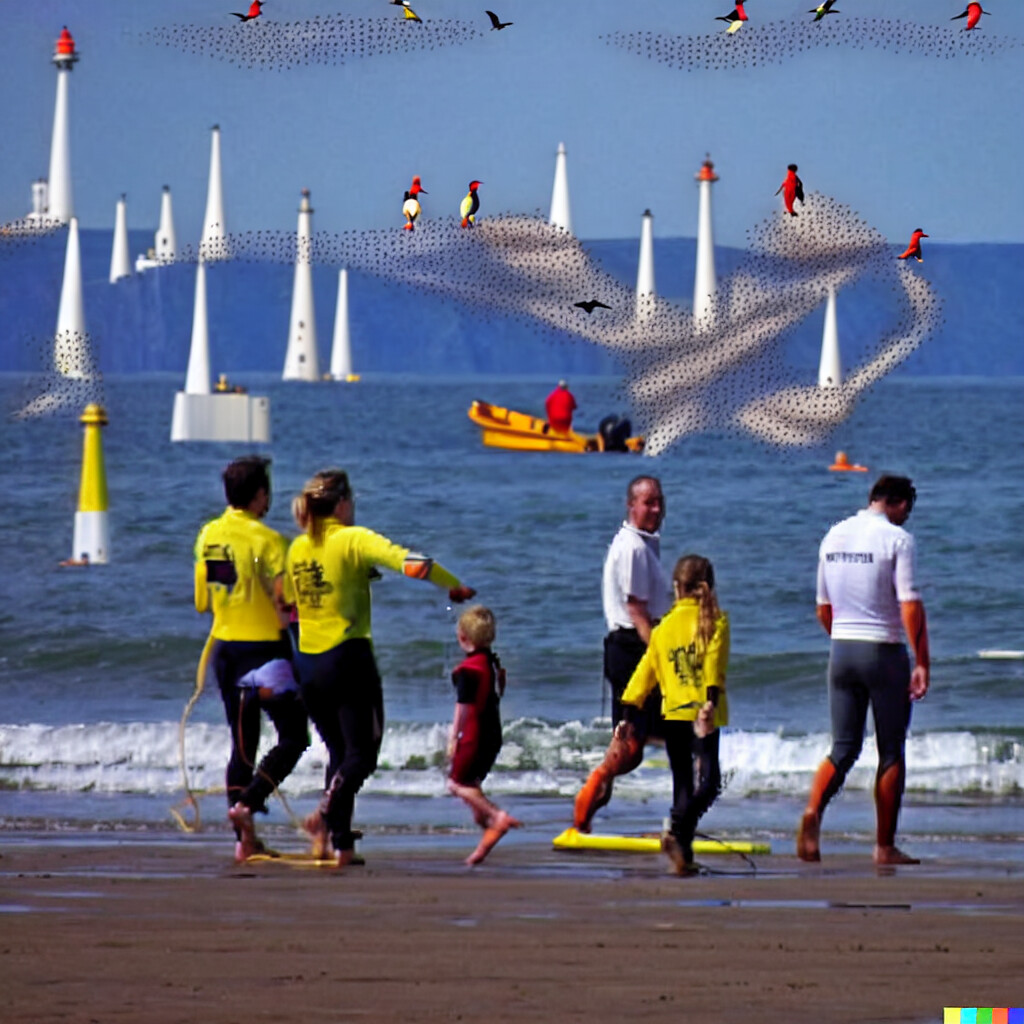}&
        \includegraphics[width=0.14\textwidth]{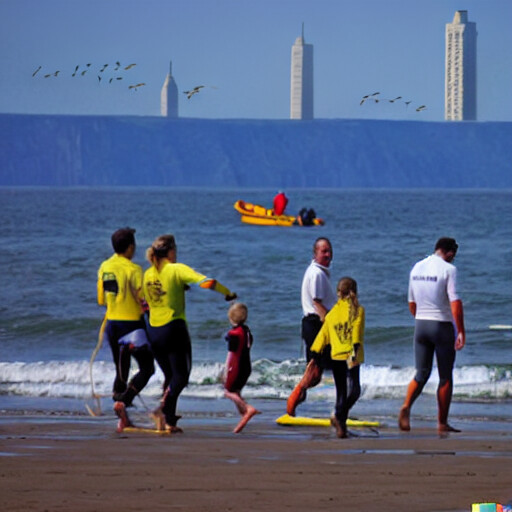}&
        \includegraphics[width=0.14\textwidth]{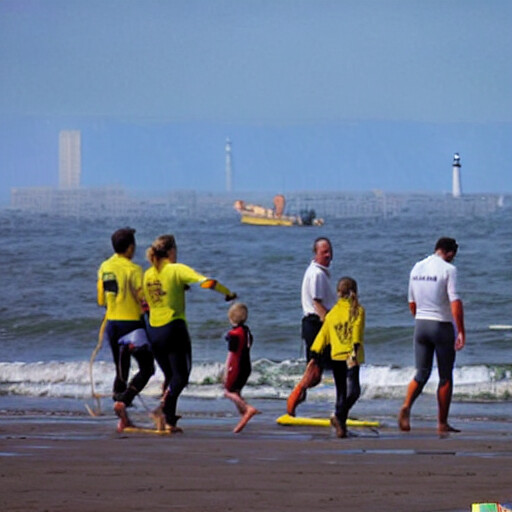}&
        \includegraphics[width=0.14\textwidth]{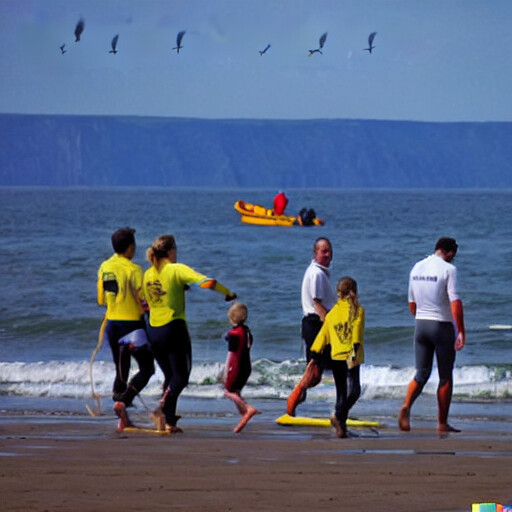}&
        \shortstack{
        \includegraphics[width=0.06\textwidth]{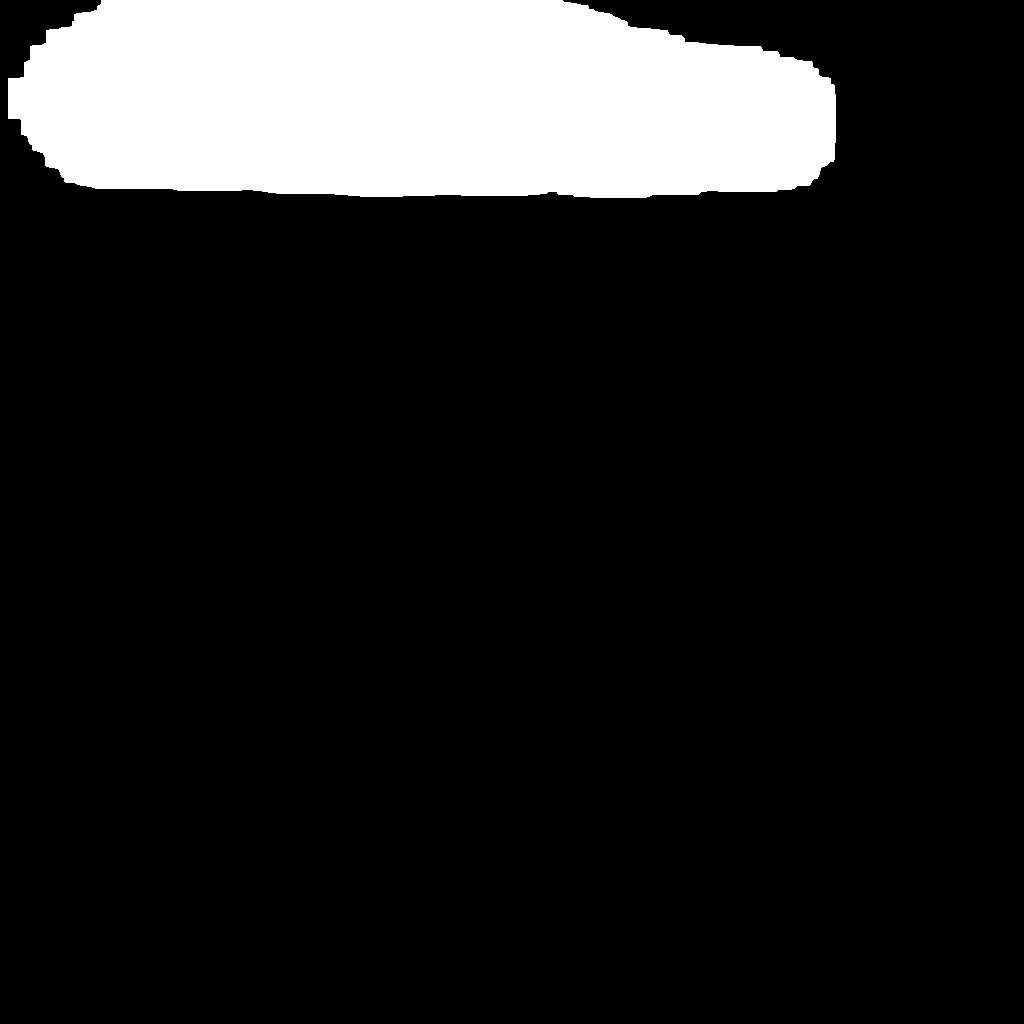}\quad\includegraphics[width=0.06\textwidth]{figure/negmask.jpg}\\
        \texttt{[MASK]~}~\texttt{[NEG]}}
        \\
        \multicolumn{7}{c}{(c) Edit instruction: \emph{``Add birds to the sky,}} \\
        \multicolumn{7}{c}{\emph{and then replace the lighthouse in the background with a tall building.''}} \\
        
        \includegraphics[width=0.14\textwidth]{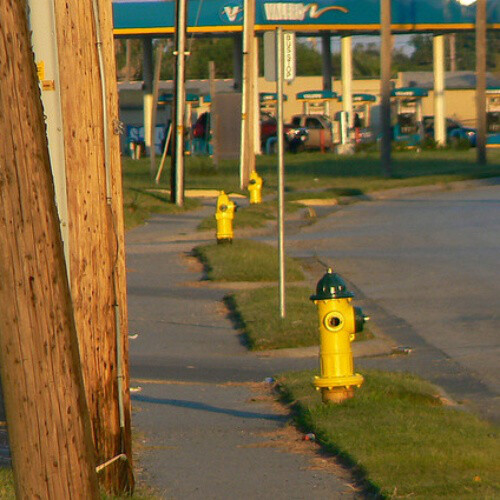}&
        \includegraphics[width=0.14\textwidth]{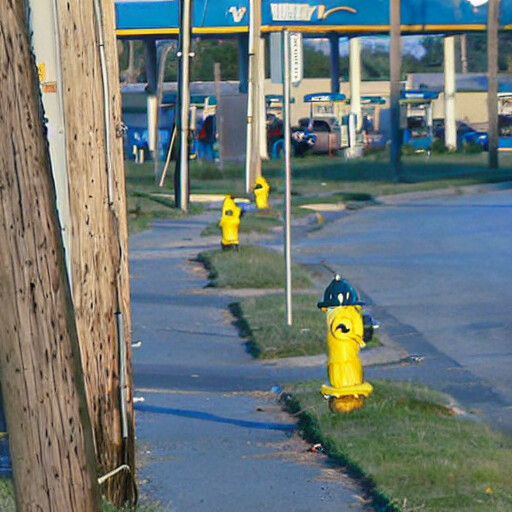}&
        \includegraphics[width=0.14\textwidth]{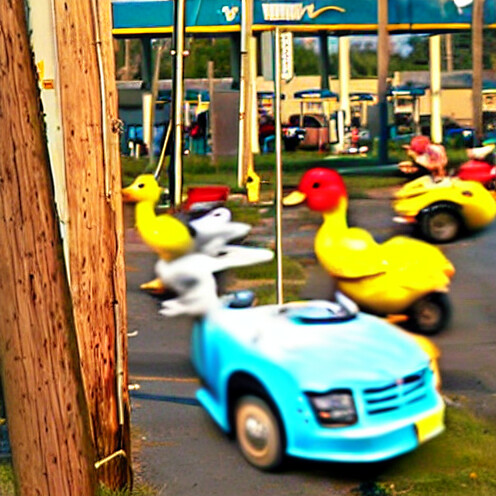}&
        \includegraphics[width=0.14\textwidth]{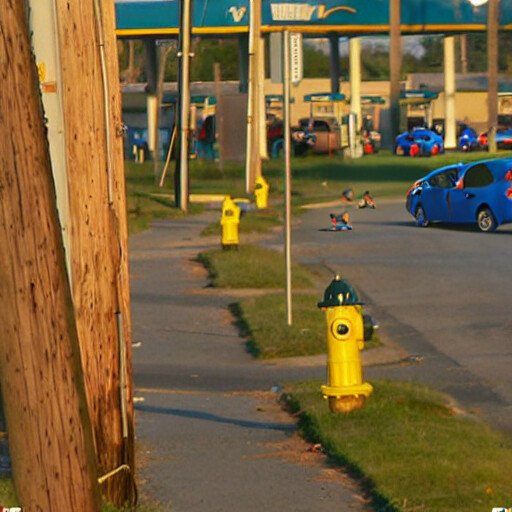}&
        \includegraphics[width=0.14\textwidth]{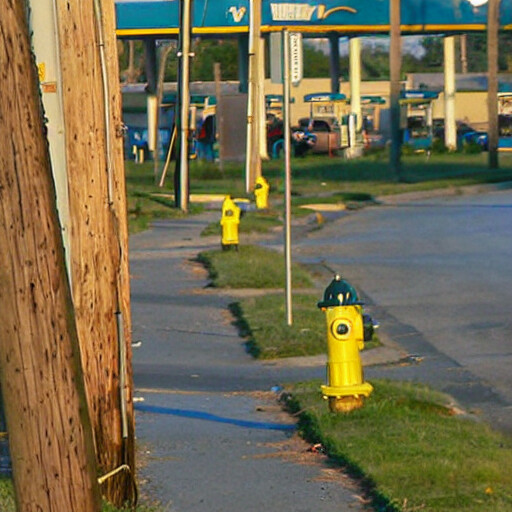}&
        \includegraphics[width=0.14\textwidth]{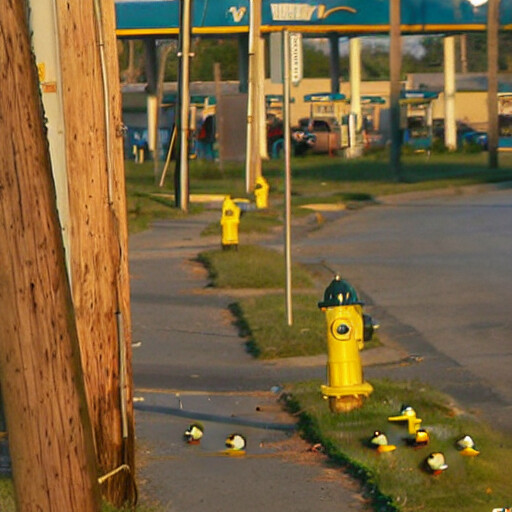}&
        \shortstack{
        \includegraphics[width=0.06\textwidth]{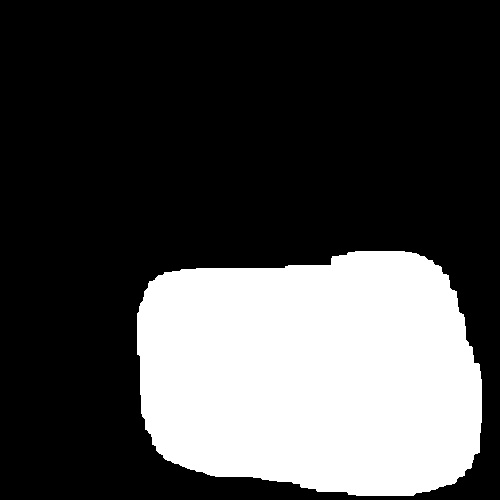}\quad\includegraphics[width=0.06\textwidth]{figure/negmask.jpg}\\
        \texttt{[MASK]~}~\texttt{[NEG]}}
        \\
        \multicolumn{7}{c}{(d) Edit instruction: \emph{``There should be lots of baby ducklings in the street.}} \\
        \multicolumn{7}{c}{\emph{Next, change the color of the car parked by the tree to blue.''}} \\
        
        \includegraphics[width=0.14\textwidth]{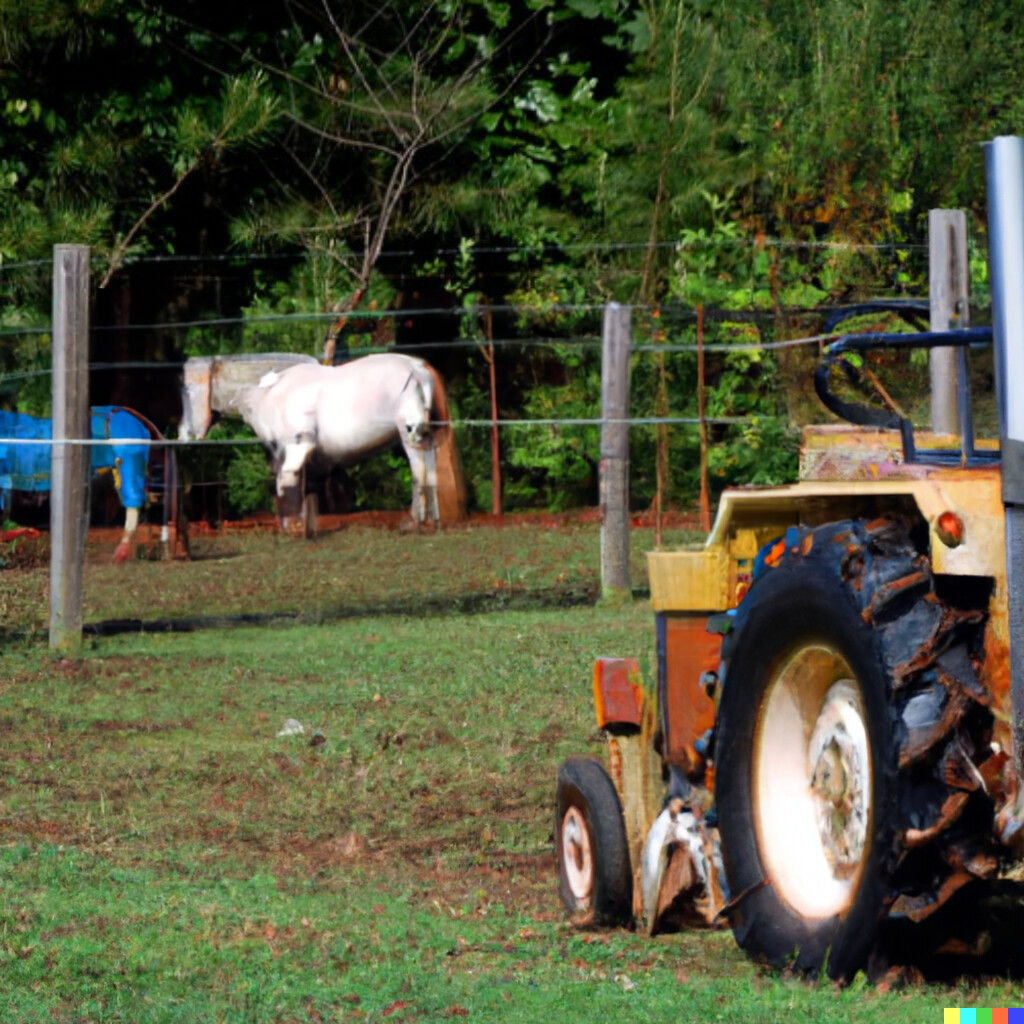}&
        \includegraphics[width=0.14\textwidth]{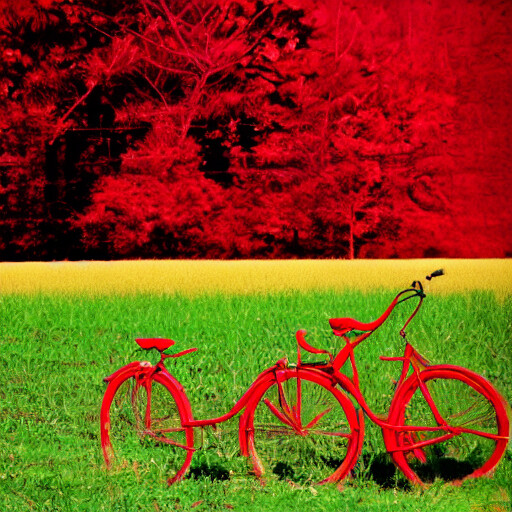}&
        \includegraphics[width=0.14\textwidth]{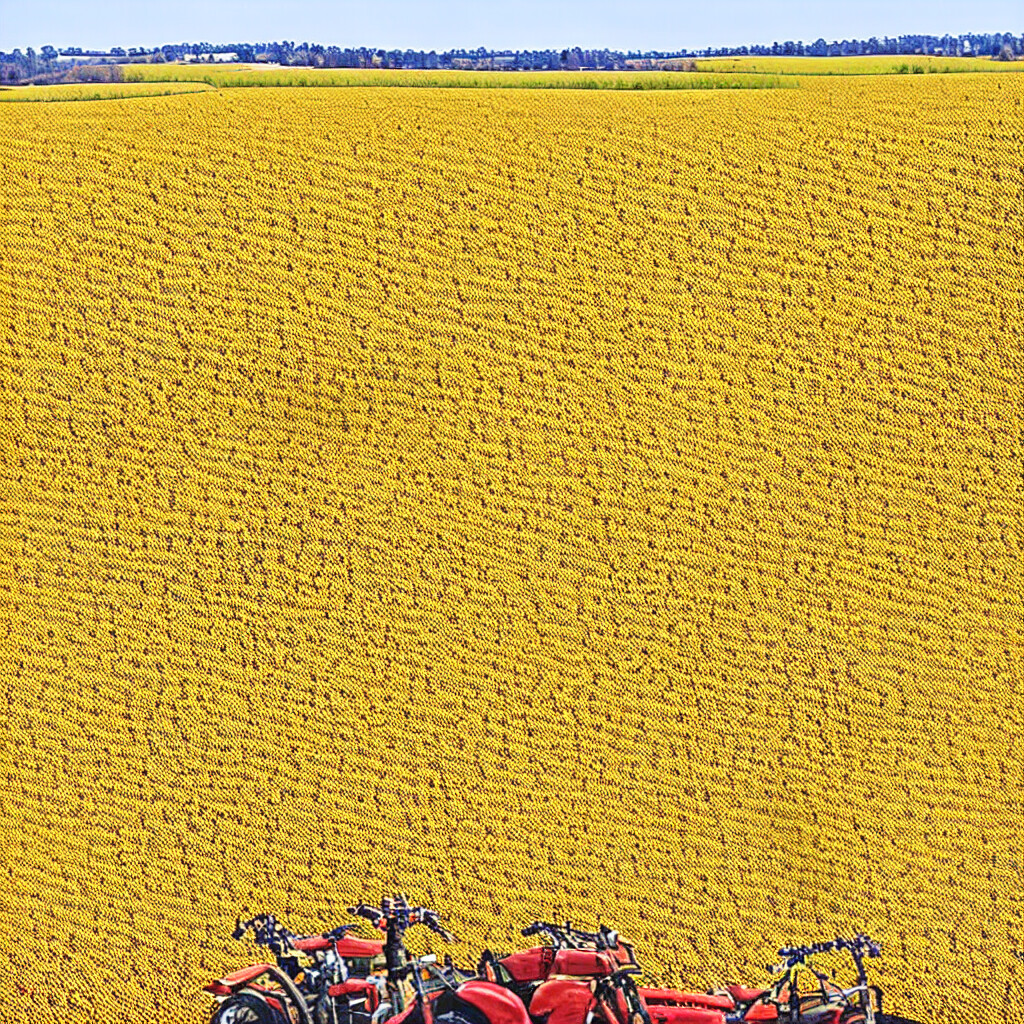}&
        \includegraphics[width=0.14\textwidth]{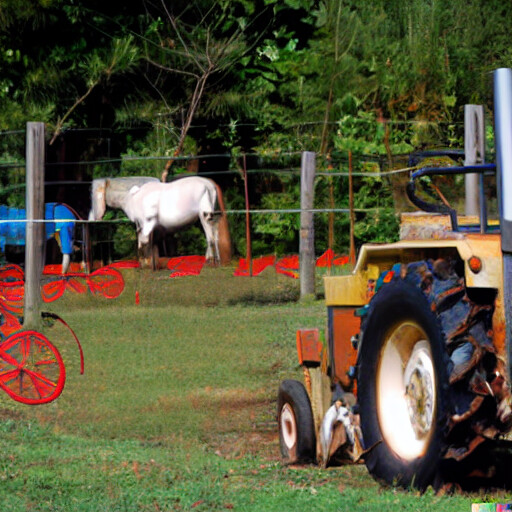}&
        \includegraphics[width=0.14\textwidth]{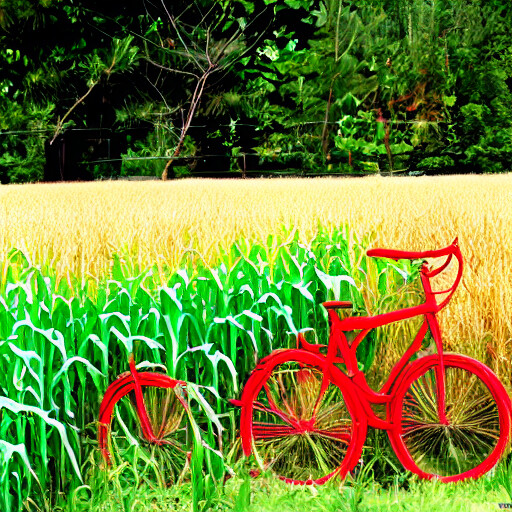}&
        \includegraphics[width=0.14\textwidth]{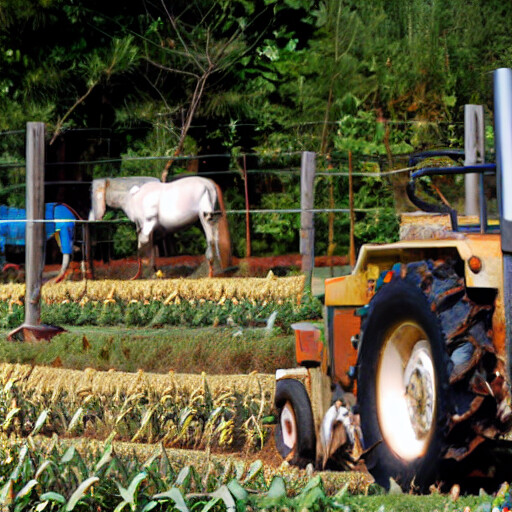}&
        \shortstack{
        \includegraphics[width=0.06\textwidth]{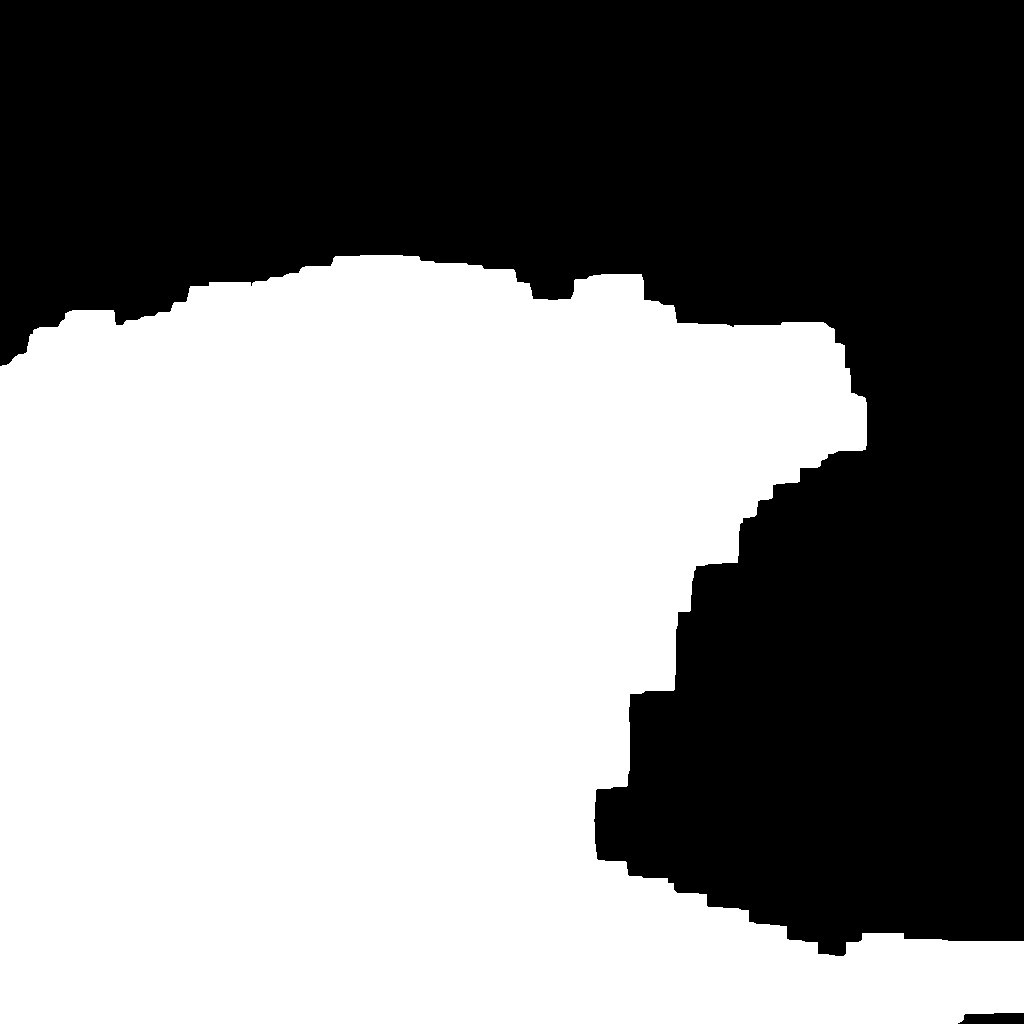}\quad\includegraphics[width=0.06\textwidth]{figure/negmask.jpg}\\
        \texttt{[MASK]~}~\texttt{[NEG]}}
        \\
        \multicolumn{7}{c}{(e) Edit instruction: \emph{``Make the field a cornfield.}} \\
        \multicolumn{7}{c}{\emph{Besides, change the color of the bicycle leaning against the tree to red.''}} \\
        
        \includegraphics[width=0.14\textwidth]{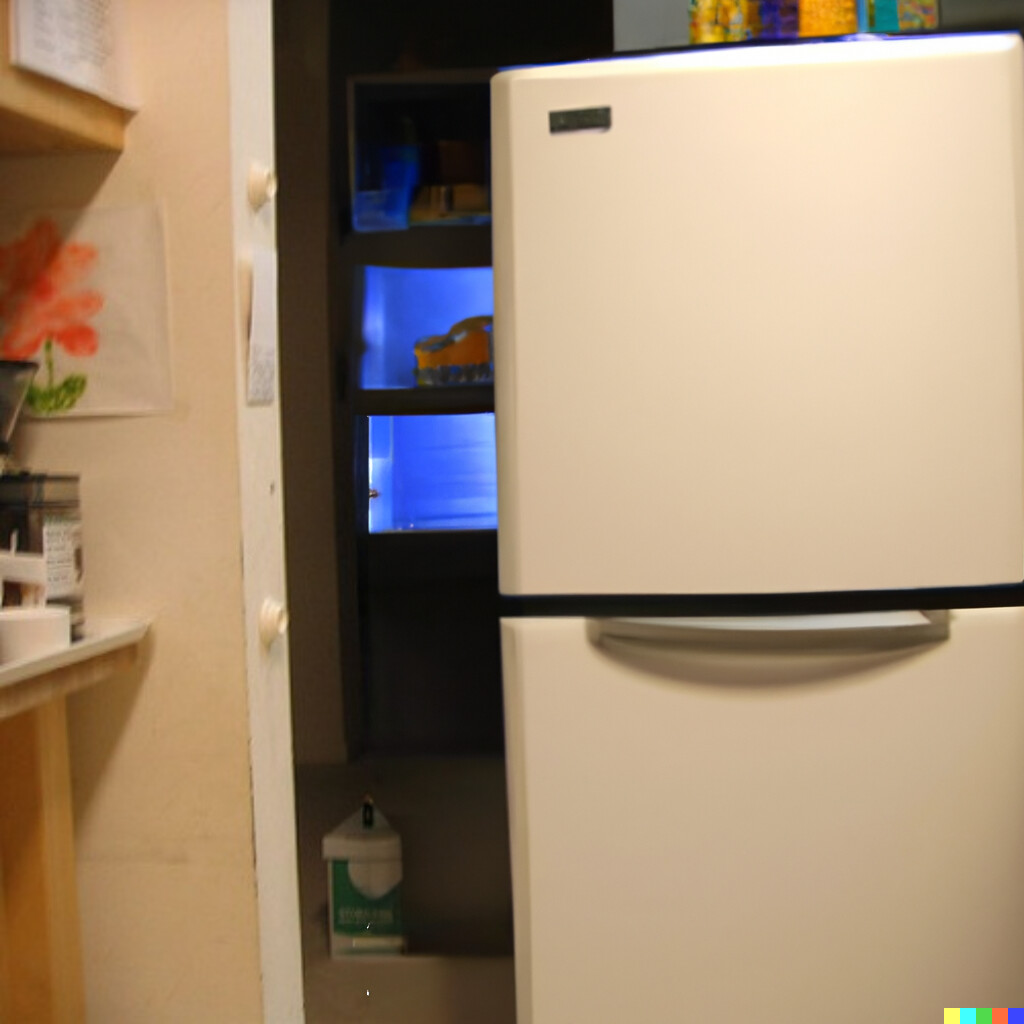}&
        \includegraphics[width=0.14\textwidth]{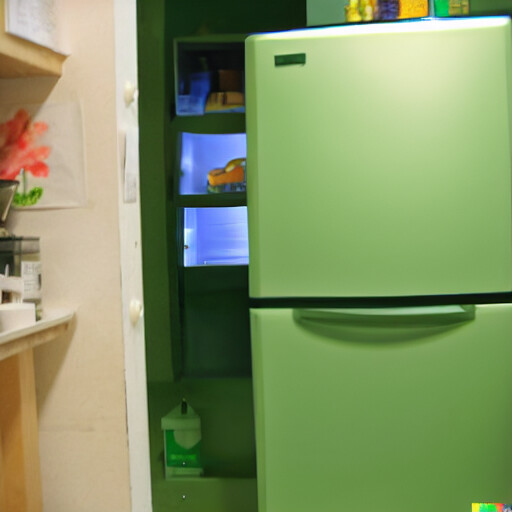}&
        \includegraphics[width=0.14\textwidth]{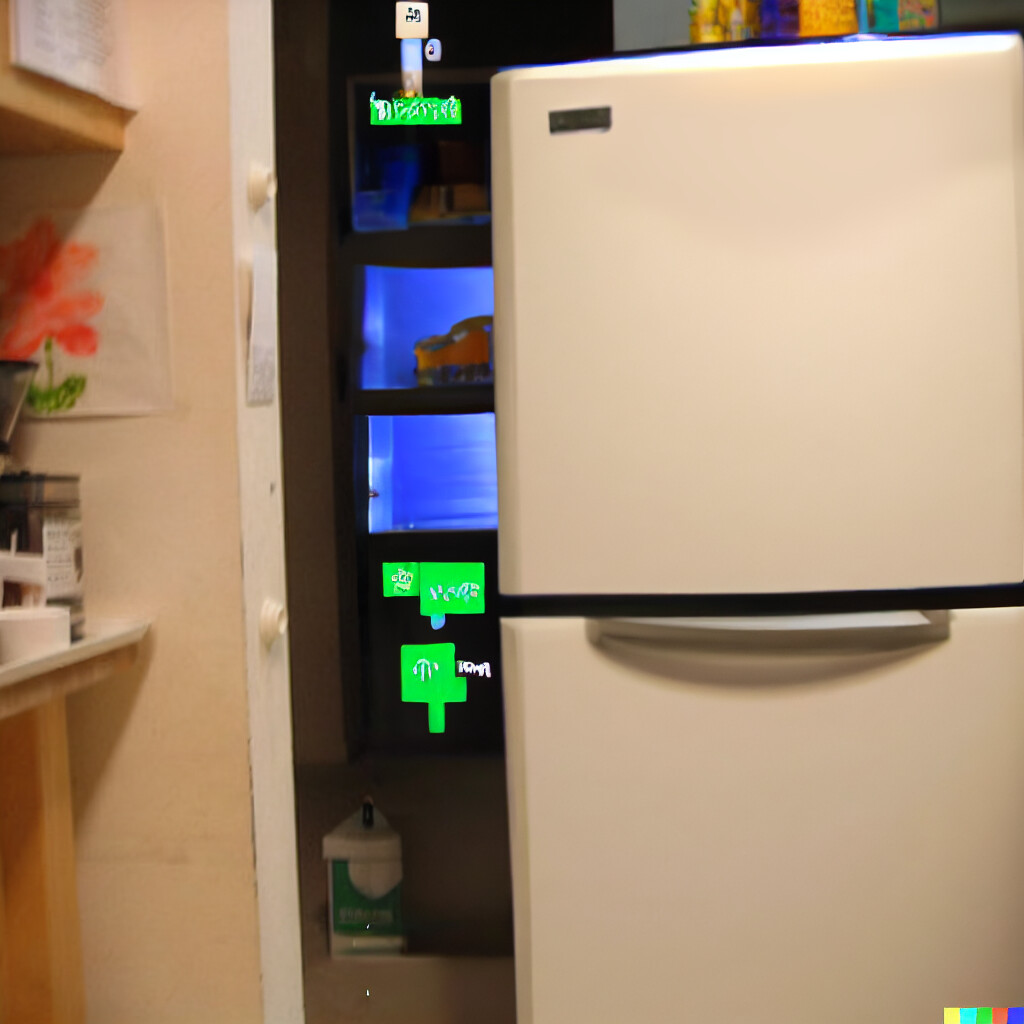}&
        \includegraphics[width=0.14\textwidth]{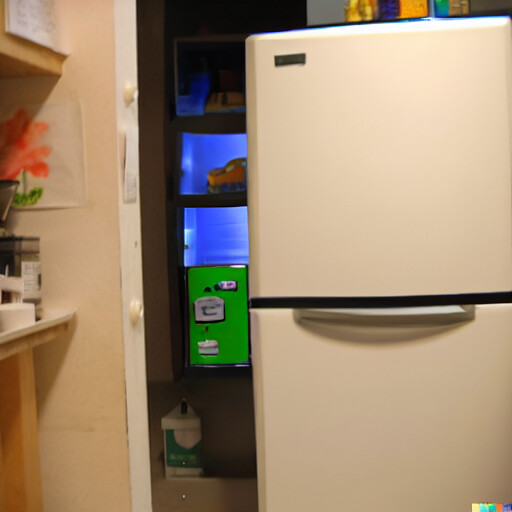}&
        \includegraphics[width=0.14\textwidth]{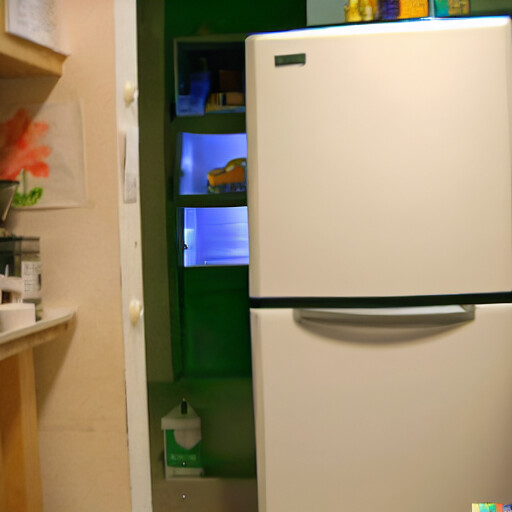}&
        \includegraphics[width=0.14\textwidth]{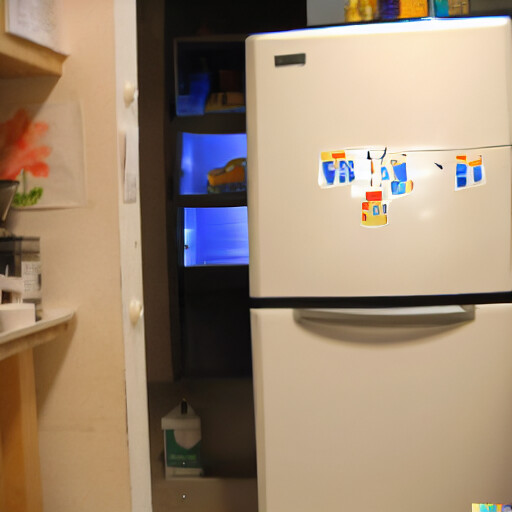}&
        \shortstack{
        \includegraphics[width=0.06\textwidth]{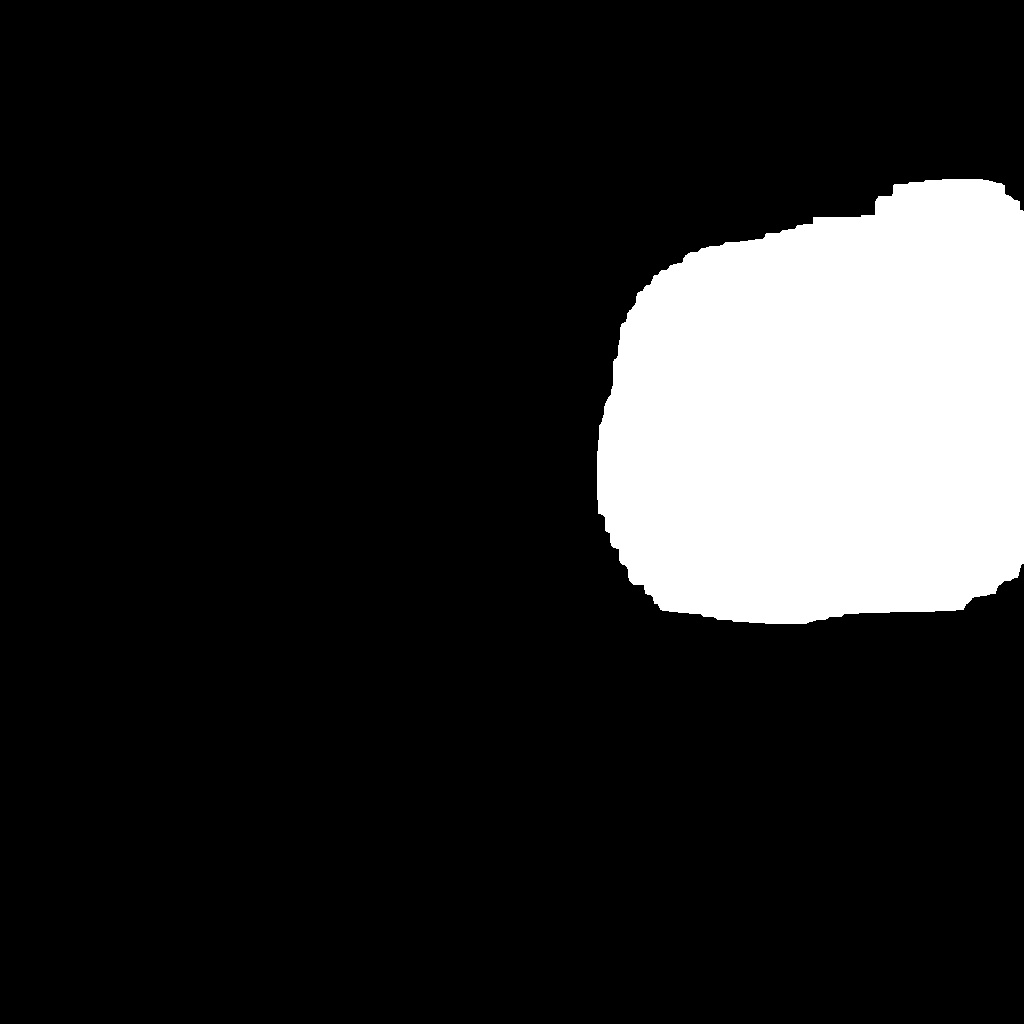}\quad\includegraphics[width=0.06\textwidth]{figure/negmask.jpg}\\
        \texttt{[MASK]~}~\texttt{[NEG]}}
        \\
        \multicolumn{7}{c}{(f) Edit instruction: \emph{``We could add some stickers to the refrigerator.}} \\
        \multicolumn{7}{c}{\emph{After that, change the color of the pot on the stove to green.''}} \\
        
        \includegraphics[width=0.14\textwidth]{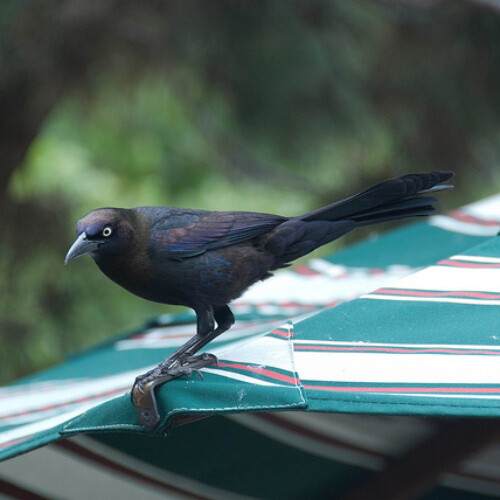}&
        \includegraphics[width=0.14\textwidth]{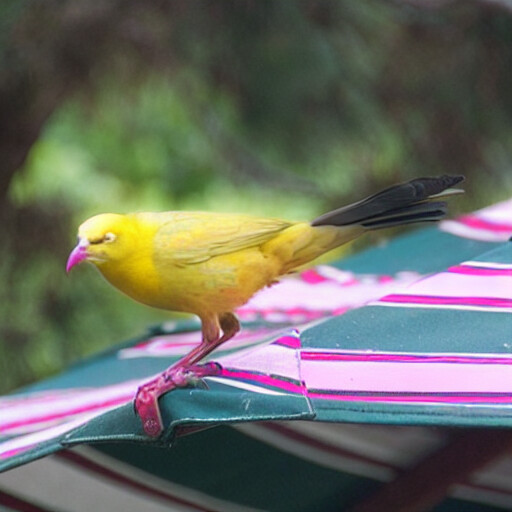}&
        \includegraphics[width=0.14\textwidth]{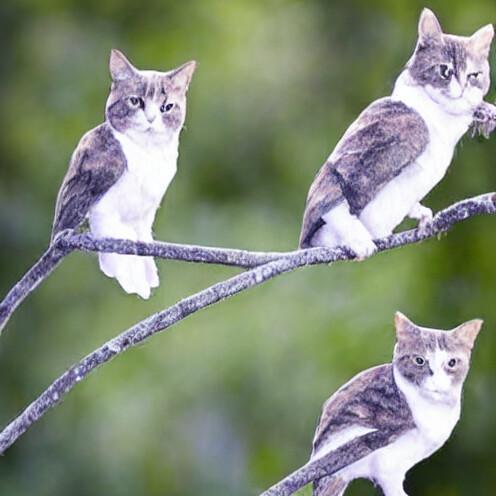}&
        \includegraphics[width=0.14\textwidth]{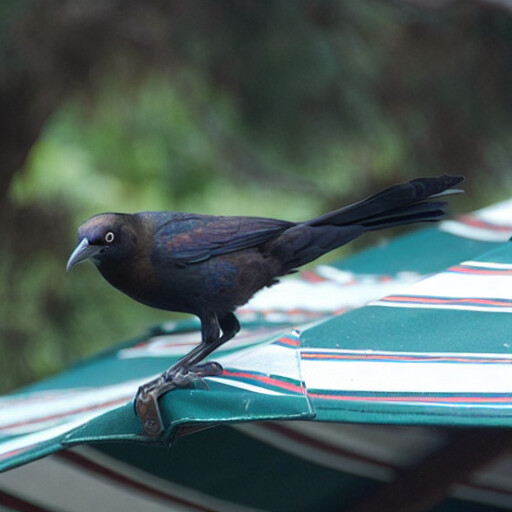}&
        \includegraphics[width=0.14\textwidth]{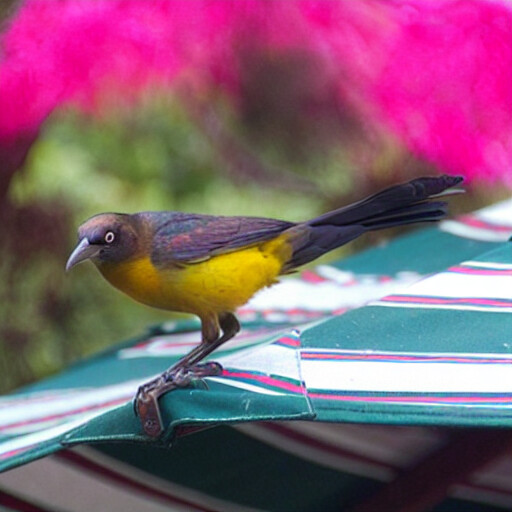}&
        \includegraphics[width=0.14\textwidth]{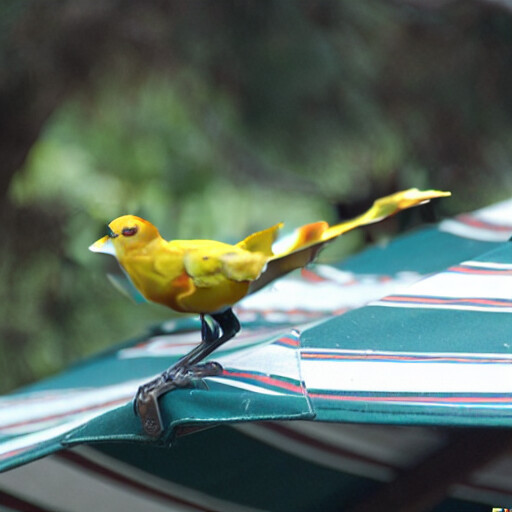}&
        \shortstack{
        \includegraphics[width=0.06\textwidth]{figure/negmask.jpg}\quad\includegraphics[width=0.06\textwidth]{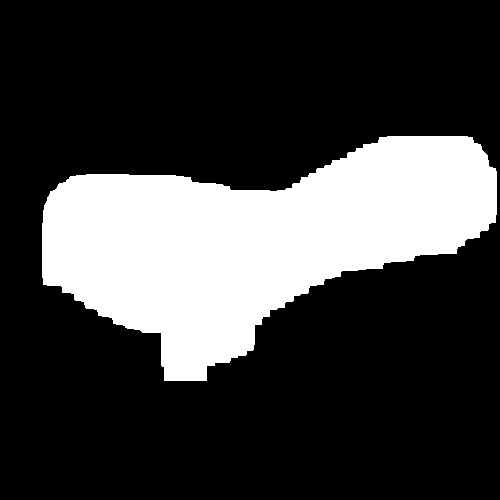}\quad\includegraphics[width=0.06\textwidth]{figure/negmask.jpg}\\
        \texttt{[NEG]~}\texttt{[MASK]}\texttt{[NEG]}}
        \\
        \multicolumn{7}{c}{(g) Edit instruction: \emph{``Remove the cat sitting on the fence, and let the bird turn yellow.}} \\
        \multicolumn{7}{c}{\emph{Then, change the white flowers on the tree to pink.''}}\\
        
        \includegraphics[width=0.14\textwidth]{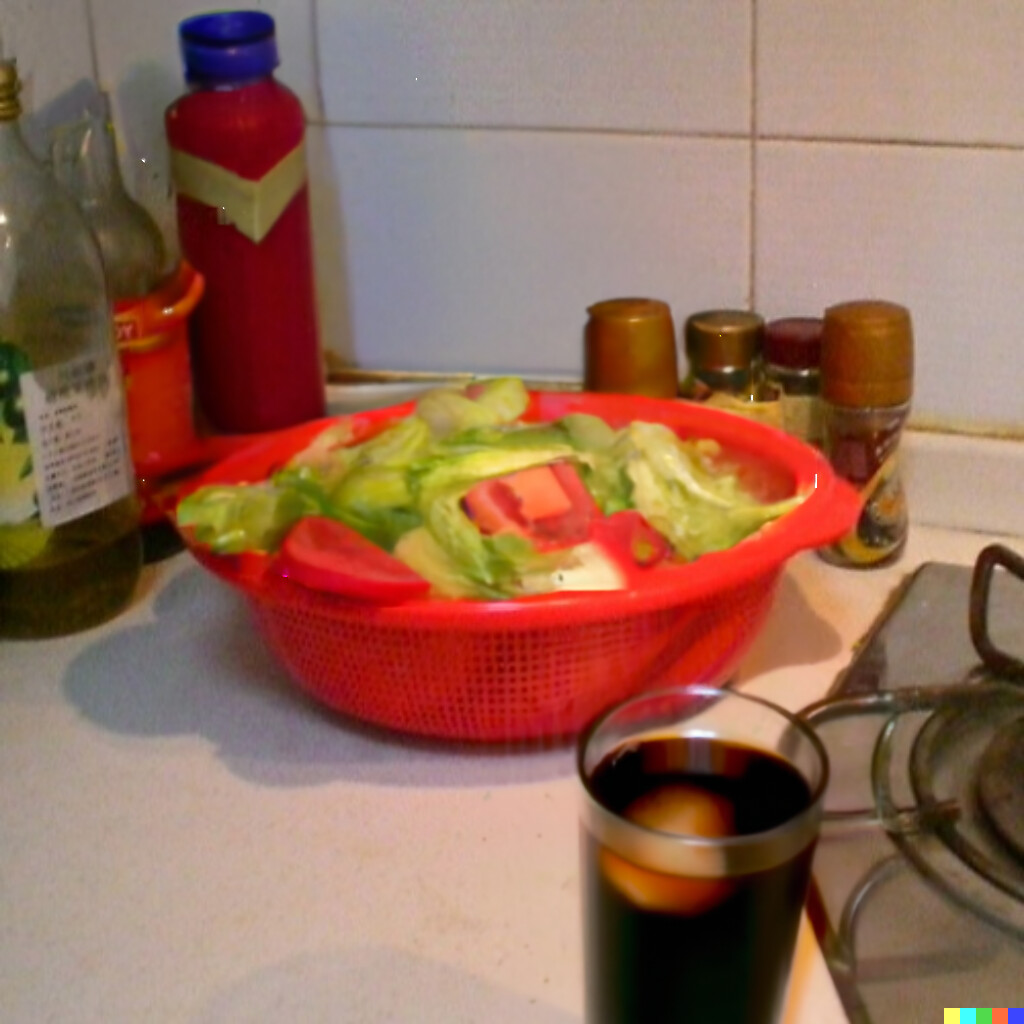}&
        \includegraphics[width=0.14\textwidth]{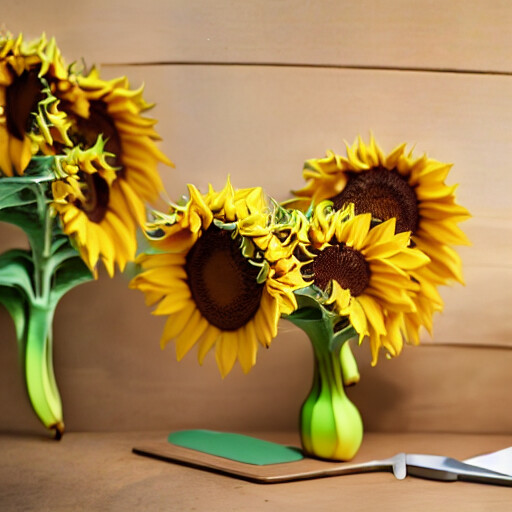}&
        \includegraphics[width=0.14\textwidth]{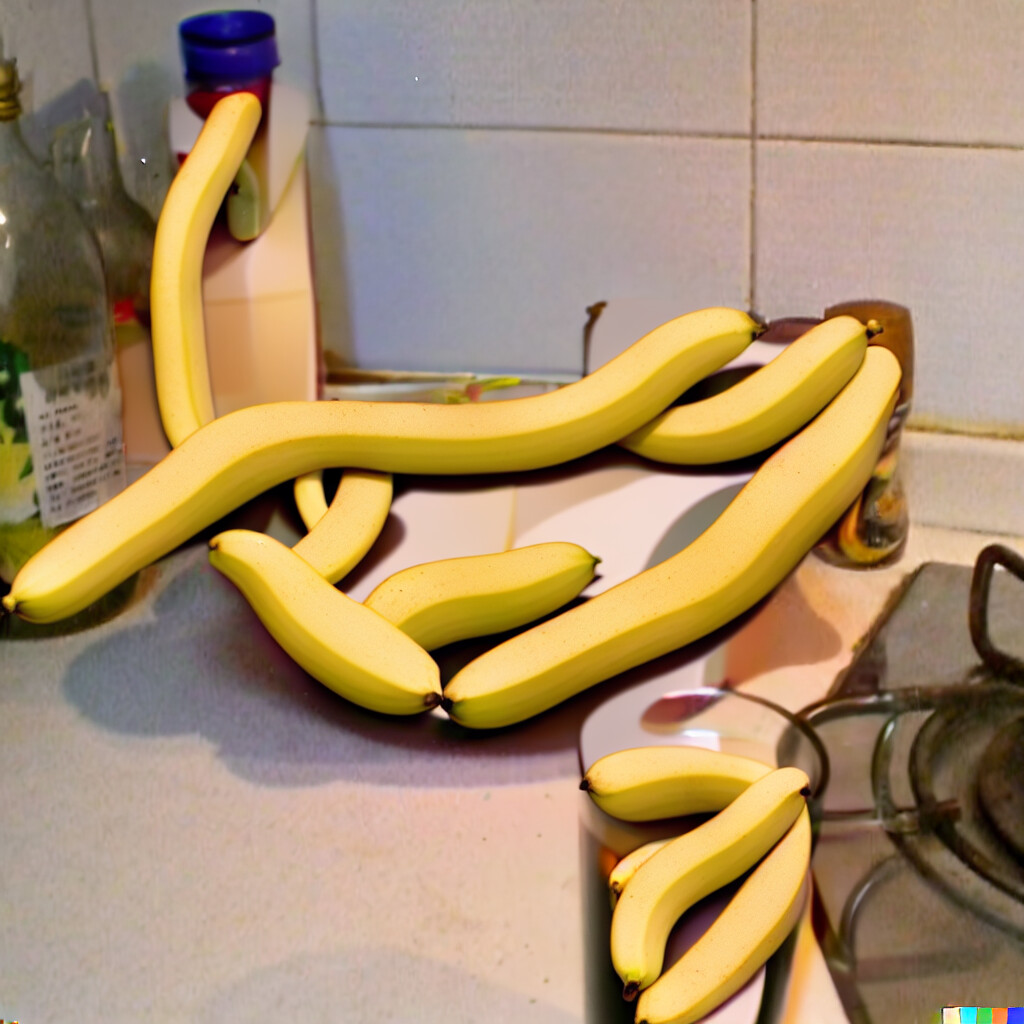}&
        \includegraphics[width=0.14\textwidth]{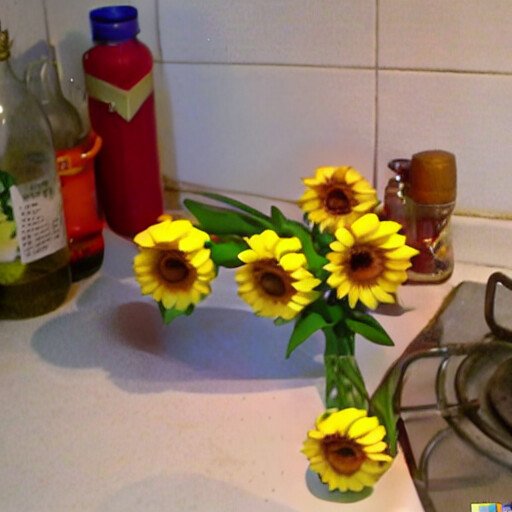}&
        \includegraphics[width=0.14\textwidth]{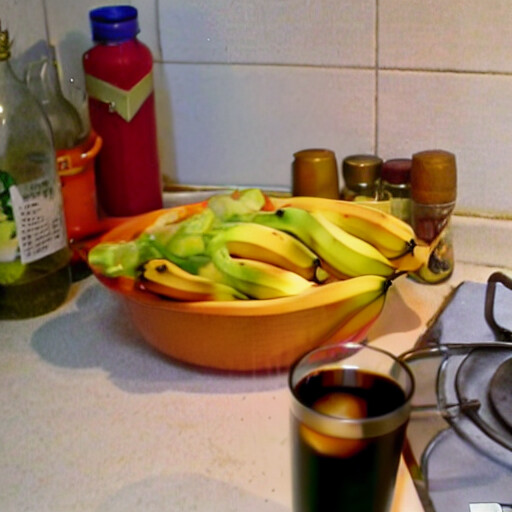}&
        \includegraphics[width=0.14\textwidth]{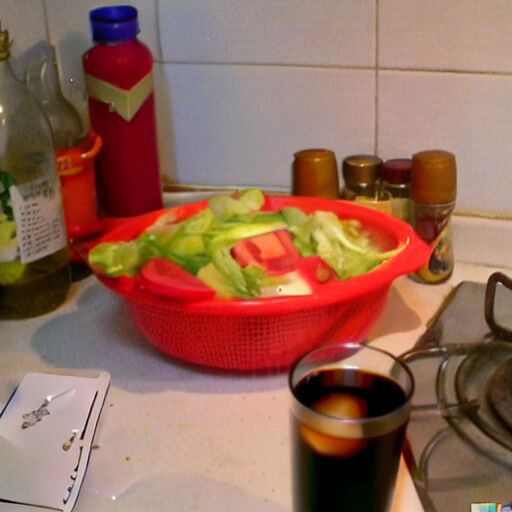}&
        \shortstack{
        \includegraphics[width=0.06\textwidth]{figure/negmask.jpg}\quad\includegraphics[width=0.06\textwidth]{figure/negmask.jpg}\quad\includegraphics[width=0.06\textwidth]{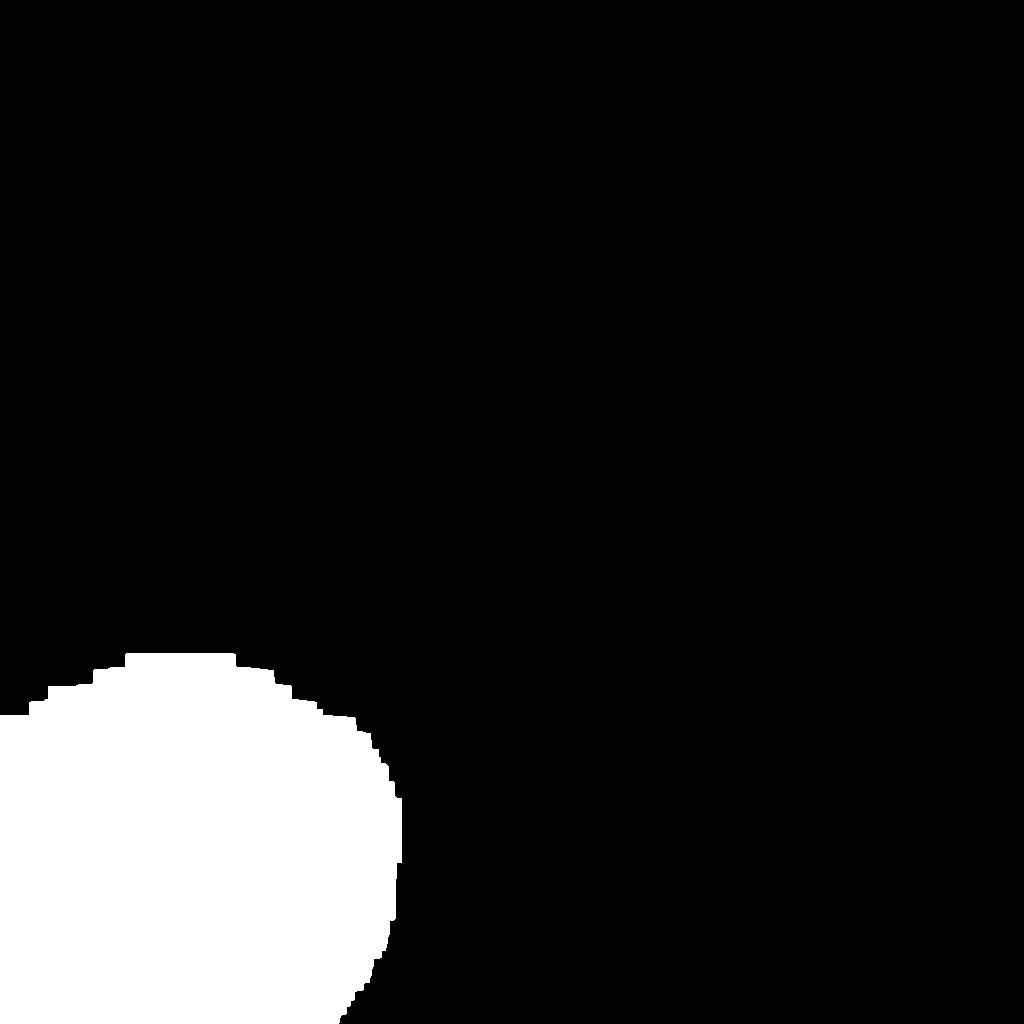}\\
        \texttt{[NEG]~}~\texttt{[NEG]~}\texttt{[MASK]}}
        \\
        \multicolumn{7}{c}{(h) Edit instruction: \emph{``Change the arrangement of the bananas next to the knife.}} \\
        \multicolumn{7}{c}{\emph{Moreover, replace the flowers in the vase with fresh sunflowers. As well, add a clipboard.''}}\\
        
    \end{tabular}}
    \caption{\textbf{Qualitative comparisons for context-aware instruction task}}
    \label{fig:appendix_qualitative_wrong}
\end{figure*}

We present additional qualitative results in \cref{fig:appendix_qualitative_single,fig:appendix_qualitative_multi,fig:appendix_qualitative_wrong}, highlighting comparisons with other baselines~\cite{brooks2023instructpix2pix,fu2024guiding,huang2023smartedit,guo2024focus} across single-instruction, multi-instruction, and context-aware instruction image editing task, respectively. Additionally, we include both the MLLM token outputs and the corresponding decoded binary mask results for further analysis.

\section{Broader Impacts}
\label{appendix:broaderimpacts}
The integration of a mechanism to filter non-applicable instructions in \name has important social implications. By preventing the execution of irrelevant or misleading inputs, the system reduces user frustration and promotes a more intuitive editing experience. This contributes to broader accessibility, enabling individuals with limited technical expertise or physical impairments to achieve their editing goals through natural language alone. Moreover, the ability to reject non-executable instructions helps safeguard against misuse, fostering more responsible and trustworthy deployment of generative image editing technologies.

\section{License of Assets}
\label{appendix:licenseofassets}
In our study, we utilize several instruction-based image editing models~\cite{brooks2023instructpix2pix,fu2024guiding,huang2023smartedit,guo2024focus} as baselines. Specifically, InstructPix2Pix~\cite{brooks2023instructpix2pix} is released under the Creative ML OpenRAIL-M license, as it is built upon Stable Diffusion~\cite{rombach2022high}. MGIE~\cite{fu2024guiding} is distributed under the Apple Custom License. SmartEdit~\cite{huang2023smartedit} is available under the MIT License. The licensing information for FoI~\cite{guo2024focus} was not specified in the available resources.

For training and evaluation, we employ the MagicBrush~\cite{zhang2024magicbrush} dataset, which is released under the Creative Commons Attribution 4.0 License. This license permits users to share and adapt the dataset for any purpose, including commercial use, provided appropriate credit is given and any changes made are indicated. Building on this dataset, we construct a new context-aware image editing dataset that enables the evaluation of models under various scenarios.

\end{document}